%% file: acl_latex.tex
\pdfoutput=1

\documentclass[11pt]{article}

\usepackage[final]{acl}

\usepackage{times}
\usepackage{latexsym}
\usepackage{array}
\usepackage{arydshln}
\usepackage{tikz}
\usepackage{floatpag}
\usepackage{makecell}
\usepackage{booktabs}
\usepackage{multirow}
\usepackage{subfig}
\usepackage{stackengine}
\usepackage[]{graphicx}
\usepackage{graphics}
\usepackage[dvipsnames]{xcolor}
\input{latex/math_commands}

\usepackage{verbatim}
\usepackage{natbib}

\usepackage[T1]{fontenc}

\usepackage[utf8]{inputenc}
\usepackage{csquotes}

\usepackage{microtype}

\usepackage{inconsolata}

\usepackage{todonotes}

\title{Patches of Nonlinearity: Instruction Vectors in Large Language Models}

\author{
  \textbf{Irina Bigoulaeva},
  \textbf{Jonas Rohweder}, 
  \textbf{Subhabrata Dutta}, 
  \textbf{Iryna Gurevych} \\
  Ubiquitous Knowledge Processing Lab (UKP Lab) \\ Department of Computer Science, Technical University of Darmstadt \\ and National Research Center for Applied Cybersecurity ATHENE, Germany \\
  \texttt{\small www.ukp.tu-darmstadt.de} \\ [-0.1mm]
\\
}

\begin{document}
\maketitle
\begin{abstract}
Despite the recent success of instruction-tuned language models and their ubiquitous usage, very little is known of how models process instructions internally. In this work, we address this gap from a mechanistic point of view by investigating how instruction-specific representations are constructed and utilized in different stages of post-training: Supervised Fine-Tuning (SFT) and Direct Preference Optimization (DPO). Via causal mediation, we identify that instruction representation is fairly localized in models. These representations, which we call \textit{Instruction Vectors} (IVs), demonstrate a curious juxtaposition of linear separability along with non-linear causal interaction, broadly questioning the scope of the linear representation hypothesis commonplace in mechanistic interpretability. To disentangle the non-linear causal interaction, we propose a novel method to localize information processing in language models that is free from the implicit linear assumptions of patching-based techniques. We find that, conditioned on the task representations formed in the early layers, different information pathways are selected in the later layers to solve that task, i.e., IVs act as \textit{circuit selectors}.\footnote{We make our code available at: \url{https://github.com/UKPLab/acl2026-instruction-vectors}.}

\end{abstract}

\pagenumbering{gobble}

\section{Introduction}
\label{sec:intro}

\begin{figure}[htb!]
    \centering
    \includegraphics[width=\linewidth,scale=1.0,trim={0.7cm 0.6cm 1cm 0.6cm},clip]{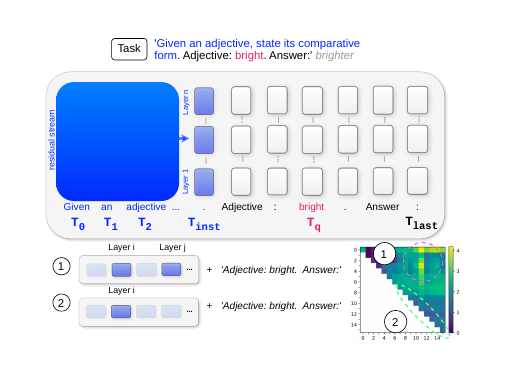}
    \caption{
    We locate instruction vectors (IVs) in the residual stream representations of the final instructional token, $\emT_\text{inst}$. Across models and tasks, we find that $\emT_\text{inst}$ stores sufficient instruction information, and that layerwise representations are more effective in combination (1) than alone (2), i.e. IVs are \textit{superadditive} (§ \ref{sec:identifying_ivs}).
    }
    \label{fig:main_figure}
\end{figure}

Instruction tuning has emerged as a staple of post-training in recent years, with some form of it now present in nearly all state-of-the-art LLMs. Originally conceived as a supervised fine-tuning (SFT) technique \cite{training-lms-follow-instrs, wei2022finetuned}, instruction tuning has since been implemented in many variations, which range from alternate loss calculations \cite{lossoverinstructions, chatterjee2025effectinstructiontuningloss}, to using adapters \cite{task-adapters-instructions} or preference optimization \cite{deepseek}.
Prior works have investigated the differences between base and instruction-tuned models in terms of output patterns and task performance \cite{zhou-lima-superficial, ghosh2024closerlooklimitationsinstruction}. Fine-tuning in general (which is the most commonplace implementation of instruction tuning) has been shown not to introduce new capabilities into a pretrained model, rather reusing existing ones \cite{jain2024mechanistically}.

However, one fundamental question remains unaddressed: By what mechanism do models represent and process instructions? Isolating this mechanism will enable an accurate diagnosis of whether a model has successfully acquired the ability to follow instructions for a given task, and which post-training methods (if any) are necessary for this capability to appear. Furthermore, instruction-following is a pillar of alignment; safer, accountable deployment of LLMs necessitates a thorough understanding of how instructions are processed by the language model.

Prior studies identify that models construct localized neural digests of input-output mappings from in-context examples \cite{hendel-etal-2023-context, todd2024function, li2025justintime}. These digests, commonly denoted as \textit{function vectors}, restore the task context when provided during a new forward pass, even without any in-context examples. Recently, \citet{davidson2025differentpromptingmethodsyield} demonstrate that similar transferrable digests can be associated with instructions as well, realized at different yet overlapping attention heads of the model. 

We formulate our starting point motivated by these works, and note the following gaps: 1) Function vectors, identified at the last token of the input prompt, are the end-products of the internal computation, but the process of their formation, geometric properties, and mechanism of action remain unknown; 2) Steering vector approaches (either in the model activation space~\citep{DBLP:conf/iclr/StolfoBYHN25} or in SAE feature space~\cite{he2025saifsparseautoencoderframework}) identify optimal perturbation locations that elicit superior instruction following, albeit without providing additional knowledge about it.

{\bf Contributions and findings.} In this work, we seek to provide a mechanistic exploration of instruction following by investigating how models represent instructions.
We discover that models construct localized digests, representative of the instruction, right after processing the instructional context   (Section~\ref{sec:identifying_ivs}). The task information in these digests is linearly recoverable (i.e., there exist linear hyperplanes that discriminate the digests corresponding to semantic paraphrases of two tasks). However, the action of these digests is \textit{superadditive} in nature: there exists a combination of layers whose cumulative causal influence is more than the sum of their individual influences. This superadditivity signals a non-linear interaction between the linearly-separable digests. This limits the applicability of existing causal discovery methods, which presume an additive interaction between the variables, for investigation of how these instruction digests are utilized. 

To mitigate this challenge, we propose a novel method of understanding information flow within a Transformer-based language model (Section~\ref{sec:path_experiments}) that is free from additive interactions between model components. Using this method, we identify that the neural digests play the role of \textit{circuit selectors} to process the query. While a pretrained model can construct the same task-specific neural digests, instruction-tuning is indispensable for introducing the circuit-selection ability into those digests. 

{\bf Broader significance.} The fact that localized instruction digests are constructed independent of the query (i.e., eager computation instead of just-in-time computation together with the query) points towards the possibility of information bottleneck~\citep{tishby2000informationbottleneckmethod} in instruction processing. Such a bottleneck can be utilized to increase robustness against adversarial perturbations in the instruction. Superadditivity of task digests bears foundational implications for interpretability research: If the influence of a variable $X$ depends on another, non-local variable $Y$, then single-component causal attributions can undervalue $X$ by recording a low impact of $X$ when $Y$ is not active. This calls for an immediate evaluation of how we think of mechanistic interpretability
models.

\section{Related Work}
\label{sec:relwork}

Several recent works have investigated how models represent tasks. \citet{todd2024function} find that models form compact representations of tasks, specifically from in-context learning (ICL) examples. 
\citet{davidson2025differentpromptingmethodsyield} extend this investigation to instructions and find that ICL- and instruction-based task representations activate different attention heads, but are beneficial when used together. An orthogonal yet noteworthy study by \citet{huang-etal-2024-chat} represents instruction-following as differential directions in the parameter space. Additionally, various works have examined the properties of instruction-tuned models, from the standpoint of behavioral differences to base models \cite{wu-etal-2024-language} and whether a particular dimension can be isolated that yields instruction-following behavior \cite{heo2024do}. Nevertheless, a gap is left in this research area, namely, how models form instruction representations and what the causal and geometric properties of these representations are.

A substantial majority of mechanistic interpretability research implicitly or explicitly associates abstract causal roles with singleton components (representation, subspace, parameter matrix such as attention heads, etc.): name-mover heads~\citep{wang2023interpretability}, function vectors~\citep{todd2024function}, factual association in early-middle MLPs~\citep{meng-rome}, etc. Circuit discovery~\citep{wang2023interpretability, conmy2023towards, ameisen2025circuit} in its current form is built on top of this assumption -- these algorithms formalize the forward pass of a model as a Directed Acyclic Graph (DAG) and search for a connected subgraph responsible for the behavior under investigation. Findings of \citet{sutter2025the} warn of the fallibility of causal abstraction without strong assumptions on how language models encode features. 

A current popular theory of model representations is the \textit{linear representation hypothesis}, which states that models represent \enquote{high-level concepts} as linear subspaces \cite{elhage2021mathematical, open-problems, gurnee2024language, nguyen2025toward}. Theoretical frameworks have additionally been developed that seek to formalize the ways in which concepts interact linearly within a model's representation space \cite{pmlr-v235-park24c, nguyen2025toward}. Our findings identify a critical orthogonality in this area: despite linear representation, causal variables can interact synergistically, rendering the DAG model of causality invalid (since a DAG allows a causal edge to connect two nodes, it cannot capture the scenario when two variables synergistically define a third variable). Subsequently, our proposed solution, via locally linear maps, searches for a suitable subset of the functions jointly implemented by a Transformer, without any component-specific abstraction. Note that our results go beyond the existing argument related to sparse vs. distributed causal structures~\citep{bahador2025localizeddefinitionsdistributedreasoning}. Even in a distributed setup, a single abstraction of the causal variable is shared across multiple layers/locations~\citep{lindsey2024sparsecrosscoders}. Each instance of the distributed variable interacts additively, whereas our discovery points toward a completely different organization of causal variables that promote superadditivity.

\section{Models and Tasks}
\label{sec:models-task-info}

\paragraph{Models.}

We use OLMo-2 \cite{olmo20252} models due to their clear progression from the base, SFT, and DPO variants -- e.g. the SFT model was built from the base model, and the DPO model was built from the SFT model. This enables us to follow the impact of the typical post-training process when comparing the models.

We implement our experiments using NNSight \cite{fiottokaufman2024nnsightndifdemocratizingaccess}, setting library hyperparameters to ensure deterministic behavior and reproducibility. We elaborate on GPU architecture and experiment runtimes in Appendix \ref{app:sec:experiment-runtimes}.

\paragraph{Tasks.}

\begin{table*}
    \centering
    \small

    \begin{tabular}{cp{2.5cm}p{10cm}}
    \toprule
    Task & Subtask & Instruction \\
    \midrule
      \multirow{2}{*}{Adjectives}  & Comparative & `Given an adjective, state its comparative form.'\\
         & Antonym  & `Given an adjective, state its antonym.'\\
      \midrule
      \multirow{2}{*}{Animals}  & Color & `Given an animal, state its most typical associated color.'\\
         & Can\_Fly & `Given an animal, state whether or not it can fly. Print 'yes' or 'no'.'\\

\midrule
         
      \midrule
        \multirow{4}{*}{BigBench}& Metaphor\_Boolean & `For a given metaphorical sentence, identify if the second sentence is the correct interpretation. Print Y for yes and N for no.'\\
           & Implicatures & `Predict whether Speaker 2's answer to Speaker 1 counts as a yes or as a no. Print Y for yes and N for no.'\\
         & Object\_Counting & `For the given sentence, count the number of objects listed and print it as a digit.'\\
         & Snarks & `Choose which of the two statements is sarcastic and print the corresponding letter:'\\
         \bottomrule
    \end{tabular}
    \caption{Our eight tasks with their corresponding instructions. During tokenization, the final token is always the final punctuation mark (either a fullstop or a colon).}
    \label{tab:contrastive_tasks}
\end{table*}

We generate two \textit{contrastive task pairs} -- i.e. pairs of tasks in which the target query is the same, while the instructions differ (see Table \ref{tab:contrastive_tasks}). In each case, the model processes an identical query but must output a different token as the answer. This is done to minimize the variation in the prompts apart from the instruction and follows the task design of similar works \cite{hendel-etal-2023-context}. We generate these tasks using ChatGPT-5-nano \cite{chatgpt5-nano} and manually evaluate the correctness of each sample ($\sim$200 samples per task).

Additionally, to show generalization, we use four tasks from the BigBench benchmark \cite{srivastava2023beyond}: \textsc{metaphor\_boolean},  \textsc{implicatures},  \textsc{object\_counting}, and  \textsc{snarks}. 

\paragraph{Confirming Instruction Following Ability.}
\label{sec:inference-experiments}
When investigating instruction representations, an important confounding factor to consider is the model's competence in a given task. If a model is unable to solve a task, then it is reasonable to suppose that it will also not have instructional representations for that task, or that the properties of these representations will differ from those of a more capable model. Thus, we evaluate each model's performance on these tasks using exact match accuracy (EMA).

Additionally, we establish a metric called \textsc{instructional accuracy} (IA), which measures whether the model's answer falls within a manually-defined scope of valid response types for the instruction, even if its output answer is incorrect. A detailed elaboration of this metric can be found in Appendix \ref{app:sec:instr_acc}.

We find that most models are able to follow instructions for most tasks with a high degree of accuracy (IA $>$ 50\%, see Figure \ref{fig:basic_acc} in Appendix \ref{app:sec:instr_acc}). This confirms that our choice of tasks is appropriate for the models, and implies that the models can in fact represent and process instructions for these tasks.

\section{Preliminaries}
In this section, we provide the theoretical preliminaries for our experiments.

Let $\sV$ be an indexed vocabulary of tokens where every token $v\in \sV$ is represented as the standard column basis vector $\ve^{(v)}\in \sR^{\lvert\sV\rvert}$ (i.e., the $v$-th element is 1, the rest are 0). An $n$-length input token sequence is defined as $\mT:={\emT_1, \cdots, \emT_n} \in \sV^*$. The forward pass computation of the $l$-th layer of an auto-regressive transformer with $H$ attention heads, for $l\in \{1, \dots, L \}$, can be written as follows:
\begin{equation}
    \begin{split}
        &\emX_i^{l, \text{att}} = \sum_{j} \eva^{l, h}_{i, j}\mW^{l, h}_{OV} \emX^\text{l}_j\\
        & \mX^{l,\text{mid}} = \operatorname{Norm}(\mX^{l, \text{att}} + \mX^l)\\
        & \mX^{l,\text{mlp}} = \mW_{2} \operatorname{PLin} \left(\mW_1\mX^{l,\text{mid}}\right)\\
        & \mX^{l+1} = \operatorname{Norm}(\mX^{l,\text{mid}} + \mX^{l,\text{mlp}})\\
    \end{split}
    \label{eq:standard-transformer}
\end{equation}
where $\mX^\text{l}, \mX^{l,\text{att}}, \mX^{l,\text{mid}}, \mX^{l,\text{mlp}}\in \sR^{n\times d}$ are the most commonly referred-to model representations, $h\in \{1, \dots, H \}$ is the index of the attention head, and $\eva^{l, h}_{i, j}\in \va^{l, h}_{i}$ is attention between $i$-th query and $j$-th key. We follow \citet{elhage2021mathematical}'s reparameterization of multi-head attention: For each attention head, the value and output projections are folded into a single transformation $\mW^{l, h}_{OV}\in \sR^{d\times d}$.  $\operatorname{Norm}$ denotes learnable normalization operation (LayerNorm or RMSNorm). Parameters $\mW^{l}_1\in \sR^{d'\times d}$, $\mW^{l}_2\in \sR^{d\times d'}$, and piecewise-linear function $\operatorname{PLin()}$ (e.g. ReLU, GELU) together construct the MLP block.

The input to the first decoder block is computed as $\mX^1 = \mX^\text{emb} = \mW_{E} \mT$, where $\mW_E\in \sR^{d\times\lvert\sV\rvert}$ is canonically called the embedding that maps each token to the $d$-dimensional representation space. The final representation, i.e., $\mX^{L+1}$ is mapped to the token logit space via multiplying by $\mW_U$ (commonly referred to as the unembedding projection).

Unless mentioned otherwise, the input token sequences in our setup can be formalized as $\mT^\text{full}:=[\mT^\text{inst}\mT^\text{q}]$, where $\mT^\text{inst}$ is the instruction describing the task and $\mT^\text{q}$ is the query information.
Since we are primarily concerned with the token at the very end of the instruction segment and the last token of the prompt (See Section \ref{sec:identifying_ivs}), we refer to these token positions as $\emT_\text{inst}$ and $\emT_\text{last}$, respectively. See Figure \ref{fig:main_figure} for an illustration of our approach.

Finally, we denote the abstract forward pass of the model as 
\[\vy = {\cal F}(\mT; \Theta)\]
where $\Theta$ denotes the set of interventions (on representation or parameter space of the model) applied. For example, switching off the $h$-th attention head on $l$-th layer can be represented as $\Theta := \mW^{l,h}_{OV}\leftarrow \vzero$.

\section{Localizing Instruction Representations}
\label{sec:identifying_ivs}

We start with posing a fundamental question about the mechanism of instruction following: Are instructions processed \textit{just-in-time} (i.e., simultaneously or after processing the query) or \textit{eagerly} (i.e., model constructs neural digests of the instruction before processing the query)? The former would be realized with distributed computation within the model, whereas the latter would result in localized representations of the instruction.

We start with causal variable discovery using activation patching~\citep{zhang2024towards} to check for localizability of the instruction representations before the query tokens. Note that our primary focus in this stage is to determine the synthesis location of instruction digests (if any exist) and not their downstream usage mechanism. Therefore, we restrict ourselves to the residual stream representations ($\mX^l$ in our formalism) instead of finer granularity like attention head or MLP neuron outputs.

\subsection{Causal Mediation Analysis}
\label{subsec:identifying_ivs_cma}

\begin{figure*}[hbt!]
    \centering

    \includegraphics[width=\linewidth,scale=0.1,trim={0 15cm 0cm 0cm},clip]{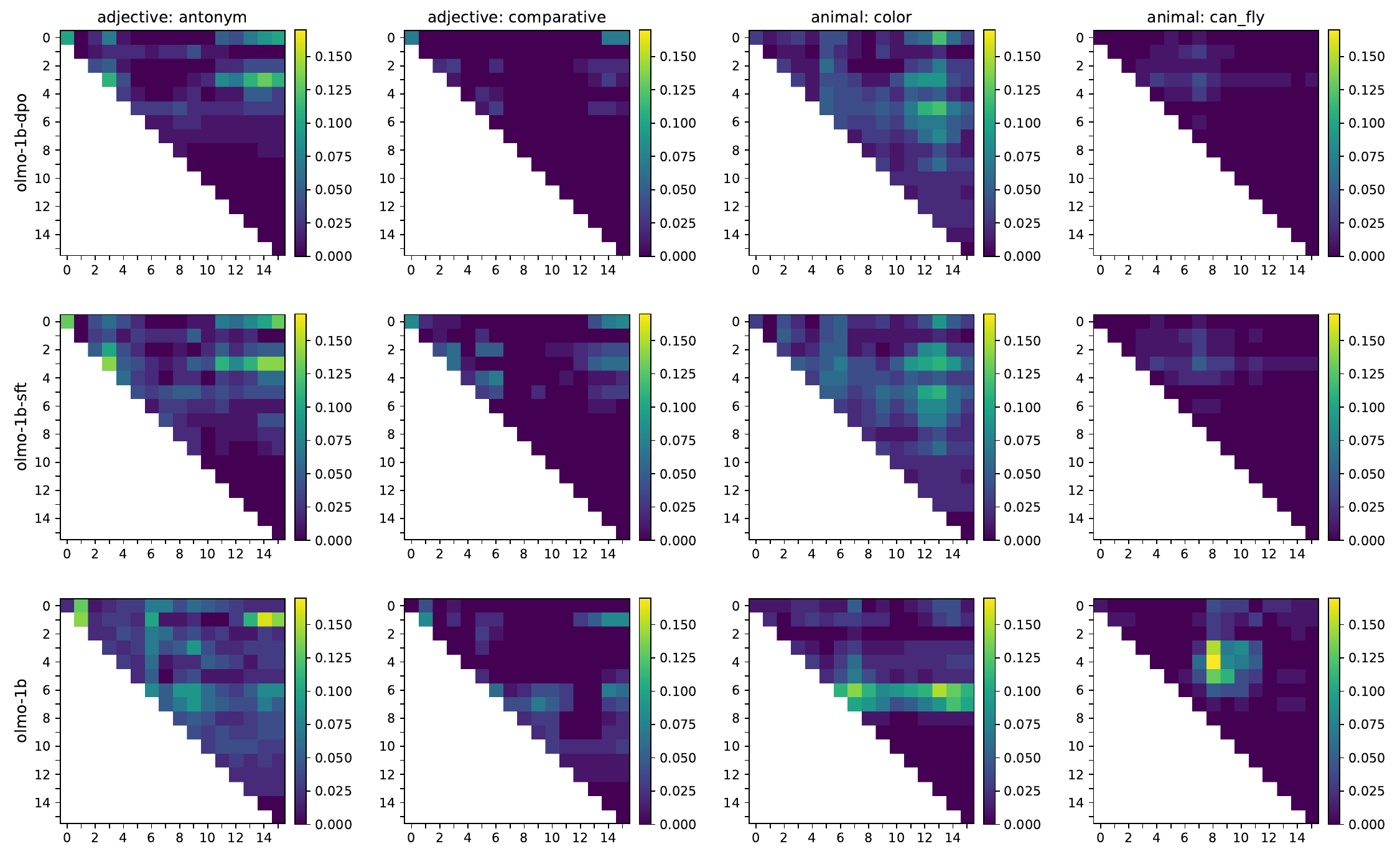}

    \caption{Effects of 1- and 2-layer patching configurations on the reciprocal rank of the target token. Each square of the x- and y-coordinate grid represents the corresponding layers of the model being patched. Coordinates where x=y represent 1-layer patching. Two important properties are shown. \textit{Localization:} Across all tasks, we observe localized points where the logit improvement is the greatest. \textit{Superadditivity:} Two-layer combinations bring greater improvements than single layers. Generalization to 7B models and other tasks is shown in the appendix.}
    \label{fig:norm_rank_pt-vs-to_1b}
\end{figure*}

Our adopted causal variable localization strategy closely follows prior work by \citet{meng-rome}. For any given internal variable $\mX^l$, causal mediation starts with two distinct forward pass:
\[\vy^\text{source} = {\cal F}\left( \mT^\text{full}; \emptyset \right)\quad \vy^\text{target} = {\cal F}\left( <\!\!s\!\!>\mT^\text{q}; \emptyset \right)\]
where $<\!\!s\!\!>$ is a filler token. Next, we run the forward pass under intervention:
\[\vy^\text{patch} = {\cal F}\left( \mT^\text{q}; \mX^{l}_{<s>} \leftarrow \mX^{l, \text{source}}_{\emT^\text{inst}} \right)\]
The causal effect of the variable $\mX^{l}$ can then be quantified as $d\left(\vy^\text{patch},  \vy^\text{target}\right)$, where $d$ is suitable metric of choice, conditioned on the answer token. Specifically, we use two metrics: (1) difference between the reciprocal rank of the answer token, and (2) difference between the absolute logit values of the answer token\footnote{Our localization is primarily based on reciprocal rank. Logit difference alone can be confounded due to high-norm representations that simply increase the logit value of all tokens.}, in $\vy^\text{patch},  \vy^\text{target}$. 

In summary, our setup checks for localizability of instruction digests by testing whether the model can recover the correct answer from the query context and a small subset of the instructional representations.

\paragraph{Superadditivity of IVs.}
\label{subsec:patching_results}
We conduct single- and multi-layer activation patching\footnote{We describe multi-layer settings as tuples, e.g. (1, 2), but since the layers are patched simultaneously, the tuple ordering does not matter. In other words, (1, 2) is the same as (2, 1). Figure \ref{fig:norm_rank_pt-vs-to_1b} and the other heatmaps are accordingly left blank in the redundant grid areas.}. For OLMo-2 1B models, Figure \ref{fig:norm_rank_pt-vs-to_1b} shows the effects of 1- and 2-layer patching on the reciprocal rank of the target token, while Figure \ref{fig:norm_logit_pt-vs-to_1b} shows the effect on the logit of the target token. The scale is normalized to better elucidate differences between the base and post-trained checkpoints; unnormalized heatmaps are presented in Appendix \ref{app:sec:numerical_scores}. 

We find that the greatest improvements in reciprocal rank tend to occur in localized areas -- for example, in the layer combinations (3, 7) and (3, 4) for the \textsc{animal: can\_fly} task (see Figure \ref{fig:norm_rank_pt-vs-to_1b}). However, when these individual layers are patched by themselves, the relative rank improvement is much lower. The same is observed with the OLMo-2-7B models for both rank and logit (see Figures \ref{fig:app:1LP+2LP_logit_olmo27b_olmo3} and \ref{fig:app:1LP+2LP_rank_olmo27b_olmo3} of the appendix).

In other words, we find that the instructional representations at various individual layers have a greater effect on the rank and logit of the target token when acting together in combination, rather than when their individual contributions are summed together. In abstract terms:
\begin{equation}
\begin{split}
f(\vx_{i} \cup \vx_{j}) \geq f(\vx_{i}) + f(\vx_{j})
\end{split}
\label{eq:superadditivity_inequality}
\end{equation}
for two given causal variables $\vx_i, \vx_j$,
and $f$ conceptually denotes the causal effect function implemented by the model.

To more rigorously test whether this inequality holds, we perform a $t$-test on the numerical scores. We find that, for all models and tasks, the inequality indeed holds to a significant degree across samples (see details and results in Tables \ref{tab:app:ttest-olmo1b-contrastive}-\ref{tab:app:ttest-olmo7b-bigbench} in Appendix \ref{app:sec:numerical_scores}). This result indicates that the mechanism of $L_{i}$ and $L_{j}$'s interaction is not a linear operation (e.g. addition), but rather one that is nonlinear. In other words, instruction vectors are \textit{superadditive.}

\paragraph{Base vs. Post-Trained Checkpoints.}
\label{subsec:base-vs-post-comparison}
Since we observe superadditivity across all model checkpoints, it is likely that the multi-component nature of instructional representations is a fundamental property, and that the role of post-training is not to change their nature, but rather to refine them. In particular, we notice that the post-trained models tend to form representations in similar layer groups, but at different ones than the base model (see Figure \ref{fig:norm_rank_pt-vs-to_1b} and Appendix \ref{app:sec:numerical_scores}), and that they are more \enquote{spread-out} than those of the base model. This implies that post-trained models rely on more complex multi-component interactions. Model size is also an influencing factor: In the 7B models, the post-trained checkpoints have higher logit and rank improvements than the base model, while the 1B models have a noisier variation among the checkpoints.

\paragraph{Geometric Properties of IVs.}
\label{subsec:spatial_analysis}
Finally, we wish to determine how the instructional representations of our eight tasks are geometrically configured in the models' representation space. For each of our contrastive tasks, we produce 200 rephrasings of the instruction to test robustness against within-task variation. We generate these using ChatGPT-5-nano \cite{chatgpt5-nano} and manually check each sample for correctness. For each instruction, we extract the residual stream representations using the same procedure as in Section \ref{sec:identifying_ivs}.

We plot the LDA representations of our models and find that the instructional representations form well-defined task clusters, albeit ones that are not linearly separable in 2 dimensions (see Appendix \ref{app:sec:vector_space_graphs}). These results partially contrast with prior works, which found task clusters that were linearly separable in two-dimensional TSNE representations \cite{hendel-etal-2023-context}.

To confirm whether or not our task clusters are linearly separable in general, we train a simple linear probe on a portion of the representations and attempt to predict the task cluster given a test sample. We find that our linear probe has high accuracy for all task clusters (98\%$<$). This implies that the instruction vectors do form linearly separable clusters, but potentially at a high dimension that was not captured by our dimensionality reduction method. Furthermore, the fact that this separation is present in both base and post-trained checkpoints implies that post-training does not newly introduce this geometric property: Base models also distinguish between instructions of different tasks.

\section{Causal Mechanism of Instruction Vectors}
\label{sec:causal_experiments}

In the previous section, we demonstrated the localization of instruction vectors and showed that these representations are indeed used for instruction following. However, it remains unclear \textit{how} exactly they are utilized. 

Existing methods of discovering causal mechanisms (i.e. circuit discovery) rely on repeated, component-wise causal mediation to disentangle the task-specific causal graph within the model~\cite{wang2023interpretability}. However, the distributed, synergistic interaction of instruction vectors imply that such component-induced graphical models will produce misleading mechanisms. 
\begin{figure}[!t]
    \centering
    \includegraphics[width=\linewidth,scale=0.2,trim={0.6cm 0.4cm 0.5cm 0.4cm},clip]{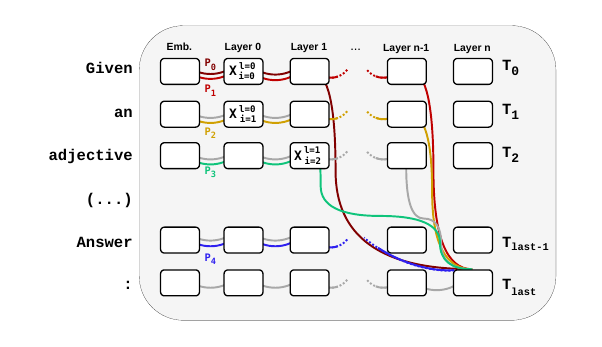}
    \caption{Conceptual decomposition of the Transformer as a collection of locally-linear, token-to-token maps, indicating how information flows through the model. For each layer and token position, high-ranking paths (in color) to the output token may exist. Other paths (in gray) may be low-ranking, or may not lead to the target token.}
    \label{fig:path_dia}
\end{figure}
\begin{figure*}
    \centering
    \includegraphics[width=\linewidth,scale=0.5,trim={0cm 0.2cm 0cm 0cm},clip]{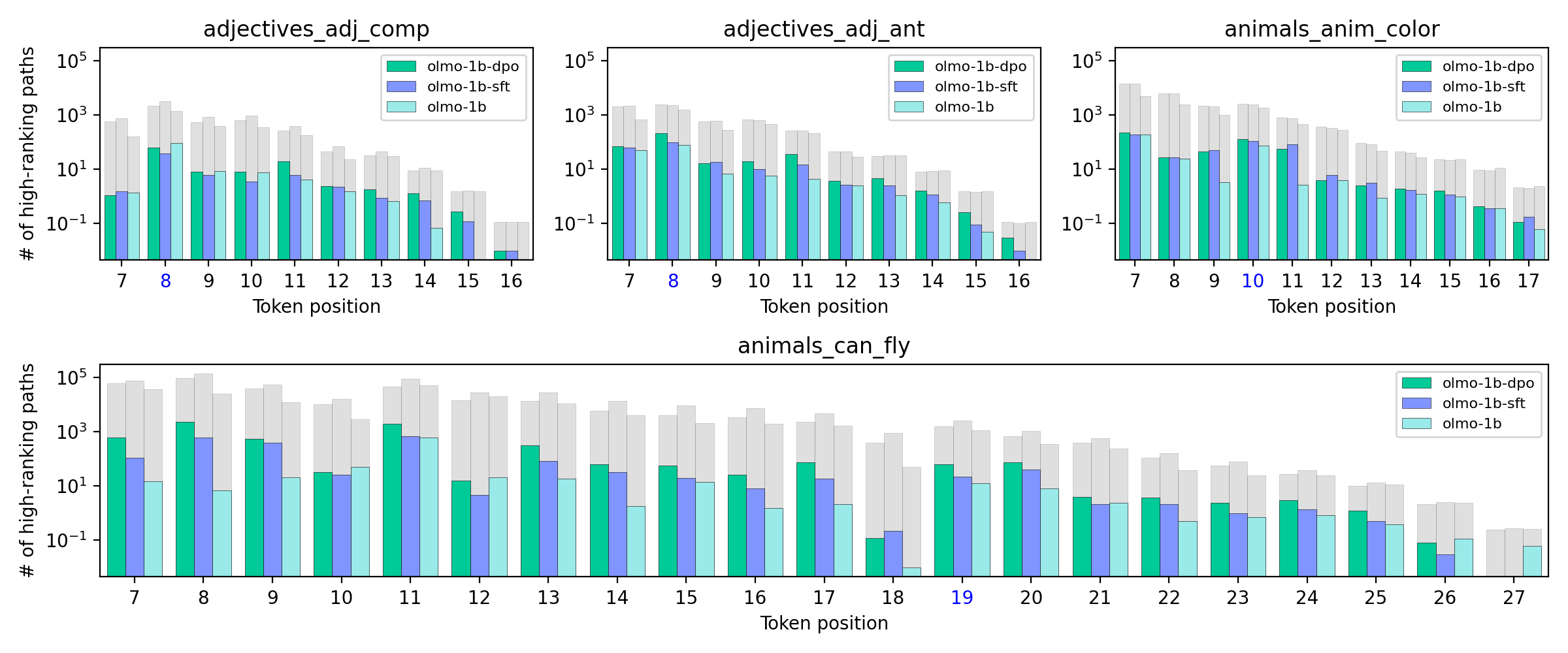}
    \caption{Path contribution by token position for 1B models. For each task, we examine a subset of token positions in the prompt, and for each token position, we average the number of high-ranking paths over 100 task samples (in color). The average total number of paths over the 100 samples is shown in gray. The $\emT_\text{inst}$ tokens for each prompt are highlighted in blue. Plots highlighting the percentage contributed by each token are shown in Appendix \ref{app:sec:path_analysis_topk}.}
    \label{fig:path_contrib_by_token}
\end{figure*}
\begin{figure}[!t]
    \centering
    \includegraphics[width=\linewidth,scale=0.7,trim={0.6cm 0cm 0.5cm 0cm},clip]{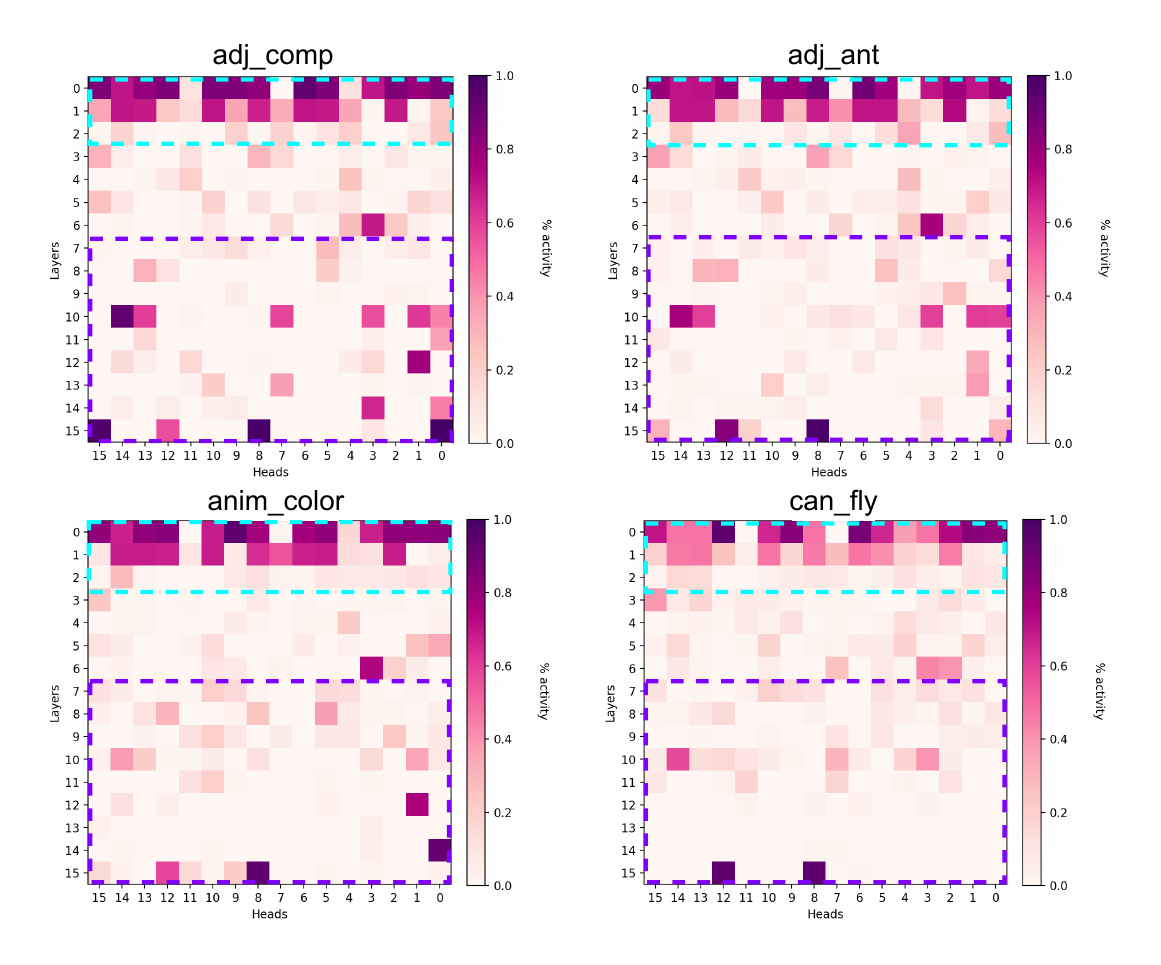}
    \caption{Attention head activity for OLMo2-1B across the contrastive tasks. The values represent how often a certain head is active (\% over 100 samples, significant). We highlight the areas where the head activity is similar (blue) and diverges (purple). Note the similarity at the $\emT_\text{inst}$ tokens for \textsc{Adjectives} tasks, and an earlier token of \textsc{anim: can\_fly}, which has a multi-sentence instruction (see Section \ref{sec:path_experiments}).
    }
    \label{fig:attn-heads-analysis}
\end{figure}

{\bf Locally-linear surrogate maps.} To circumvent this, we propose a novel, intervention-free method to attribute the components with task-specific information propagation. In this approach, we represent the Transformer as a collection of interacting, token-token, locally-linear maps. Note that in the absence of the MLP layers,~\citet{elhage2021mathematical} already provide reparameterization of 1- and 2-layer attention only transformers as linear maps in $\sR^{|\sV|\times |\sV|}$. To handle the intermediate non-linear operations such as normalization and MLP, we draw motivation from the neural polytope literature~\cite{black2022interpretingneuralnetworkspolytope, NEURIPS2019_9766527f}. For any given layer $l$, we can rewrite the MLP operation as:
\[\emX^{l,\text{mlp}}_i = \mV^l_i\emX^{l,\text{mid}}_i = \mW_{2} \mD \mW_1\emX^{l,\text{mid}}_i\]
where $\mD\in \sR^{d'\times d'}$ is a diagonal matrix that is a surrogate of $\operatorname{PLin} \left(\right)$ at the point $\mW_1\emX_i^{l,\text{mid}}$. For example, if $\operatorname{PLin} \left(\right)$ is ReLU, then 
\[diag(\mD):=\mathds{1}_{\mW_1\emX_i^{l,\text{mid}}>0}\]
Similarly, normalization can be `linearized' in this manner by treating it as an affine transformation determined by the input mean and standard deviation, resulting in a similar surrogate $\mU^{l}_i$. Note that, while these surrogates allow treating the non-linearities as linear maps, they are restricted within an $\epsilon$-neighborhood of the current input.
Using these surrogates, we can rewrite the $l$-th layer operation (Eq.~\ref{eq:standard-transformer}) as:
\begin{align}
    \begin{split}
        &\emX^{l+1}_i = \mU^{l,\text{mlp}}_i \left(\mI + \mV^l_i\right) 
        \mU^{l, \text{att}}_i\emX^l_i\\
        &+
        \mU^{l,\text{mlp}}_i \left(\mI + \mV^l_i\right) 
        \mU^{l, \text{att}}_i\sum_h\sum_j \eva^h_{i,j}\mW_{h, OV} \emX^l_j
    \end{split}
\end{align}
Structurally, this corresponds to a sum of linear transformations, each corresponding to $2(H+1)$-many different paths from input to output. By stitching together similar paths across layers, we represent the forward pass as a sum of multiple token-to-token locally-linear maps (see Figure \ref{fig:path_dia}).

\subsection{Determining the Mechanism}
\label{sec:path_experiments}

Instead of the component configuration space (i.e., set of all connected subgraphs of the computation graph), we use its dual, token-token function space to localize and characterize the causal mechanism. Since no interventions are applied and interacting paths (i.e., two paths sharing $\mV^l_i$) are allowed to use the shared channel, the effects of these paths remain additive, although instruction vectors themselves are not. 

However, such a function space grows combinatorially with size of the model and input token sequence, making it infeasible to examine each existing path. Thus, for computational practicality, we trace only those paths that gather maximal attention, i.e., we consider an edge (via attention head $h$) between $\emX^l_j$ and $\emX^{l+1}_i$ only when  
\[j = \argmax \va^{l,h}_i\]
For each path $\mP$ that maps input token $\emT_i$ to the output, we can compute the logit contribution as $\mW_U\mP\mE_E\emT_i$. Subsequently, we consider paths that result in answer token rank $<100$. 

We use these token-token maps to conduct three angles of analysis, each of which reveals an aspect by which instructional representations are utilized. Due to memory constraints, we conduct these experiments on OLMo 1B models.

{\bf Path Contribution by Token.} A general observation is the strong sparsity of the traced paths: despite a combinatorial possibility, the number of high-ranking paths (<100) per token remains bounded by ${\cal O}(10^3)$. This high degree of sparsity implies the practical appeal of our proposed path-tracing for causal discovery. Any path emanating from a token closer to the last input token has fewer choices of next node compared to distant ones. Therefore, random distribution of path geometry would result in a strictly decreasing total number of paths as we go from first to last input token. {Yet, across tasks and model variations, we find that $\emT_\text{inst}$-emitted high-ranking paths form the upper bound among other tokens (Figure \ref{fig:path_contrib_by_token}). This indicates that $\emT_\text{inst}$ indeed plays a crucial role in the correct processing of an input instruction, leading to the target output token. However, the exception to this pattern is equally informative: In \textsc{anim: can\_fly}, the token position with the most high-ranking paths is $\emT_\text{11}$, which occurs before $\emT_\text{inst}$. This is explainable due to the greater complexity of this task instruction: \textit{``Given an animal, state whether or not it can fly. Print `yes' or `no'.''} $\emT_\text{11}$ represents the fullstop at the end of the first sentence, while $\emT_\text{inst}$ is the final fullstop. In fact, the first fullstop already forms a complete instruction, while the second sentence supplements it. This reinforces the eager nature of instruction representations and suggests that the models prefer to utilize a complete representation of the sub-instruction ending at $\emT_\text{11}$.
}

\begin{table*}[h]
\centering
\resizebox{\linewidth}{!}{
\begin{tabular}{l *{16}{r}}
\toprule
\multicolumn{17}{c}{Spearman $r$ per layer (k=1 vs k=2)} \\
\midrule
\multicolumn{17}{c}{\textbf{Adjectives: Antonym}} \\
\midrule
 Model & \textbf{\color{magenta}{0}} & \textbf{\color{magenta}{1}} & 2 & 3 & \textbf{\color{magenta}{4}} & \textbf{\color{magenta}{5}} & 6 & 7 & \textbf{\color{magenta}{8}} & 9 & 10 & \textbf{\color{magenta}{11}} & 12 & 13 & \textbf{\color{magenta}{14}} & \textbf{\color{magenta}{15}} \\
\midrule
olmo-1b & \textbf{0.65} & \textbf{0.63} & 0.34 & 0.32 & 0.08 & 0.40 & 0.43 & 0.28 & \textbf{0.65} & 0.47 & 0.50 & -0.18 & -0.14 & 0.07 & 0.44 & 0.59 \\
olmo-1b-sft & \textbf{0.80} & 0.47 & 0.31 & -0.10 & \textbf{0.86} & \textbf{0.80} & 0.52 & 0.49 & 0.24 & -0.32 & 0.28 & 0.57 & -0.17 & 0.07 & \textbf{0.62} & \textbf{0.80} \\
olmo-1b-dpo & \textbf{0.79} & 0.50 & 0.24 & -0.04 & \textbf{0.72} & \textbf{0.87} & 0.25 & 0.45 & 0.36 & -0.11 & 0.09 & \textbf{0.64} & 0.00 & -0.02 & 0.44 & \textbf{0.79} \\
\midrule
\multicolumn{17}{c}{\textbf{Adjectives: Comparative}} \\
\midrule
 Model & \textbf{\color{magenta}{0}} & \textbf{\color{magenta}{1}} & 2 & 3 & \textbf{\color{magenta}{4}} & \textbf{\color{magenta}{5}} & 6 & 7 & 8 & 9 & 10 & \textbf{\color{magenta}{11}} & 12 & \textbf{\color{magenta}{13}} & 14 & \textbf{\color{magenta}{15}} \\
\midrule
olmo-1b & 0.58 & \textbf{0.61} & 0.41 & 0.15 & 0.23 & \textbf{0.60} & 0.39 & 0.15 & 0.51 & 0.29 & 0.59 & 0.27 & 0.26 & \textbf{0.60} & 0.37 & \textbf{0.69} \\
olmo-1b-sft & \textbf{0.78} & 0.37 & 0.36 & -0.14 & \textbf{0.79} & 0.21 & 0.52 & 0.56 & 0.42 & 0.07 & 0.11 & \textbf{0.61} & 0.07 & 0.57 & 0.10 & \textbf{0.74} \\
olmo-1b-dpo & \textbf{0.77} & 0.42 & 0.28 & 0.13 & \textbf{0.71} & 0.42 & 0.49 & 0.59 & 0.51 & 0.26 & 0.27 & \textbf{0.61} & 0.22 & 0.10 & 0.02 & \textbf{0.84} \\
\bottomrule
\end{tabular}
}
\caption{Spearman $r$ correlations between $k$=1 vs $k$=2 settings for the \textsc{Adjectives} tasks. Layers containing heads with a strong positive correlation ($\geq$ 0.6) for at least one model are shown in color, and the corresponding values are bolded. The significantly active layers correspond roughly to the locations of IVs found in Section \ref{sec:identifying_ivs}.}
\label{tab:spearman-per-layer}
\end{table*}

{\bf Attention Head Activity.} Next, we investigate whether the information-processing paths elicited by different instructions differ structurally (i.e., different instructions = different circuits). We use \textit{attention head activity} along the high-ranking paths emitted from $\emT_\text{inst}$ as the proxy of instruction-specific circuit structures. Concretely, for a given task, we record the fraction of times a particular attention head participates in a high-ranking path. This measure indicates a structural summary of the causal mechanism post-instruction. 

We enumerate the paths as follows: For each query token, we take the top-$k$ most attended-to tokens, first with $k{=}1$, then $k{=}2$. Within each \textit{k} setting, we calculate a particular head's mean rate of activity over a sample size of $n=100$, along with bootstrapped confidence intervals with 10,000 resamplings. Additionally, we calculate the variance of each head's activity against the Gaussian distribution using a $t$-test, to determine whether the head's activity pattern is significantly different from noise. Figures \ref{fig:bootstrapped-app} - \ref{fig:bootstrapped-8} in Appendix \ref{app:sec:attention_head_activity_stat_tests} provide a visualization.

Across all models, we find that most heads have significant non-noise activity. The remaining heads are mainly deterministic (always on or off); noise heads are very rare. With increasing $k$, more significant heads appear, and head activity increases overall. This is to be expected, since more paths are added to the computation graph.

Despite the increase in total head activity, we notice a pattern that stays robust across $k$: namely, that earlier layers have similar active heads, while later ones differ. We quantify this by computing the Jaccard similarity of the numerical matrices corresponding to the heatmaps, between both $k$ settings (shown in Table \ref{tab:jaccard-similarity}).

Figure \ref{fig:attn-heads-analysis} provides a visualization of these patterns. For \textsc{Adj: Comparative}, \textsc{Adj: Antonym}, and \textsc{Anim: Color}, the head activity pattern remains nearly identical in the earlier layers (particularly 0-1), then begins to differ in the later layers between the tasks. Interestingly, we observe that for \textsc{anim: can\_fly}, the attention head activity patterns at $\emT^{\text{fly}}_\text{11}$ most closely resemble those for \textsc{anim: color} at $\emT^{\text{color}}_\text{10} = \emT^{\text{color}}_\text{inst}$. Meanwhile, the head activity patterns for $\emT^{\text{fly}}_\text{19} = \emT^{\text{fly}}_\text{inst}$ are markedly different from both of these token positions. Correspondingly, $\emT^{\text{fly}}_\text{10}$ was the token position with the most high-ranking paths for \textsc{anim: can\_fly}.

\begin{table}[h]
\centering
\resizebox{\linewidth}{!}{
\begin{tabular}{l ccc ccc}
\toprule
& \multicolumn{3}{c}{\texttt{adj\_comp}} & \multicolumn{3}{c}{\texttt{adj\_ant}} \\
\cmidrule(lr){2-4} \cmidrule(lr){5-7}
Model & L0--1 & L4--5 & L14--15 & L0--1 & L4--5 & L14--15 \\
\midrule
OLMo-1B         & \textbf{0.91} & 0.25 & 0.24 & \textbf{0.88} & 0.18 & 0.38 \\
OLMo-1B-SFT     & \textbf{0.84} & 0.26 & 0.48 & \textbf{0.81} & 0.19 & 0.43 \\
OLMo-1B-DPO     & \textbf{0.84} & 0.26 & 0.35 & \textbf{0.81} & 0.27 & 0.36 \\
\bottomrule
\end{tabular}
}
\caption{Jaccard similarity of active heads (threshold $> 0.1$) between $k{=}1$ and $k{=}2$, by layer group at $\emT_\text{inst}$. Generalization to $k{=}3$ is shown in Appendix \ref{app:subsec:jaccard-tables}.}
\label{tab:jaccard-similarity}
\end{table}

\paragraph{Active Heads and Instruction Vectors.} The observed head activity patterns have important connections to instruction vectors. Namely, we notice that the layers which contain significantly active heads (Table \ref{tab:spearman-per-layer}) for a particular model and task are also the layers that are involved in most of the 2- or 3-layer tuples that constitute instructional representations, presented in Section \ref{sec:identifying_ivs}.

For example, in Table \ref{tab:contrastive-tasks-rank-1b}, we see that most top layer triples for the post-trained models in the \textsc{Adjective: Antonym} task involve Layers 14 or 15, which are strongly active across $k$. Meanwhile, for the base model, most triples involve Layers 0 and 1. Additionally, while inactive layers such as 14 and 15 are sometimes involved in the top triples, they tend to appear alongside a strongly-active layer.

A similar trend holds for \textsc{Adjective: Comparative}. Here, Layer 11 appears most often in the top layer triples for SFT and DPO, while it appears only 3 times for the base model, and always in combination with the more active Layer 15. We do note minority cases where top-contributing tuples consist solely of weakly active heads -- however, these tuples tend to be tied with the same score as tuples containing highly active heads. This implies that further multi-layer interaction might be occurring.

Overall, we observe that \textit{instruction vectors corresponding to different tasks are constructed in the early layers} using nearly identical, instruction-agnostic circuits (similar head activity), \textit{followed by task-specific circuits that are conditioned by the instruction vector} (different head activity).

\section{Conclusion}
\label{conclusion}

In this work, we examine the internals of language models to gain insight into the mechanisms of instruction following, comparing the base model with its post-trained counterparts. We causally locate neural digests of task representation, which are formed in an eager manner right after the instruction is processed. While these digests are linearly separable according to task semantics, they interact with each other in a nonlinear, synergistic manner. This finding provides counter-evidence against the assumption of `one component, one causal role', commonplace in the current landscape of mechanistic interpretability \cite{wang2023interpretability}. Subsequently, this makes computational graphs unreliable for identifying the causal mechanisms of instruction processing. To mitigate this challenge, we propose a novel method of tracing information flow within Transformers that is free from additive assumptions, allowing us to pinpoint the importance of each model component in processing an instruction. Using this approach, we identify the role of instruction vectors as circuit selectors.

\section*{Limitations}

Due to memory constraints, our experiments regarding the causal mechanism are conducted on 1B models. However, these models may be limited in their general task-solving capabilities compared to 7B models. While the general trend of our conclusions remains the same, more elaborate patterns may emerge at larger model scales. Additionally, while our results regarding the base, SFT, and DPO checkpoints provides initial insight into how instruction representations evolve during the course of model training, a finer-grained analysis on a greater number of intermediate checkpoints (e.g. with Pythia models \cite{pmlr-v202-biderman23a}) could be conducted.

Our definition of contrastive task pairs is based upon human-defined distinctions - i.e. we give differing instructions while keeping the target query the same. Nevertheless, this does not guarantee that an LLM will also perceive these tasks to be different. An analysis could be conducted that involves contrastive instructions as determined by an LLM's notion of task similarity, e.g. defined using distance metrics between individual sample representations. However, conducting such an analysis is beyond the scope of this work.

Similarly, we choose instructions which we consider to be fairly simple, both in terms of their linguistic structure as well as in the underlying task that they represent. Studying the impact of task difficulty and instructional complexity is worthwhile, but is beyond the scope of this work due to the significant theoretical and experimental foundation required (i.e. providing a satisfactory definition of \enquote{difficulty} and conducting a thorough ablation).

Finally, the instructions that we examine in this work all have a clear syntactic separation between the instruction and query. This is purposefully done to make it possible to investigate the point in the prompt when the instruction is fully processed, but the query hasn't yet been. This separation is also important for our path analysis experiments, since the case where the query can be cleanly switched out across samples without altering the instruction results in the fewest confounding factors. However, we acknowledge that real-world instructions are often intermixed, and may yield additional interesting insights.

We acknowledge these limitations with the intention that our work may build a starting foundation for future research in this area. 

\section*{Acknowledgements}
This work was funded by the LOEWE Distinguished Chair “Ubiquitous Knowledge Processing”, LOEWE initiative, Hesse, Germany (Grant Number: LOEWE/4a//519/05/00.002(0002)/81), as well as by the German Federal Ministry of Education and Research and the Hessian Ministry of Higher Education, Research, Science and the Arts within their joint support of the National Research Center for Applied Cybersecurity ATHENE.

We would like to thank Prof. Kentaro Inui and Phu Hoang for the insightful and helpful discussions regarding an early version of this paper.

\bibliography{custom}
\appendix

\section{Hardware Details and Runtimes}
\label{app:sec:experiment-runtimes}
In this section, we provide information regarding the computational resources used as well as the experiment runtimes. 

We run our activation patching experiments (Section \ref{sec:identifying_ivs}) and our path analysis experiments (Section \ref{sec:causal_experiments}) on H100 and A100 GPUs, with RAM capacities of either 40GB or 80GB. We use non-quantized models. For experiments which were not as resource-intensive, such as the linear probe (Section \ref{subsec:spatial_analysis}) and the inference experiments (Section \ref{sec:inference-experiments}), we use an L40 48GB GPU.

The average runtimes for our main experiments are listed in Table \ref{tab:app:experiment-runtimes}.

\begin{table}
    \centering
    \resizebox{\linewidth}{!}{
    \begin{tabular}{lcccc}
    \toprule
        Experiment (1 Task) & Num. Samples & OLMo-2 1B & OLMo-2 7B \\
        \midrule
         Activation Patching & 100 & 4 hrs. & 24 hrs. \\
         Path Analysis ($k{=}1) $ & 100 & 6 hrs. & - \\
         Path Analysis ($k{=}2) $ & 20 & 36 hrs. & - \\
         Path Analysis ($k{=}3) $ & 15 & 40 hrs. & - \\
         \bottomrule
    \end{tabular}
    }
    \caption{Average runtimes for 1 task in each experiment phase. We observe that the SFT and DPO checkpoints of each model typically have longer runtimes than the base model.}
    \label{tab:app:experiment-runtimes}
\end{table}

\section{Full Activation Patching Results}
\label{app:sec:numerical_scores}

\begin{figure*}
    \centering
    \includegraphics[width=\linewidth,scale=0.7,trim={0 14cm 0cm 0cm},clip]{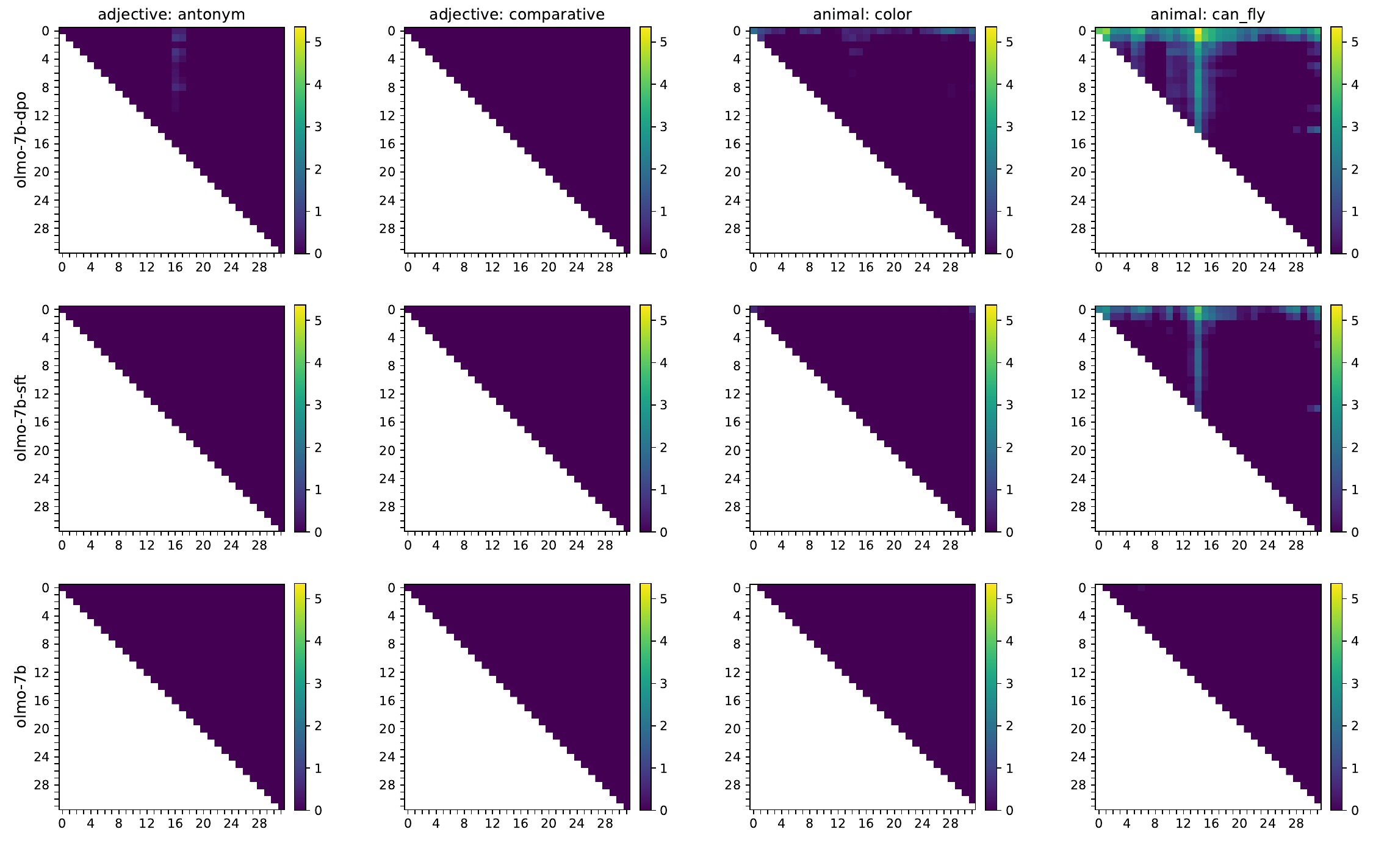}

    \caption{Effects of 1- and 2-layer patching configurations on the logit of the target token for OLMo-2 7B models. Each square of the x- and y-coordinate grid represents the corresponding layers of the model being patched. Coordinates where x=y represent 1-layer patching.}
    \label{fig:app:1LP+2LP_logit_olmo27b_olmo3}
\end{figure*}

\begin{figure*}
    \centering
    \includegraphics[width=\linewidth,scale=0.7,trim={0 14cm 0cm 0cm},clip]{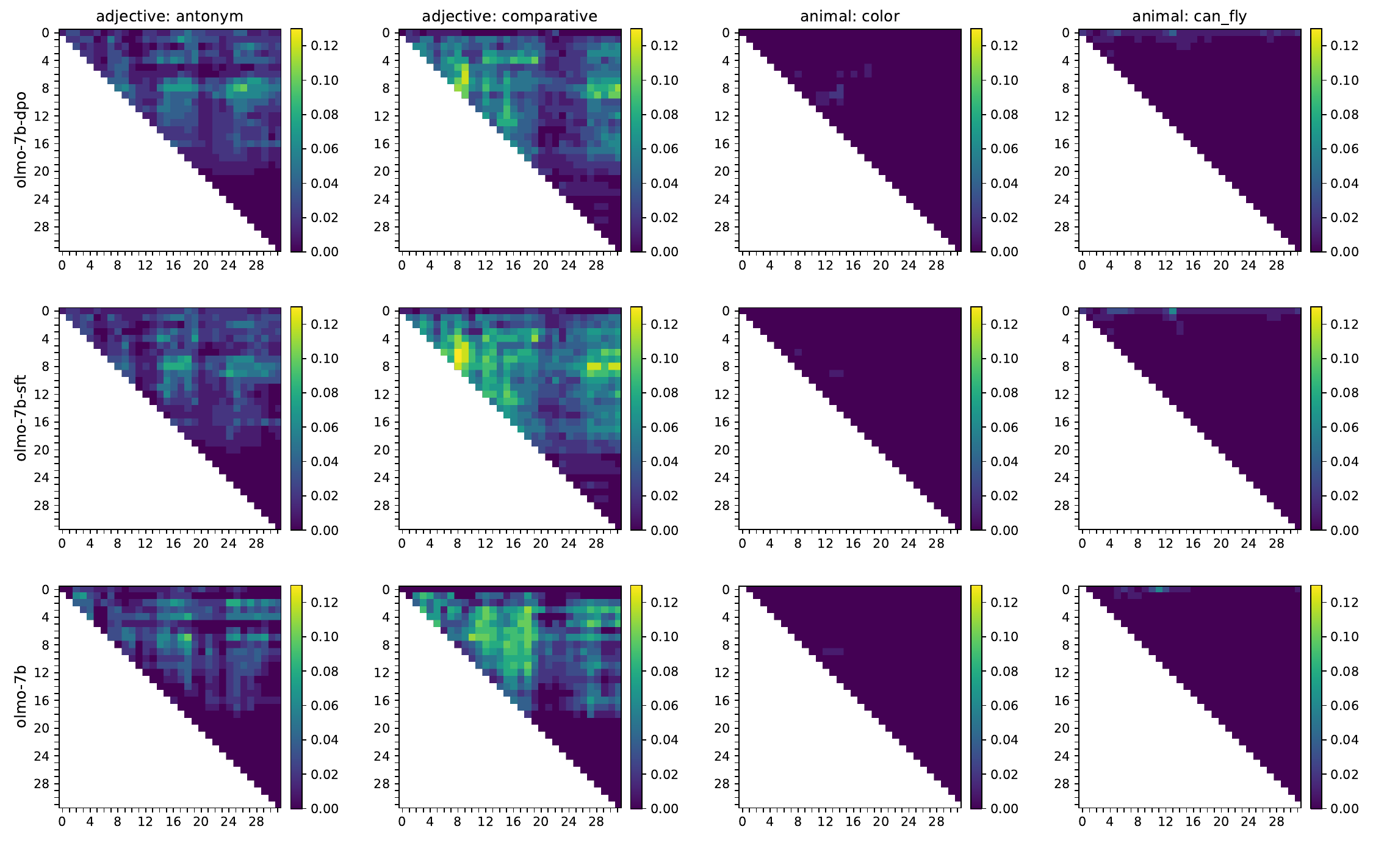}
    
    \caption{Effects of 1- and 2-layer patching configurations on the rank of the target token for OLMo-2 7B models. Each square of the x- and y-coordinate grid represents the corresponding layers of the model being patched. Coordinates where x=y represent 1-layer patching.}
    \label{fig:app:1LP+2LP_rank_olmo27b_olmo3}
\end{figure*}

\begin{figure*}
    \centering
    \includegraphics[width=\linewidth,scale=0.7,trim={0 14cm 0cm 0cm},clip]{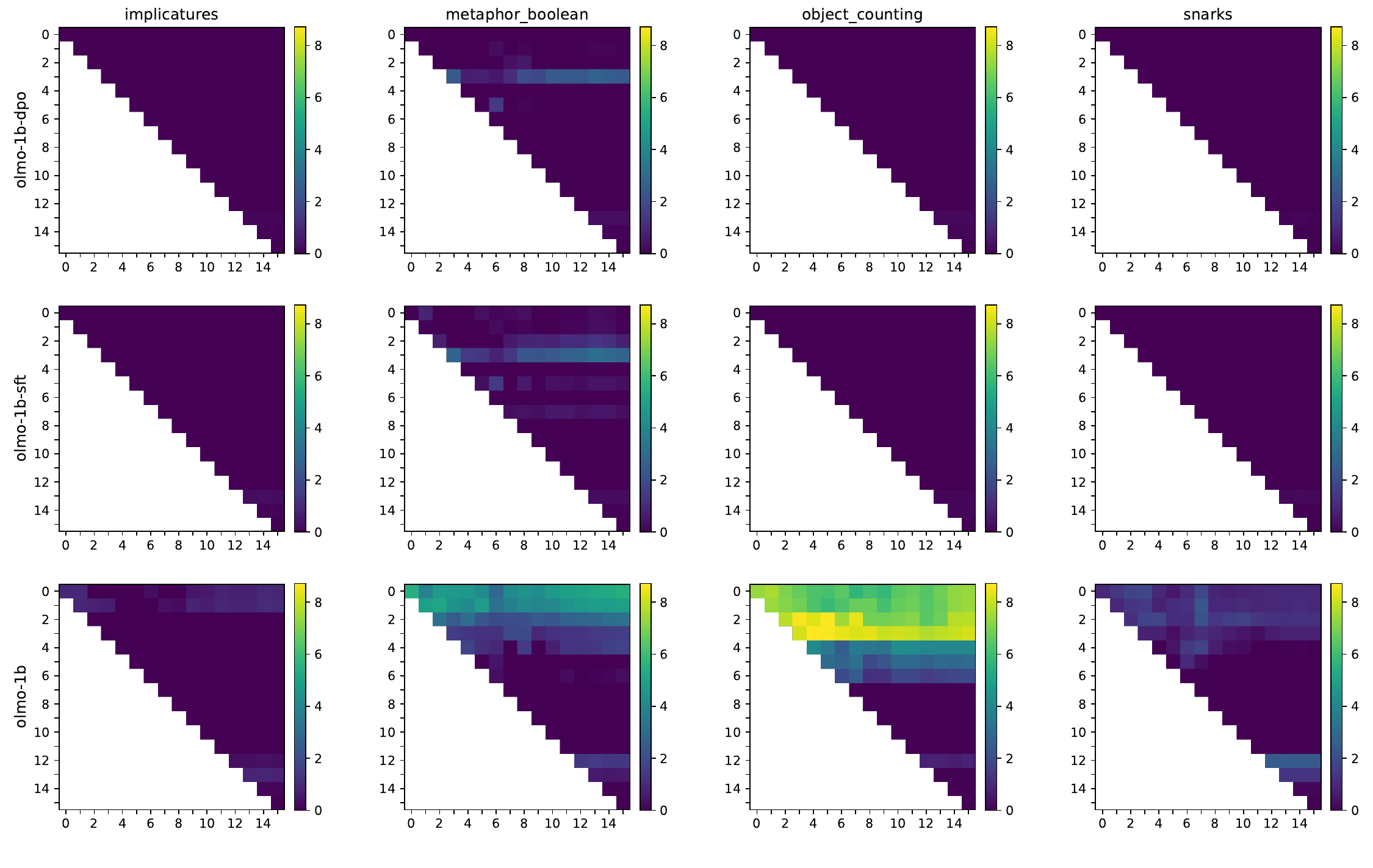}

\includegraphics[width=\linewidth,scale=0.7,trim={0.5cm 13cm 0.6cm 5cm},clip]{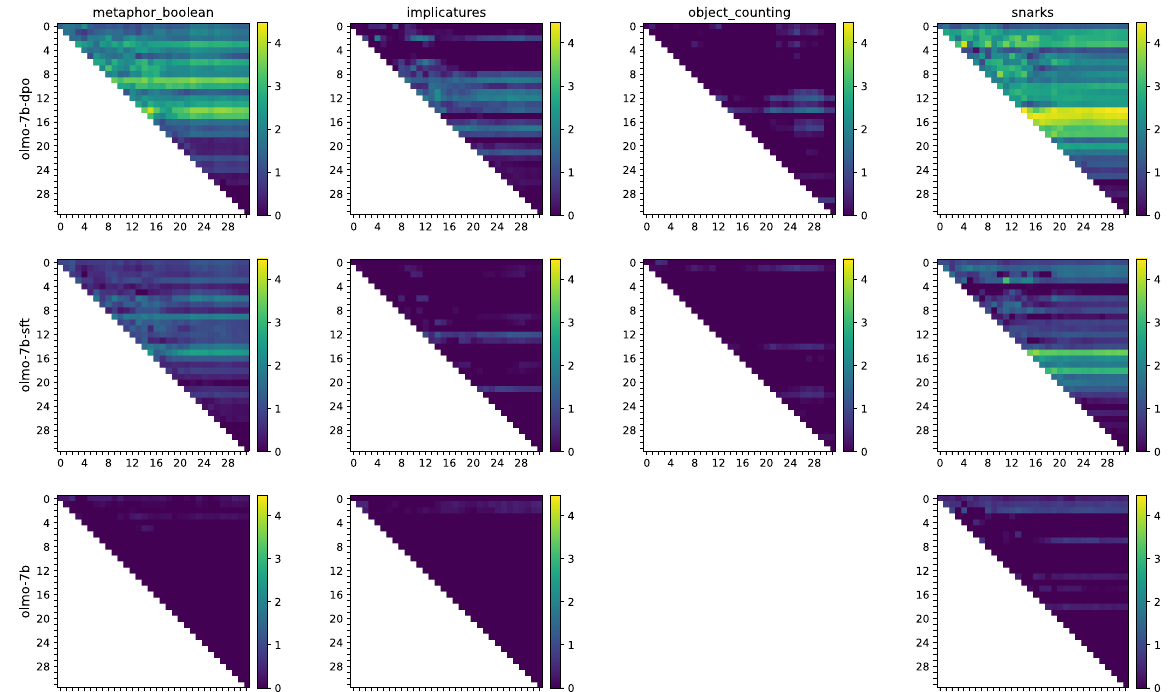}

    \caption{Effects of 1- and 2-layer patching configurations on the logit of the target token for BigBench tasks. Each square of the x- and y-coordinate grid represents the corresponding layers of the model being patched. Coordinates where x=y represent 1-layer patching. \textsc{object\_counting} scores for OLMo-2 7B are missing due to a model error while running the task.}
    \label{fig:app:bb-norm-logits}
\end{figure*}

\begin{figure*}
    \centering
    \includegraphics[width=\linewidth,scale=0.7,trim={0 14cm 0cm 0cm},clip]{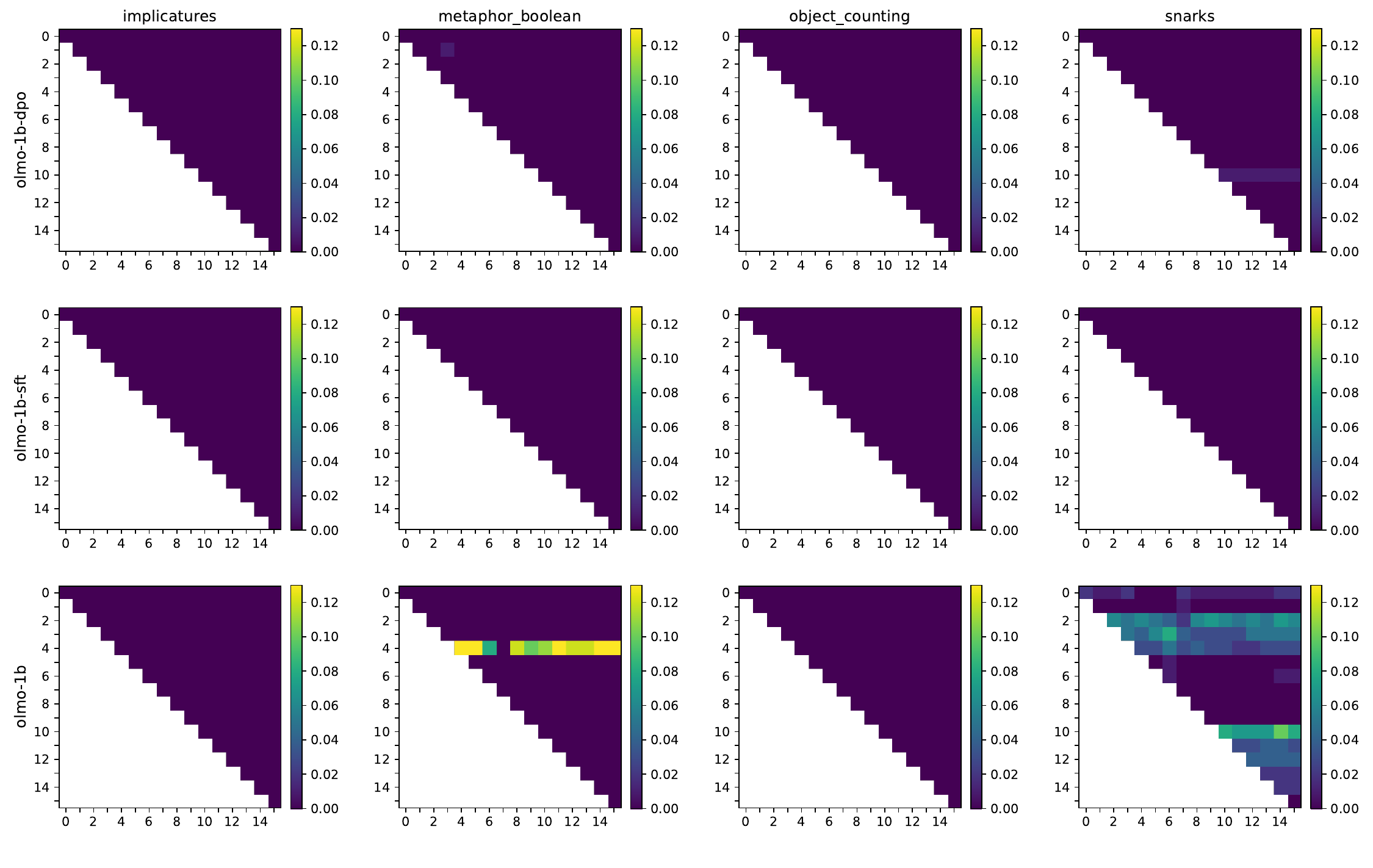}

\includegraphics[width=\linewidth,scale=0.7,trim={0.5cm 0cm 0cm 0cm},clip]{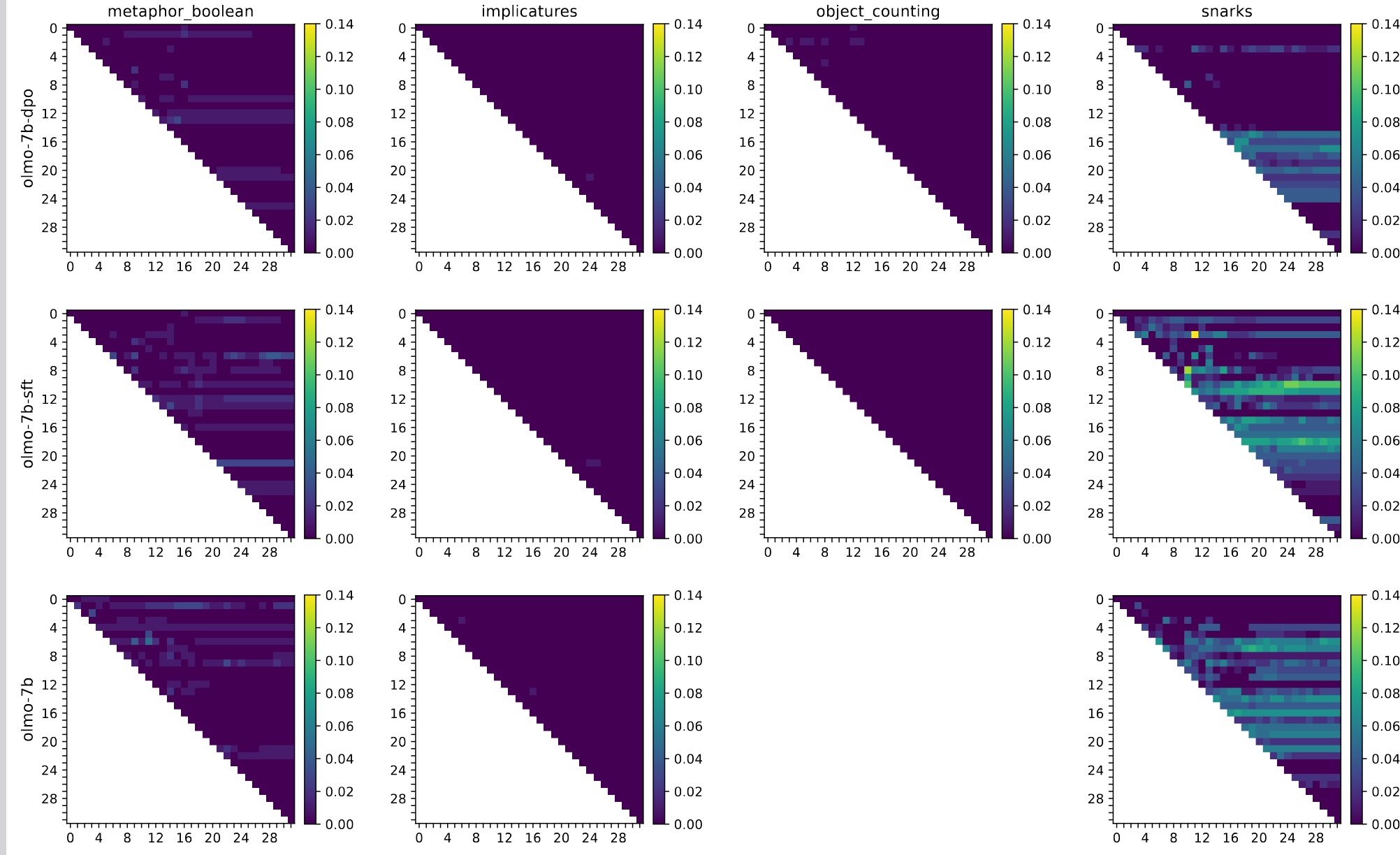}

    \caption{Effects of 1- and 2-layer patching configurations on the rank of the target token for BigBench tasks. Each square of the x- and y-coordinate grid represents the corresponding layers of the model being patched. Coordinates where x=y represent 1-layer patching. \textsc{object\_counting} scores for OLMo-2 7B are missing due to a model error while running the task.}
    \label{fig:app:bb-norm-rank}
\end{figure*}

\begin{figure*}[hbt!]
    \centering

    \includegraphics[width=\linewidth,scale=0.1,trim={0 15cm 0cm 0cm},clip]{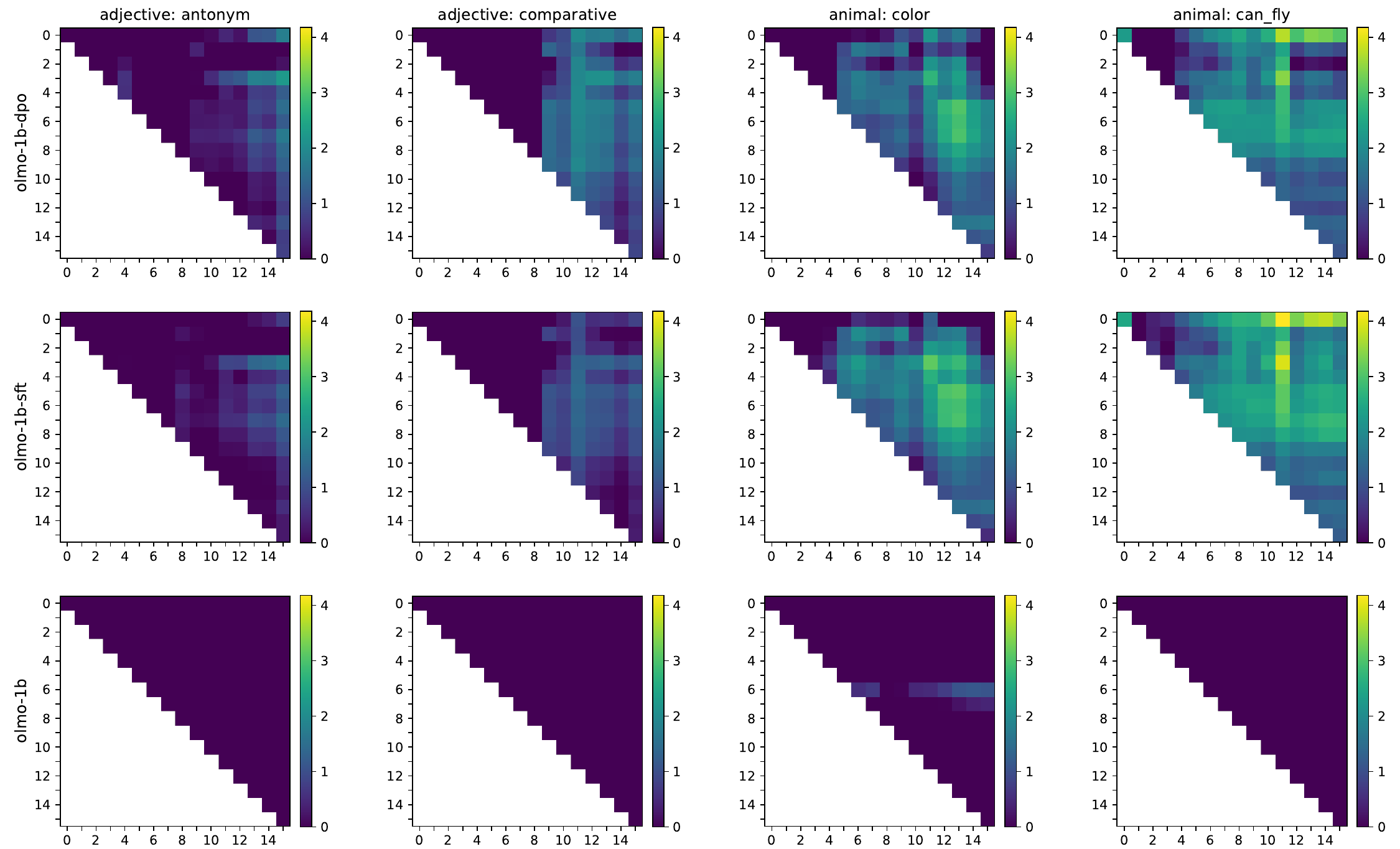}

    \caption{Effects of 1- and 2-layer patching configurations on the logit of the target token. Each square of the x- and y-coordinate grid represents the corresponding layers of the model being patched. Coordinates where x=y represent 1-layer patching. Two important properties are shown. \textit{Localization:} Across all tasks, we observe localized points where the logit improvement is the greatest. \textit{Superadditivity:} Two-layer combinations bring greater improvements than single layers.}
    \label{fig:norm_logit_pt-vs-to_1b}
\end{figure*}

In this section, we present the activation patching heatmaps for all models (Figures \ref{fig:app:1LP+2LP_logit_olmo27b_olmo3} - \ref{fig:1LP+2LP_logit}), as well as the numerical logit and rank contributions (Tables \ref{tab:app:contrastive-rank-olmo1b-num} - \ref{tab:app:contrastive-logit-olmo7b-num}). For each model, we display the results of the 10 highest-contributing tokens.

\begin{figure*}
    \centering
    \includegraphics[width=\linewidth,scale=0.7,trim={0 15cm 0cm 0cm},clip]{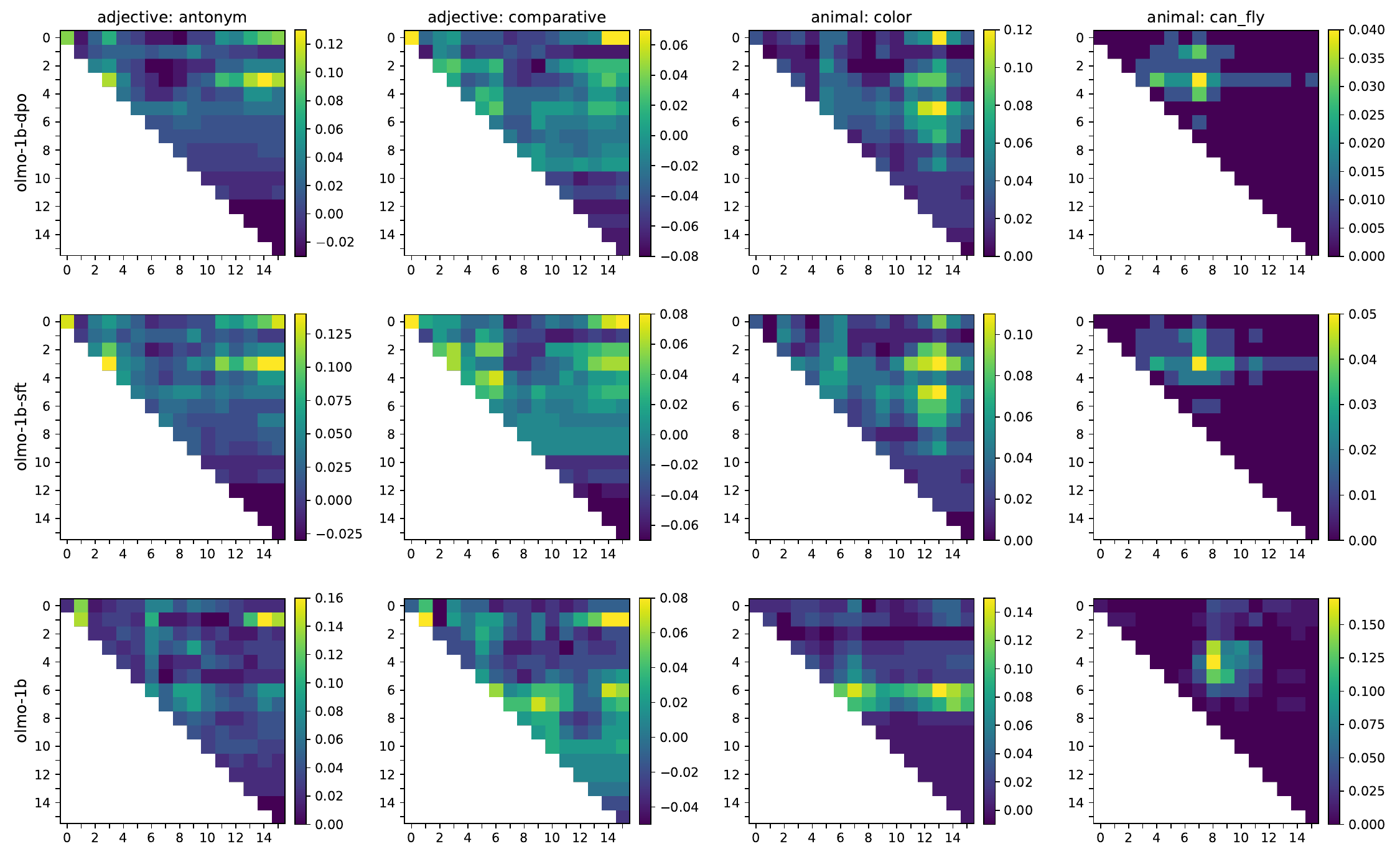}
    \caption{Effects of 1- and 2-layer patching configurations on the reciprocal rank of the target token for OLMo-2 1B models, without normalization across panels. Each square of the x- and y-coordinate grid represents the corresponding layers of the model being patched. Coordinates where x=y represent 1-layer patching.}
    \label{fig:unnorm_1b/unnorm_contr_ranks}
\end{figure*}

\begin{figure*}
    \centering
    \includegraphics[width=\linewidth,scale=0.7,trim={0 15cm 0cm 0cm},clip]{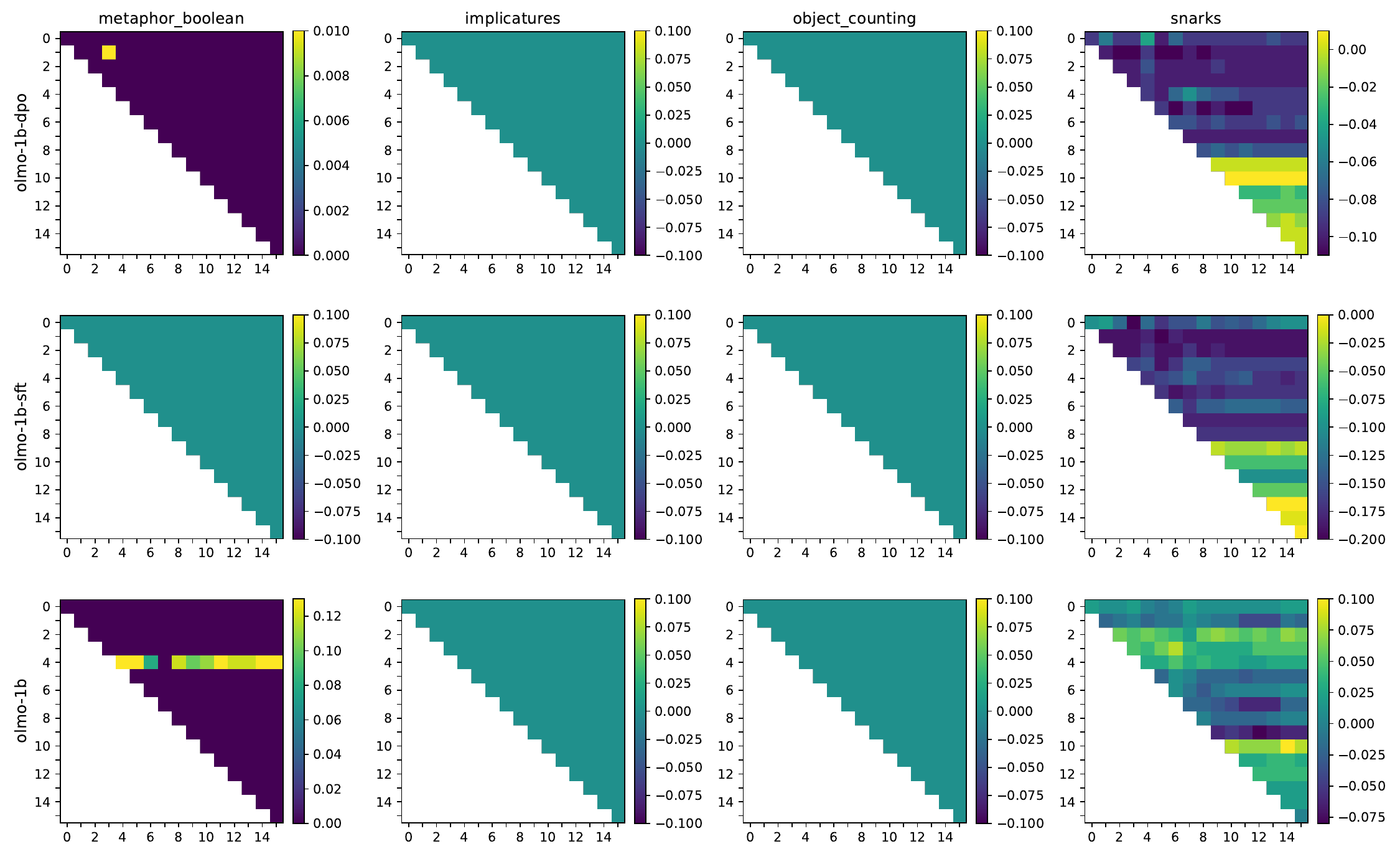}
    \caption{Effects of 1- and 2-layer patching configurations on the reciprocal rank of the target token for OLMo-2 1B models, without normalization across panels. Each square of the x- and y-coordinate grid represents the corresponding layers of the model being patched. Coordinates where x=y represent 1-layer patching.}
    \label{fig:unnorm_1b/unnorm_bb_ranks}
\end{figure*}

\begin{figure*}
    \centering
    \includegraphics[width=\linewidth,scale=0.7,trim={0 15cm 0cm 0cm},clip]{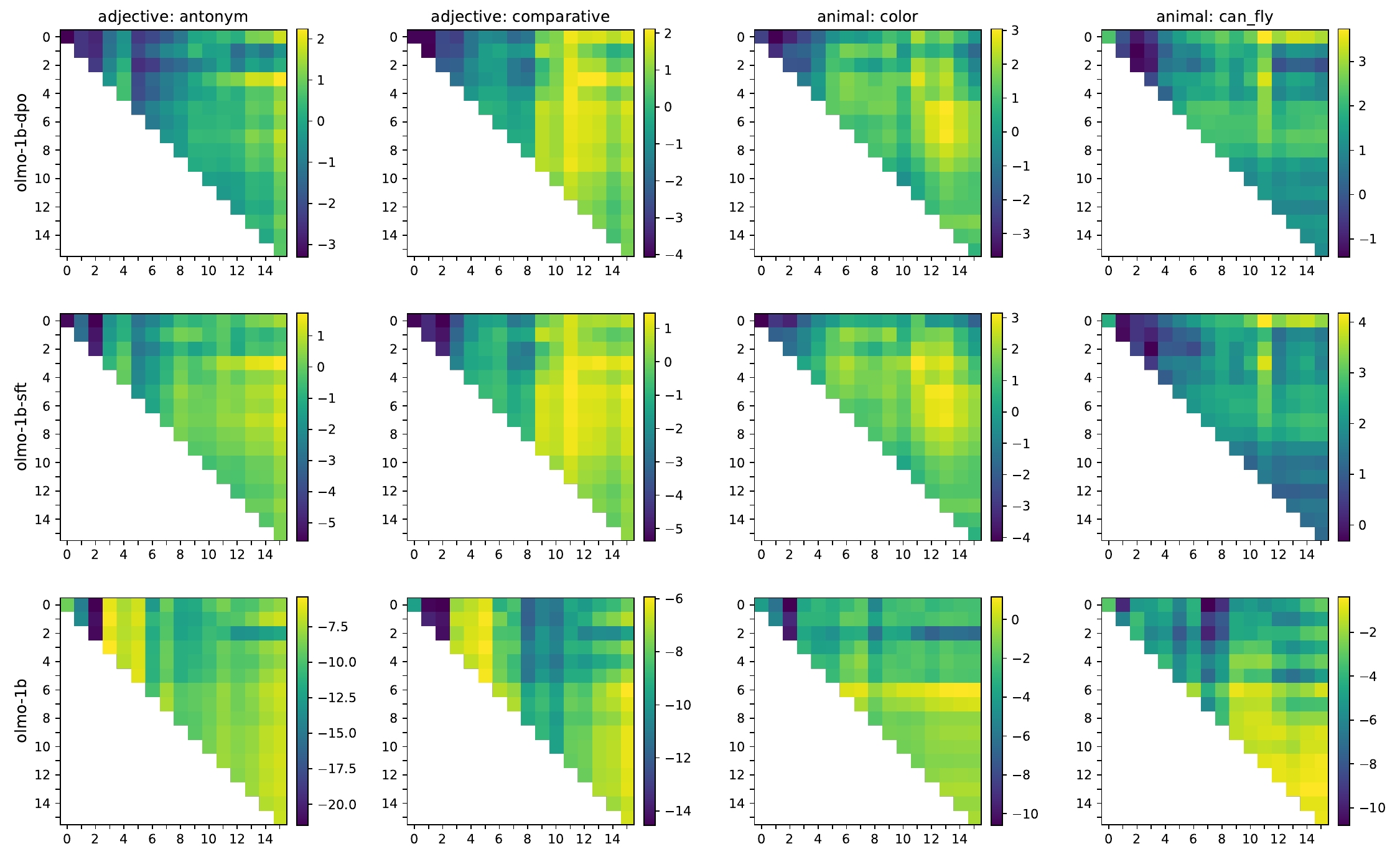}
    \caption{Effects of 1- and 2-layer patching configurations on the logit of the target token for OLMo-2 1B models, without normalization across panels. Each square of the x- and y-coordinate grid represents the corresponding layers of the model being patched. Coordinates where x=y represent 1-layer patching.}
    \label{fig:unnorm_1b/unnorm_contr_logits}
\end{figure*}

\begin{figure*}
    \centering
    \includegraphics[width=\linewidth,scale=0.7,trim={0 15cm 0cm 0cm},clip]{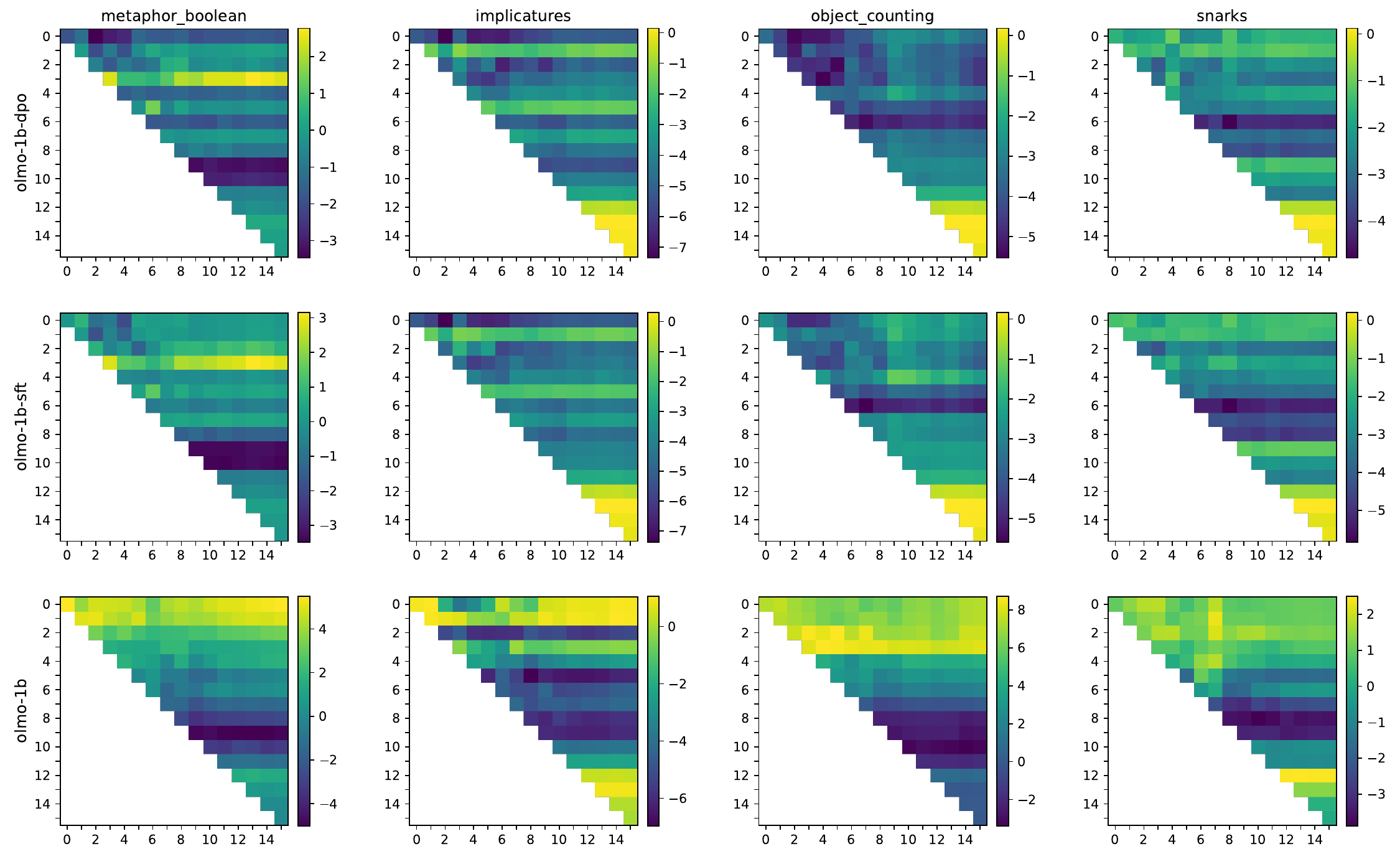}
    \caption{Effects of 1- and 2-layer patching configurations on the logit of the target token for OLMo-2 1B models, without normalization across panels. Each square of the x- and y-coordinate grid represents the corresponding layers of the model being patched. Coordinates where x=y represent 1-layer patching.}
    \label{fig:unnorm_1b/unnorm_bb_logits}
\end{figure*}

\begin{figure*}
    \centering
    \includegraphics[width=\linewidth,scale=0.7,trim={0 15cm 0cm 0cm},clip]{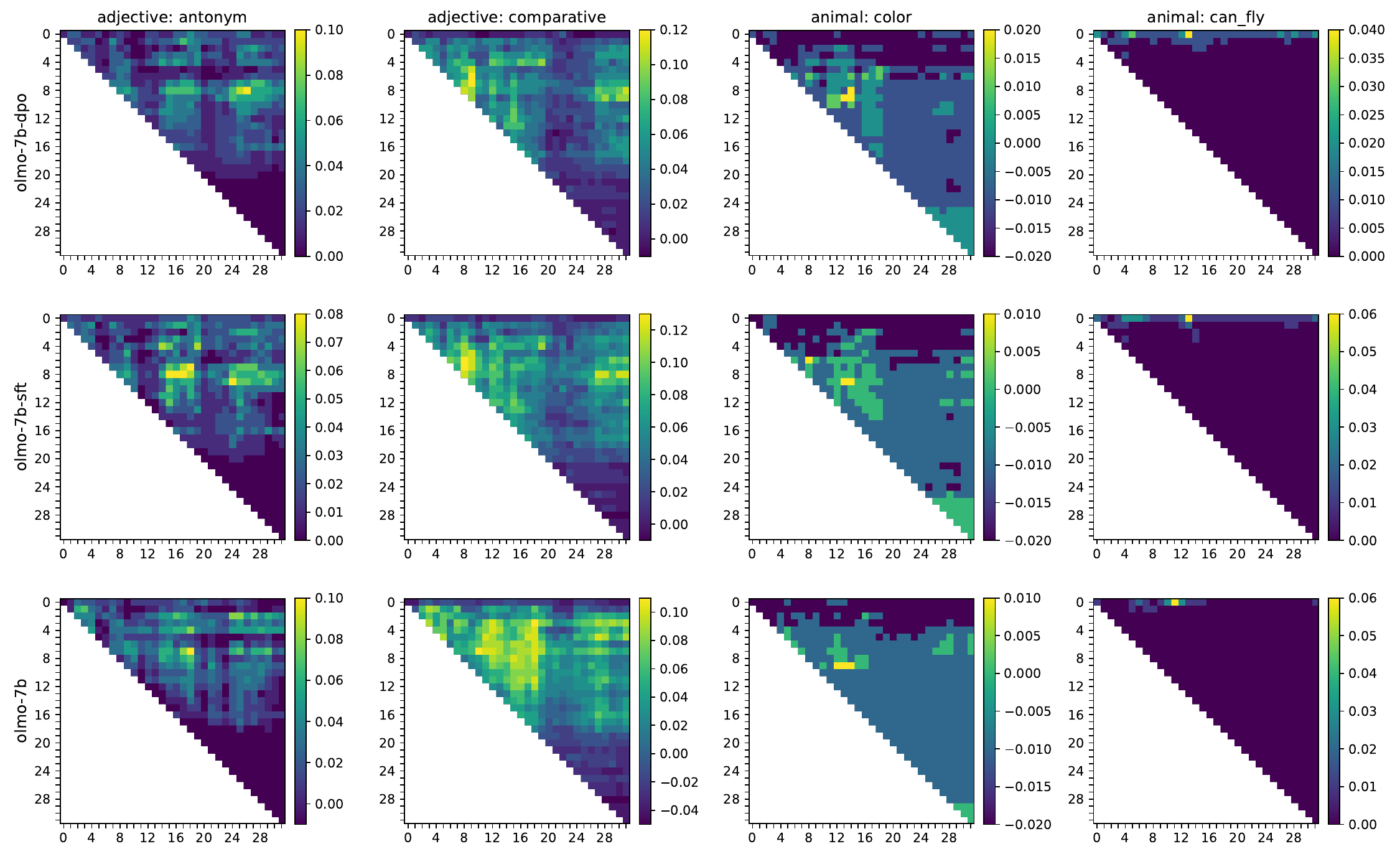}
    \caption{Effects of 1- and 2-layer patching configurations on the reciprocal rank of the target token for OLMo-2 7B models, without normalization across panels. Each square of the x- and y-coordinate grid represents the corresponding layers of the model being patched. Coordinates where x=y represent 1-layer patching.}
    \label{fig:unnorm_7b/unnorm_contr_ranks}
\end{figure*}

\begin{figure*}
    \centering
    \includegraphics[width=\linewidth,scale=0.7,trim={0 15cm 0cm 0cm},clip]{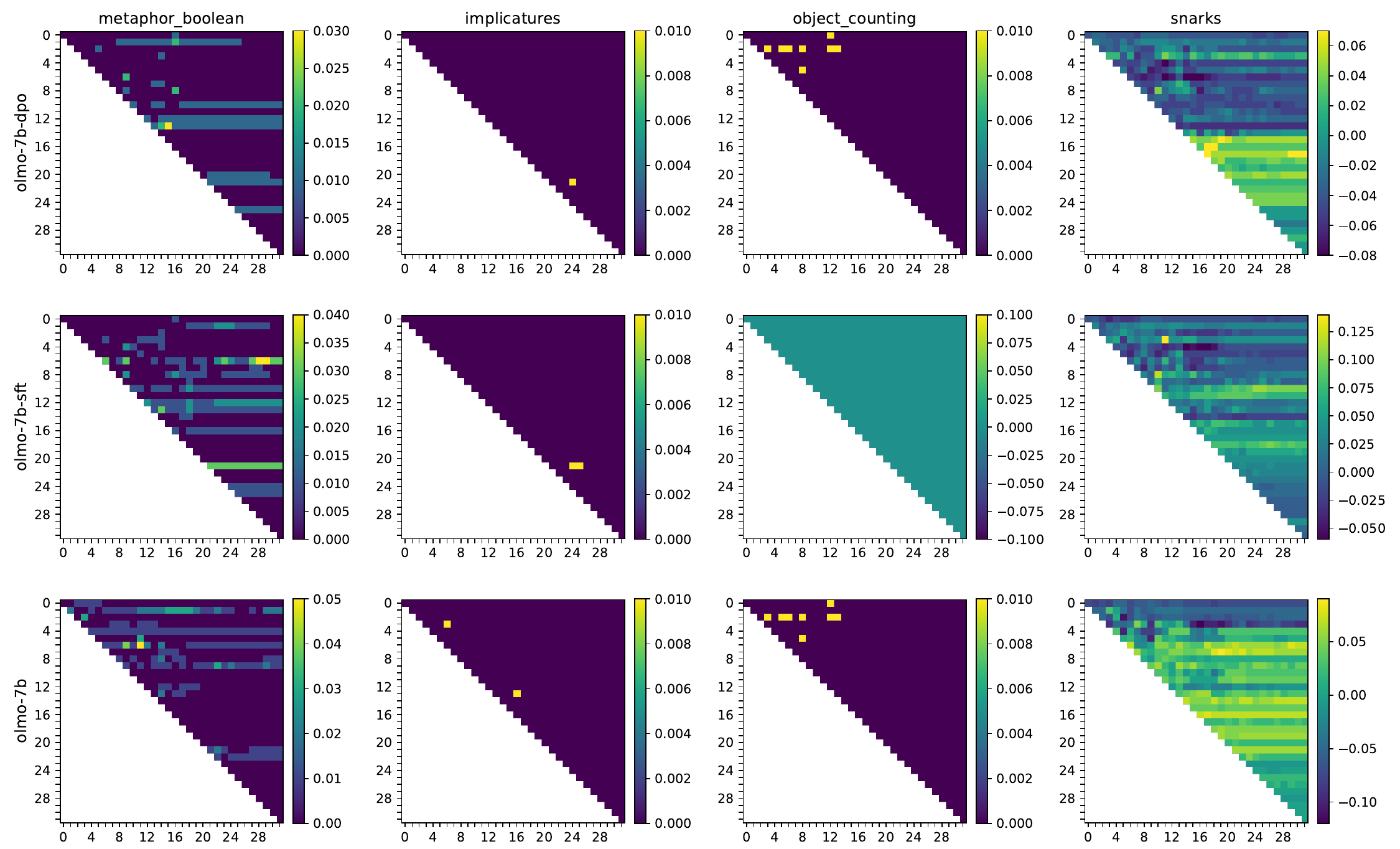}
    \caption{Effects of 1- and 2-layer patching configurations on the reciprocal rank of the target token for OLMo-2 7B models, without normalization across panels. Each square of the x- and y-coordinate grid represents the corresponding layers of the model being patched. Coordinates where x=y represent 1-layer patching.}
    \label{fig:unnorm_7b/unnorm_bb_ranks}
\end{figure*}

\begin{figure*}
    \centering
    \includegraphics[width=\linewidth,scale=0.7,trim={0 15cm 0cm 0cm},clip]{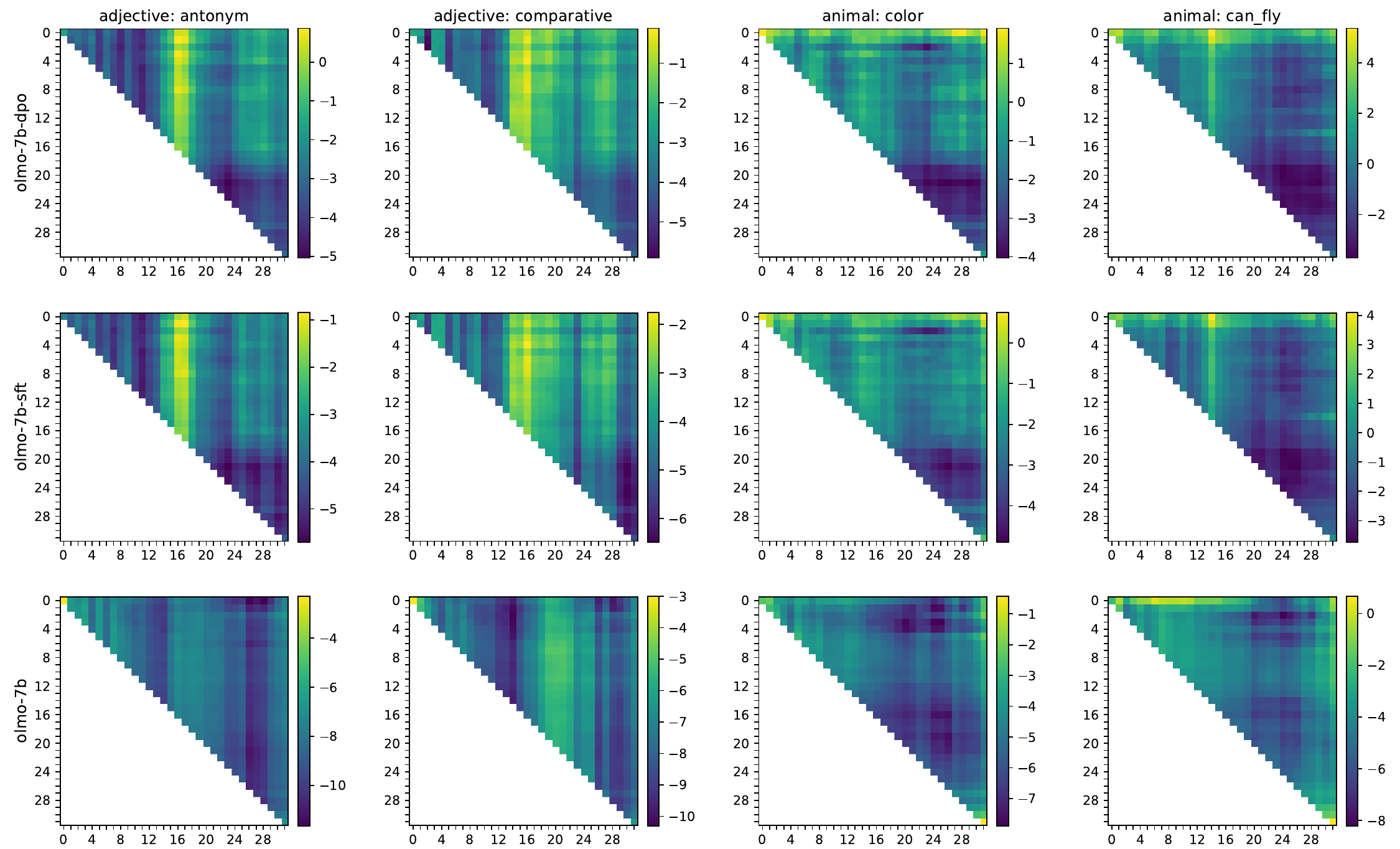}
    \caption{Effects of 1- and 2-layer patching configurations on the logit of the target token for OLMo-2 7B models, without normalization across panels. Each square of the x- and y-coordinate grid represents the corresponding layers of the model being patched. Coordinates where x=y represent 1-layer patching.}
    \label{fig:unnorm_7b/unnorm_contr_logits}
\end{figure*}

\begin{figure*}
    \centering
    \includegraphics[width=\linewidth,scale=0.7,trim={0 15cm 0cm 0cm},clip]{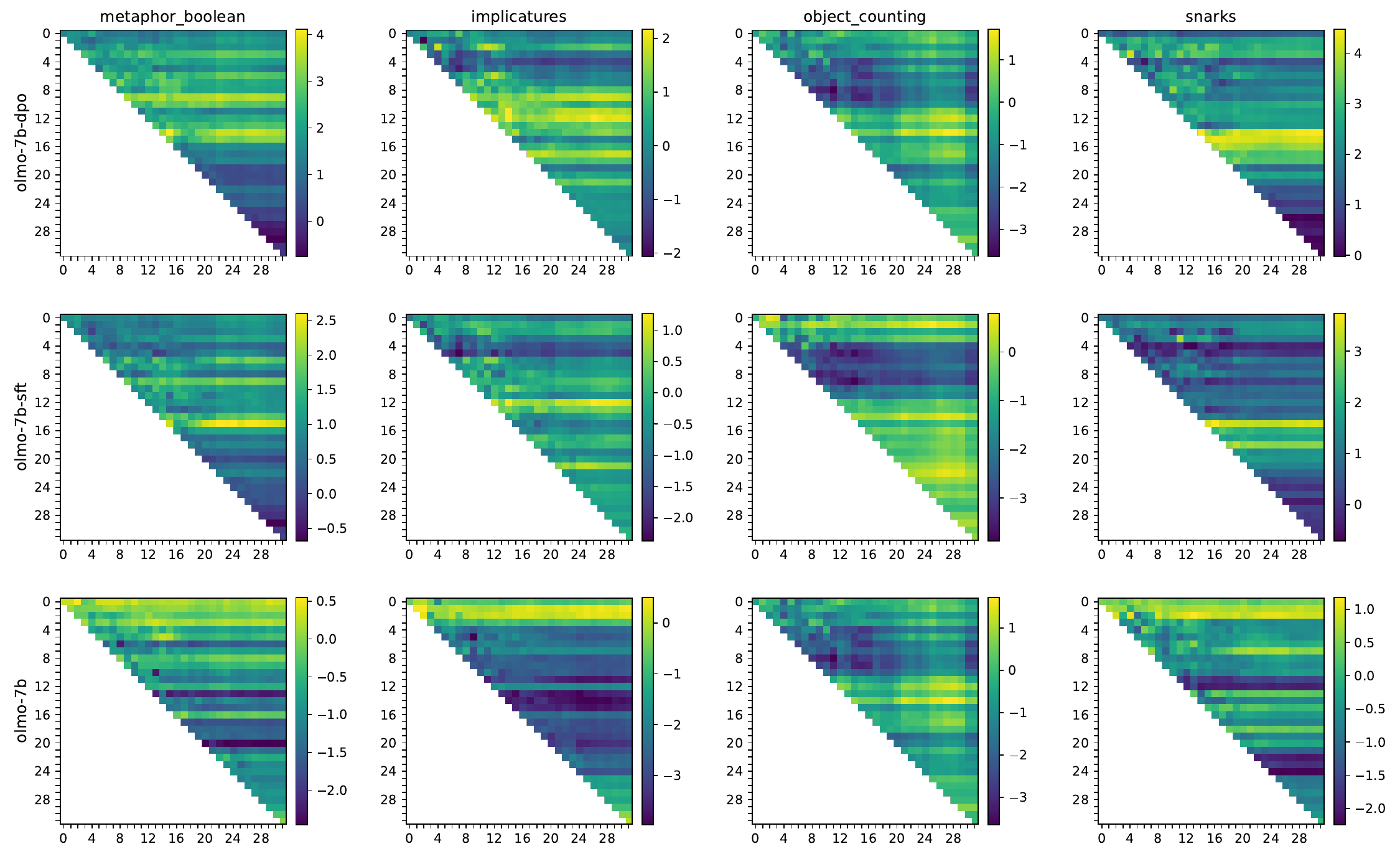}
    \caption{Effects of 1- and 2-layer patching configurations on the logit of the target token for OLMo-2 7B models, without normalization across panels. Each square of the x- and y-coordinate grid represents the corresponding layers of the model being patched. Coordinates where x=y represent 1-layer patching.}
    \label{fig:unnorm_7b/unnorm_bb_logits}
\end{figure*}

As a supplement, in Figures \ref{fig:unnorm_1b/unnorm_contr_ranks} - \ref{fig:unnorm_7b/unnorm_bb_logits}, we additionally present the activation patching heatmaps in their \textit{unnormalized} form - that is, without normalizing against the mininum and maximum scores across all panels in the diagram. While the underlying data is the same as in Figures \ref{fig:app:1LP+2LP_logit_olmo27b_olmo3} - \ref{fig:1LP+2LP_logit}, this alternate view makes it easier to see the areas of localization in each model.

Finally, we present the $t$-test results conducted on OLMo 1B and OLMo 7B models (Tables \ref{tab:app:ttest-olmo1b-contrastive} - \ref{tab:app:ttest-olmo7b-bigbench}). For each task, we pick the top-10 patching combinations in terms of average rank contribution (as shown in the heatmaps, as well as the numerical score tables). Then, for each of these layer combinations, we take their unnormalized patching scores and evaluate Eq. \ref{eq:superadditivity_inequality} (as a boolean truth value 1 or 0).

 \begin{table*}
     \centering
     \resizebox{\textwidth}{!}{
     \begin{tabular}{cccccccccccc}

    \toprule
     \multicolumn{12}{c}{\textbf{Contrastive Tasks: T-Test}} \\
    \toprule
     \multicolumn{12}{c}{\textbf{OLMo-2 1B SFT}} \\
     \midrule
     \multicolumn{3}{c}{adj: antonym} & \multicolumn{3}{c}{adj: comparative} & \multicolumn{3}{c}{anim: color} & \multicolumn{3}{c}{anim: can\_fly}\\
     \midrule
     Layer(s) &  t-stat.  &  p-value & Layer(s) &  t-stat. & p-value &  Layer(s) & t-stat. &  p-value &  Layer(s) & t-stat.  & p-value  \\ 
     \midrule
(3, 3) & -1.003e+03 & 3.487e-200 & (0, 0) & -1.132e+03 & 2.327e-205 & (3, 13) & -1.334e+03 & 1.941e-212 & (3, 7) & -1.012e+03 & 1.483e-200  \\
(3, 14) & -1.504e+03 & 1.377e-217 & (0, 15) & -inf & 0.000e+00 & (5, 13) & -9.890e+02 & 1.447e-199 & (1, 7) & -9.862e+02 & 1.911e-199  \\
(3, 15) & -2.239e+03 & 1.061e-234 & (0, 14) & -2.490e+03 & 2.901e-239 & (3, 12) & -1.504e+03 & 1.377e-217 & (2, 7) & -1.036e+03 & 1.498e-201  \\
(0, 0) & -1.182e+03 & 3.077e-207 & (4, 6) & -9.963e+02 & 6.999e-200 & (5, 12) & -1.081e+03 & 2.208e-203 & (3, 4) & -9.934e+02 & 9.296e-200  \\
(0, 15) & -inf & 0.000e+00 & (2, 3) & -1.117e+03 & 8.176e-205 & (0, 13) & -1.562e+03 & 3.312e-219 & (3, 8) & -9.890e+02 & 1.446e-199  \\
(3, 11) & -1.092e+03 & 7.928e-204 & (3, 3) & -1.036e+03 & 1.412e-201 & (2, 13) & -2.055e+03 & 5.115e-231 & (3, 9) & -9.874e+02 & 1.696e-199  \\
(3, 13) & -1.081e+03 & 2.208e-203 & (3, 14) & -2.239e+03 & 1.061e-234 & (3, 11) & -1.504e+03 & 1.377e-217 & (1, 6) & -1.007e+03 & 2.337e-200  \\
(0, 14) & -1.409e+03 & 8.728e-215 & (3, 15) & -2.490e+03 & 2.901e-239 & (3, 14) & -2.055e+03 & 5.115e-231 & (3, 5) & -9.890e+02 & 1.446e-199  \\
(2, 3) & -1.248e+03 & 1.493e-209 & (2, 5) & -1.164e+03 & 1.437e-206 & (2, 12) & -2.490e+03 & 2.901e-239 & (3, 6) & -1.007e+03 & 2.337e-200  \\
(0, 13) & -1.117e+03 & 8.176e-205 & (2, 6) & -1.182e+03 & 3.077e-207 & (6, 12) & -1.104e+03 & 2.646e-204 & (3, 11) & -1.334e+03 & 1.941e-212  \\

    \toprule
     \multicolumn{12}{c}{\textbf{OLMo-2 1B DPO}} \\
     \midrule
     \multicolumn{3}{c}{adj: antonym} & \multicolumn{3}{c}{adj: comparative} & \multicolumn{3}{c}{anim: color} & \multicolumn{3}{c}{anim: can\_fly}\\
     \midrule
     Layer(s) &  t-stat.  &  p-value & Layer(s) &  t-stat. & p-value &  Layer(s) & t-stat. &  p-value &  Layer(s) & t-stat.  & p-value  \\ 
     \midrule
(3, 14) & -1.504e+03 & 1.377e-217 & (0, 0) & -1.092e+03 & 7.928e-204 & (0, 13) & -1.800e+03 & 2.590e-225 & (3, 7) & -9.850e+02 & 2.153e-199  \\
(3, 3) & -9.934e+02 & 9.296e-200 & (0, 14) & -2.055e+03 & 5.115e-231 & (5, 13) & -9.850e+02 & 2.153e-199 & (1, 7) & -9.874e+02 & 1.696e-199  \\
(3, 13) & -1.092e+03 & 7.928e-204 & (0, 15) & -inf & 0.000e+00 & (5, 12) & -1.104e+03 & 2.646e-204 & (3, 4) & -1.061e+03 & 1.400e-202  \\
(3, 15) & -1.800e+03 & 2.590e-225 & (2, 3) & -1.061e+03 & 1.400e-202 & (3, 12) & -1.504e+03 & 1.377e-217 & (4, 7) & -1.036e+03 & 1.498e-201  \\
(0, 0) & -1.117e+03 & 8.176e-205 & (3, 14) & -2.239e+03 & 1.061e-234 & (3, 13) & -1.504e+03 & 1.377e-217 & (1, 6) & -1.003e+03 & 3.523e-200  \\
(0, 15) & -inf & 0.000e+00 & (5, 6) & -9.854e+02 & 2.070e-199 & (3, 11) & -1.504e+03 & 1.377e-217 & (3, 5) & -9.862e+02 & 1.912e-199  \\
(0, 14) & -1.504e+03 & 1.377e-217 & (2, 2) & -1.092e+03 & 7.928e-204 & (7, 13) & -1.029e+03 & 2.730e-201 & (3, 6) & -9.874e+02 & 1.696e-199  \\
(3, 11) & -1.061e+03 & 1.400e-202 & (2, 6) & -1.061e+03 & 1.400e-202 & (2, 13) & -3.484e+03 & 1.053e-253 & (3, 8) & -9.862e+02 & 1.912e-199  \\
(0, 3) & -1.044e+03 & 6.922e-202 & (2, 13) & -1.707e+03 & 5.030e-223 & (5, 11) & -1.008e+03 & 2.303e-200 & (0, 5) & -1.389e+03 & 3.596e-214  \\
(0, 13) & -1.117e+03 & 8.176e-205 & (2, 14) & -1.914e+03 & 6.080e-228 & (5, 14) & -9.874e+02 & 1.696e-199 & (0, 7) & -1.389e+03 & 3.596e-214  \\

    \toprule
     \multicolumn{12}{c}{\textbf{OLMo-2 1B}} \\
     \midrule
     \multicolumn{3}{c}{adj: antonym} & \multicolumn{3}{c}{adj: comparative} & \multicolumn{3}{c}{anim: color} & \multicolumn{3}{c}{anim: can\_fly}\\
     \midrule
     Layer(s) &  t-stat.  &  p-value & Layer(s) &  t-stat. & p-value &  Layer(s) & t-stat. &  p-value &  Layer(s) & t-stat.  & p-value  \\ 
     \midrule
(1, 14) & -1.224e+03 & 1.002e-208 & (1, 1) & -1.334e+03 & 1.941e-212 & (6, 13) & -9.874e+02 & 1.697e-199 & (4, 8) & -9.934e+02 & 9.296e-200  \\
(1, 1) & -1.202e+03 & 5.898e-208 & (1, 14) & -1.274e+03 & 1.925e-210 & (6, 7) & -9.934e+02 & 9.313e-200 & (3, 8) & -1.092e+03 & 7.928e-204  \\
(1, 15) & -inf & 0.000e+00 & (1, 15) & -inf & 0.000e+00 & (6, 14) & -1.334e+03 & 1.941e-212 & (5, 8) & -1.504e+03 & 1.377e-217  \\
(0, 1) & -1.707e+03 & 5.030e-223 & (6, 14) & -1.274e+03 & 1.925e-210 & (7, 14) & -1.147e+03 & 6.063e-206 & (5, 9) & -1.707e+03 & 5.030e-223  \\
(1, 13) & -1.036e+03 & 1.412e-201 & (7, 9) & -1.224e+03 & 1.002e-208 & (6, 6) & -1.069e+03 & 6.546e-203 & (3, 10) & -1.248e+03 & 1.493e-209  \\
(1, 6) & -1.104e+03 & 2.646e-204 & (6, 6) & -1.081e+03 & 2.208e-203 & (6, 8) & -9.850e+02 & 2.153e-199 & (4, 9) & -1.070e+03 & 5.744e-203  \\
(3, 9) & -1.081e+03 & 2.208e-203 & (6, 15) & -inf & 0.000e+00 & (6, 12) & -1.023e+03 & 5.009e-201 & (3, 9) & -1.274e+03 & 1.925e-210  \\
(6, 8) & -1.224e+03 & 1.002e-208 & (1, 13) & -1.023e+03 & 5.009e-201 & (6, 15) & -4.901e+03 & 2.232e-268 & (4, 10) & -1.070e+03 & 5.744e-203  \\
(6, 9) & -1.248e+03 & 1.493e-209 & (7, 10) & -1.202e+03 & 5.898e-208 & (6, 11) & -1.070e+03 & 5.744e-203 & (4, 7) & -1.117e+03 & 8.176e-205  \\
(7, 9) & -1.454e+03 & 4.028e-216 & (0, 1) & -1.409e+03 & 8.728e-215 & (7, 7) & -9.995e+02 & 5.083e-200 & (4, 11) & -1.147e+03 & 6.063e-206  \\

    \bottomrule
     \end{tabular}
     }
     \caption{Contrastive tasks T-Test results for OLMo-2 1B models.}
     \label{tab:app:ttest-olmo1b-contrastive}
 \end{table*}

 \begin{table*}
     \centering
     \resizebox{\textwidth}{!}{
     \begin{tabular}{cccccccccccc}

    \toprule
     \multicolumn{12}{c}{\textbf{BigBench Tasks: T-Test}} \\
    \toprule
     \multicolumn{12}{c}{\textbf{OLMo-2 1B SFT}} \\
     \midrule
     \multicolumn{3}{c}{metaphor\_boolean} & \multicolumn{3}{c}{implicatures} & \multicolumn{3}{c}{object\_counting} & \multicolumn{3}{c}{snarks}\\
     \midrule
     Layer(s) &  t-stat.  &  p-value & Layer(s) &  t-stat. & p-value &  Layer(s) & t-stat. &  p-value &  Layer(s) & t-stat.  & p-value  \\ 
     \midrule
(3, 13) & -1.029e+03 & 2.730e-201 & (13, 14) & -1.831e+03 & 4.827e-226 & (13, 13) & -1.177e+03 & 4.881e-207 & (13, 14) & -1.104e+03 & 2.646e-204  \\
(3, 14) & -1.409e+03 & 8.728e-215 & (13, 13) & -1.430e+03 & 2.068e-215 & (13, 14) & -inf & 0.000e+00 & (13, 13) & -1.353e+03 & 4.982e-213  \\
(3, 3) & -1.290e+03 & 5.353e-211 & (13, 15) & -inf & 0.000e+00 & (13, 15) & -inf & 0.000e+00 & (13, 15) & -inf & 0.000e+00  \\
(3, 15) & -inf & 0.000e+00 & (14, 14) & -4.999e+03 & 3.144e-269 & (14, 14) & -inf & 0.000e+00 & (15, 15) & -inf & 0.000e+00  \\
(3, 12) & -1.202e+03 & 5.898e-208 & (14, 15) & -inf & 0.000e+00 & (14, 15) & -inf & 0.000e+00 & (14, 14) & -1.104e+03 & 2.646e-204  \\
(3, 11) & -9.963e+02 & 6.999e-200 & (15, 15) & -inf & 0.000e+00 & (15, 15) & -inf & 0.000e+00 & (14, 15) & -inf & 0.000e+00  \\
(3, 10) & -3.484e+03 & 1.053e-253 & (12, 14) & -1.061e+03 & 1.400e-202 & (12, 13) & -1.069e+03 & 6.546e-203 & (12, 13) & -1.143e+03 & 8.560e-206  \\
(3, 8) & -1.052e+03 & 3.206e-202 & (12, 13) & -1.263e+03 & 4.322e-210 & (12, 14) & -4.999e+03 & 3.144e-269 & (12, 14) & -1.104e+03 & 2.646e-204  \\
(3, 9) & -2.055e+03 & 5.115e-231 & (12, 12) & -3.484e+03 & 1.053e-253 & (12, 12) & -1.628e+03 & 5.250e-221 & (12, 12) & -inf & 0.000e+00  \\
(3, 4) & -1.090e+03 & 9.535e-204 & (12, 15) & -inf & 0.000e+00 & (12, 15) & -inf & 0.000e+00 & (12, 15) & -inf & 0.000e+00  \\

    \toprule
     \multicolumn{12}{c}{\textbf{OLMo-2 1B DPO}} \\
     \midrule
     \multicolumn{3}{c}{metaphor\_boolean} & \multicolumn{3}{c}{implicatures} & \multicolumn{3}{c}{object\_counting} & \multicolumn{3}{c}{snarks}\\
     \midrule
     Layer(s) &  t-stat.  &  p-value & Layer(s) &  t-stat. & p-value &  Layer(s) & t-stat. &  p-value &  Layer(s) & t-stat.  & p-value  \\ 
     \midrule
(3, 13) & -9.890e+02 & 1.446e-199 & (13, 14) & -1.947e+03 & 1.089e-228 & (13, 13) & -1.216e+03 & 1.849e-208 & (13, 14) & -1.029e+03 & 2.730e-201  \\
(3, 14) & -1.224e+03 & 1.002e-208 & (13, 13) & -1.043e+03 & 7.448e-202 & (13, 14) & -inf & 0.000e+00 & (13, 13) & -1.069e+03 & 6.546e-203  \\
(3, 10) & -2.055e+03 & 5.115e-231 & (13, 15) & -inf & 0.000e+00 & (13, 15) & -inf & 0.000e+00 & (13, 15) & -inf & 0.000e+00  \\
(3, 3) & -1.177e+03 & 4.881e-207 & (14, 14) & -4.999e+03 & 3.144e-269 & (14, 14) & -inf & 0.000e+00 & (15, 15) & -inf & 0.000e+00  \\
(3, 15) & -inf & 0.000e+00 & (14, 15) & -inf & 0.000e+00 & (14, 15) & -inf & 0.000e+00 & (14, 14) & -1.017e+03 & 8.739e-201  \\
(3, 11) & -1.061e+03 & 1.400e-202 & (15, 15) & -inf & 0.000e+00 & (15, 15) & -inf & 0.000e+00 & (14, 15) & -inf & 0.000e+00  \\
(3, 12) & -1.454e+03 & 4.028e-216 & (12, 14) & -1.147e+03 & 6.063e-206 & (12, 13) & -1.102e+03 & 3.288e-204 & (12, 14) & -1.023e+03 & 5.009e-201  \\
(3, 8) & -9.963e+02 & 6.999e-200 & (12, 13) & -1.012e+03 & 1.483e-200 & (12, 14) & -2.280e+03 & 1.754e-235 & (12, 12) & -1.454e+03 & 4.028e-216  \\
(3, 9) & -1.454e+03 & 4.028e-216 & (12, 12) & -inf & 0.000e+00 & (12, 12) & -1.914e+03 & 6.080e-228 & (12, 15) & -inf & 0.000e+00  \\
(5, 6) & -inf & 0.000e+00 & (12, 15) & -inf & 0.000e+00 & (12, 15) & -inf & 0.000e+00 & (12, 13) & -1.060e+03 & 1.561e-202  \\

    \toprule
     \multicolumn{12}{c}{\textbf{OLMo-2 1B}} \\
     \midrule
     \multicolumn{3}{c}{metaphor\_boolean} & \multicolumn{3}{c}{implicatures} & \multicolumn{3}{c}{object\_counting} & \multicolumn{3}{c}{snarks}\\
     \midrule
     Layer(s) &  t-stat.  &  p-value & Layer(s) &  t-stat. & p-value &  Layer(s) & t-stat. &  p-value &  Layer(s) & t-stat.  & p-value  \\ 
     \midrule
(0, 0) & -4.999e+03 & 3.144e-269 & (1, 14) & -1.475e+03 & 9.170e-217 & (2, 5) & -1.947e+03 & 1.089e-228 & (12, 13) & -inf & 0.000e+00  \\
(0, 15) & -inf & 0.000e+00 & (0, 1) & -1.060e+03 & 1.561e-202 & (3, 4) & -4.999e+03 & 3.144e-269 & (12, 14) & -inf & 0.000e+00  \\
(0, 14) & -1.947e+03 & 1.089e-228 & (1, 1) & -1.216e+03 & 1.849e-208 & (3, 5) & -1.947e+03 & 1.089e-228 & (12, 12) & -3.552e+03 & 1.544e-254  \\
(0, 13) & -inf & 0.000e+00 & (1, 15) & -inf & 0.000e+00 & (2, 3) & -inf & 0.000e+00 & (12, 15) & -inf & 0.000e+00  \\
(1, 2) & -1.655e+03 & 1.060e-221 & (0, 14) & -2.537e+03 & 4.608e-240 & (3, 6) & -1.353e+03 & 4.982e-213 & (1, 7) & -2.490e+03 & 2.901e-239  \\
(0, 12) & -3.552e+03 & 1.544e-254 & (0, 0) & -1.060e+03 & 1.561e-202 & (2, 7) & -1.029e+03 & 2.730e-201 & (2, 7) & -1.707e+03 & 5.030e-223  \\
(0, 11) & -1.224e+03 & 1.002e-208 & (0, 15) & -inf & 0.000e+00 & (3, 8) & -1.800e+03 & 2.590e-225 & (0, 3) & -9.890e+02 & 1.447e-199  \\
(1, 1) & -3.552e+03 & 1.544e-254 & (1, 11) & -inf & 0.000e+00 & (2, 4) & -3.552e+03 & 1.544e-254 & (0, 7) & -1.800e+03 & 2.590e-225  \\
(1, 15) & -inf & 0.000e+00 & (13, 14) & -inf & 0.000e+00 & (3, 7) & -9.890e+02 & 1.446e-199 & (2, 4) & -1.003e+03 & 3.487e-200  \\
(1, 13) & -4.999e+03 & 3.144e-269 & (1, 9) & -inf & 0.000e+00 & (3, 3) & -inf & 0.000e+00 & (4, 7) & -2.860e+03 & 3.234e-245  \\
    \bottomrule
     \end{tabular}
     }
     \caption{BigBench tasks T-Test results for OLMo-2 1B models.}
     \label{tab:app:ttest-olmo1b-bigbench}
 \end{table*}

 \begin{table*}
     \centering
     \resizebox{\textwidth}{!}{
     \begin{tabular}{cccccccccccc}

    \toprule
     \multicolumn{12}{c}{\textbf{Contrastive Tasks: T-Test}} \\
    \toprule
     \multicolumn{12}{c}{\textbf{OLMo-2 7B SFT}} \\
     \midrule
     \multicolumn{3}{c}{adj: antonym} & \multicolumn{3}{c}{adj: comparative} & \multicolumn{3}{c}{anim: color} & \multicolumn{3}{c}{anim: can\_fly}\\
     \midrule
     Layer(s) &  t-stat.  &  p-value & Layer(s) &  t-stat. & p-value &  Layer(s) & t-stat. &  p-value &  Layer(s) & t-stat.  & p-value  \\ 
     \midrule
(7, 18) & -1.370e+03 & 1.458e-213 & (6, 8) & -1.044e+03 & 6.922e-202 & (6, 8) & -9.910e+02 & 1.184e-199 & (0, 13) & -9.963e+02 & 6.999e-200  \\
(8, 15) & -1.182e+03 & 3.077e-207 & (7, 8) & -1.061e+03 & 1.400e-202 & (9, 13) & -1.061e+03 & 1.400e-202 & (0, 4) & -9.850e+02 & 2.153e-199  \\
(8, 16) & -1.202e+03 & 5.898e-208 & (5, 9) & -1.092e+03 & 7.928e-204 & (9, 14) & -1.117e+03 & 8.176e-205 & (0, 5) & -9.910e+02 & 1.184e-199  \\
(8, 17) & -1.274e+03 & 1.925e-210 & (6, 9) & -1.061e+03 & 1.400e-202 & (2, 16) & -1.008e+03 & 2.303e-200 & (0, 6) & -1.061e+03 & 1.400e-202  \\
(9, 24) & -2.490e+03 & 2.901e-239 & (7, 9) & -1.070e+03 & 5.744e-203 & (2, 17) & -1.070e+03 & 5.744e-203 & (0, 0) & -1.128e+03 & 3.131e-205  \\
(4, 14) & -1.202e+03 & 5.898e-208 & (8, 8) & -1.036e+03 & 1.412e-201 & (3, 15) & -1.164e+03 & 1.437e-206 & (0, 7) & -9.996e+02 & 5.049e-200  \\
(4, 19) & -1.454e+03 & 4.028e-216 & (8, 27) & -2.490e+03 & 2.901e-239 & (4, 13) & -9.996e+02 & 5.049e-200 & (0, 12) & -9.862e+02 & 1.912e-199  \\
(7, 17) & -1.248e+03 & 1.493e-209 & (8, 28) & -inf & 0.000e+00 & (4, 14) & -1.008e+03 & 2.303e-200 & (0, 31) & -3.484e+03 & 1.053e-253  \\
(8, 18) & -1.504e+03 & 1.377e-217 & (8, 30) & -4.901e+03 & 2.232e-268 & (5, 16) & -1.081e+03 & 2.208e-203 & (1, 13) & -9.890e+02 & 1.447e-199  \\
(4, 15) & -1.202e+03 & 5.898e-208 & (8, 31) & -inf & 0.000e+00 & (6, 6) & -1.003e+03 & 3.487e-200 & (0, 1) & -1.043e+03 & 7.448e-202  \\

    \toprule
     \multicolumn{12}{c}{\textbf{OLMo-2 7B DPO}} \\
     \midrule
     \multicolumn{3}{c}{adj: antonym} & \multicolumn{3}{c}{adj: comparative} & \multicolumn{3}{c}{anim: color} & \multicolumn{3}{c}{anim: can\_fly}\\
     \midrule
     Layer(s) &  t-stat.  &  p-value & Layer(s) &  t-stat. & p-value &  Layer(s) & t-stat. &  p-value &  Layer(s) & t-stat.  & p-value  \\ 
     \midrule
(8, 26) & -inf & 0.000e+00 & (6, 9) & -1.061e+03 & 1.400e-202 & (8, 14) & -9.850e+02 & 2.153e-199 & (0, 13) & -1.023e+03 & 5.196e-201  \\
(8, 25) & -4.901e+03 & 2.232e-268 & (7, 9) & -1.036e+03 & 1.412e-201 & (9, 13) & -9.934e+02 & 9.296e-200 & (0, 5) & -1.060e+03 & 1.561e-202  \\
(1, 17) & -1.274e+03 & 1.925e-210 & (5, 9) & -1.117e+03 & 8.176e-205 & (9, 14) & -9.910e+02 & 1.184e-199 & (0, 0) & -1.655e+03 & 1.060e-221  \\
(7, 27) & -2.490e+03 & 2.901e-239 & (8, 8) & -1.036e+03 & 1.412e-201 & (5, 18) & -1.052e+03 & 3.206e-202 & (0, 4) & -1.060e+03 & 1.561e-202  \\
(7, 28) & -inf & 0.000e+00 & (8, 31) & -inf & 0.000e+00 & (6, 8) & -9.862e+02 & 1.912e-199 & (0, 12) & -1.017e+03 & 8.988e-201  \\
(8, 15) & -1.132e+03 & 2.327e-205 & (4, 19) & -1.454e+03 & 4.028e-216 & (6, 14) & -1.012e+03 & 1.452e-200 & (0, 26) & -1.104e+03 & 2.646e-204  \\
(8, 16) & -1.164e+03 & 1.437e-206 & (7, 8) & -1.044e+03 & 6.922e-202 & (6, 16) & -9.934e+02 & 9.296e-200 & (0, 28) & -1.334e+03 & 1.941e-212  \\
(8, 17) & -1.224e+03 & 1.002e-208 & (8, 9) & -1.036e+03 & 1.412e-201 & (6, 18) & -1.052e+03 & 3.206e-202 & (0, 31) & -1.454e+03 & 4.028e-216  \\
(8, 18) & -1.370e+03 & 1.458e-213 & (9, 9) & -1.044e+03 & 6.922e-202 & (9, 11) & -9.962e+02 & 7.026e-200 & (0, 1) & -1.290e+03 & 5.353e-211  \\
(8, 24) & -2.860e+03 & 3.234e-245 & (9, 28) & -4.901e+03 & 2.232e-268 & (9, 12) & -9.890e+02 & 1.447e-199 & (0, 3) & -1.389e+03 & 3.596e-214  \\

    \toprule
     \multicolumn{12}{c}{\textbf{OLMo-2 7B}} \\
     \midrule
     \multicolumn{3}{c}{adj: antonym} & \multicolumn{3}{c}{adj: comparative} & \multicolumn{3}{c}{anim: color} & \multicolumn{3}{c}{anim: can\_fly}\\
     \midrule
     Layer(s) &  t-stat.  &  p-value & Layer(s) &  t-stat. & p-value &  Layer(s) & t-stat. &  p-value &  Layer(s) & t-stat.  & p-value  \\ 
     \midrule
(7, 18) & -2.490e+03 & 2.901e-239 & (3, 18) & -2.490e+03 & 2.901e-239 & (9, 12) & -inf & 0.000e+00 & (0, 11) & -inf & 0.000e+00  \\
(2, 24) & -4.901e+03 & 2.232e-268 & (7, 10) & -2.055e+03 & 5.115e-231 & (9, 13) & -inf & 0.000e+00 & (0, 10) & -inf & 0.000e+00  \\
(2, 25) & -inf & 0.000e+00 & (3, 12) & -2.055e+03 & 5.115e-231 & (9, 14) & -inf & 0.000e+00 & (0, 12) & -inf & 0.000e+00  \\
(4, 18) & -2.239e+03 & 1.061e-234 & (4, 12) & -1.800e+03 & 2.590e-225 & (2, 16) & -1.914e+03 & 6.080e-228 & (0, 6) & -3.484e+03 & 1.053e-253  \\
(7, 17) & -2.239e+03 & 1.061e-234 & (4, 18) & -1.914e+03 & 6.080e-228 & (4, 5) & -inf & 0.000e+00 & (0, 0) & -1.164e+03 & 1.437e-206  \\
(1, 3) & -1.628e+03 & 5.250e-221 & (5, 11) & -1.504e+03 & 1.377e-217 & (5, 5) & -2.490e+03 & 2.901e-239 & (0, 5) & -2.055e+03 & 5.115e-231  \\
(2, 16) & -1.202e+03 & 5.898e-208 & (5, 17) & -1.628e+03 & 5.250e-221 & (5, 12) & -4.901e+03 & 2.232e-268 & (0, 7) & -2.490e+03 & 2.901e-239  \\
(2, 27) & -4.901e+03 & 2.232e-268 & (5, 18) & -1.562e+03 & 3.312e-219 & (5, 28) & -4.901e+03 & 2.232e-268 & (0, 9) & -4.901e+03 & 2.232e-268  \\
(7, 25) & -inf & 0.000e+00 & (5, 27) & -3.484e+03 & 1.053e-253 & (5, 31) & -inf & 0.000e+00 & (0, 13) & -inf & 0.000e+00  \\
(7, 29) & -inf & 0.000e+00 & (6, 18) & -2.239e+03 & 1.061e-234 & (6, 6) & -4.901e+03 & 2.232e-268 & (0, 14) & -inf & 0.000e+00  \\

    \bottomrule
     \end{tabular}
     }
     \caption{Contrastive tasks T-Test results for OLMo-2 7B models.}
     \label{tab:app:ttest-olmo7b-contrastive}
 \end{table*}

  \begin{table*}
     \centering
    \resizebox{\textwidth}{!}{
     \begin{tabular}{cccccccccccc}

     \toprule
     \multicolumn{12}{c}{\textbf{BigBench Tasks: T-Test}} \\
    \toprule
     \multicolumn{12}{c}{\textbf{OLMo-2 7B SFT}} \\
     \midrule
     \multicolumn{3}{c}{metaphor\_boolean} & \multicolumn{3}{c}{implicatures} & \multicolumn{3}{c}{object\_counting} & \multicolumn{3}{c}{snarks}\\
     \midrule
     Layer(s) &  t-stat.  &  p-value & Layer(s) &  t-stat. & p-value &  Layer(s) & t-stat. &  p-value &  Layer(s) & t-stat.  & p-value  \\ 
     \midrule
(15, 24) & -9.850e+02 & 2.153e-199 & (12, 26) & -1.061e+03 & 1.400e-202 & (1, 25) & -1.562e+03 & 3.312e-219 & (15, 16) & -1.430e+03 & 2.068e-215  \\
(15, 22) & -1.029e+03 & 2.861e-201 & (12, 14) & -1.044e+03 & 6.922e-202 & (1, 24) & -1.800e+03 & 2.590e-225 & (15, 28) & -1.012e+03 & 1.483e-200  \\
(15, 27) & -1.070e+03 & 5.744e-203 & (12, 27) & -2.055e+03 & 5.115e-231 & (0, 2) & -2.239e+03 & 1.061e-234 & (15, 30) & -1.302e+03 & 2.114e-211  \\
(15, 23) & -9.850e+02 & 2.153e-199 & (12, 28) & -1.070e+03 & 5.744e-203 & (14, 27) & -inf & 0.000e+00 & (15, 15) & -1.947e+03 & 1.089e-228  \\
(15, 26) & -9.862e+02 & 1.911e-199 & (12, 29) & -4.901e+03 & 2.232e-268 & (22, 27) & -inf & 0.000e+00 & (15, 31) & -inf & 0.000e+00  \\
(15, 28) & -1.117e+03 & 8.176e-205 & (12, 30) & -2.860e+03 & 3.234e-245 & (1, 27) & -inf & 0.000e+00 & (15, 29) & -9.962e+02 & 7.026e-200  \\
(15, 21) & -9.995e+02 & 5.083e-200 & (12, 25) & -1.117e+03 & 8.176e-205 & (1, 23) & -1.104e+03 & 2.646e-204 & (15, 18) & -2.537e+03 & 4.608e-240  \\
(15, 29) & -inf & 0.000e+00 & (12, 12) & -1.239e+03 & 3.019e-209 & (1, 28) & -3.484e+03 & 1.053e-253 & (15, 26) & -1.182e+03 & 3.077e-207  \\
(15, 25) & -1.003e+03 & 3.523e-200 & (12, 24) & -1.036e+03 & 1.412e-201 & (22, 26) & -4.901e+03 & 2.232e-268 & (15, 19) & -1.586e+03 & 6.960e-220  \\
(15, 30) & -2.239e+03 & 1.061e-234 & (12, 31) & -inf & 0.000e+00 & (14, 21) & -1.029e+03 & 2.730e-201 & (15, 27) & -1.051e+03 & 3.507e-202  \\

    \toprule
     \multicolumn{12}{c}{\textbf{OLMo-2 7B DPO}} \\
     \midrule
     \multicolumn{3}{c}{metaphor\_boolean} & \multicolumn{3}{c}{implicatures} & \multicolumn{3}{c}{object\_counting} & \multicolumn{3}{c}{snarks}\\
     \midrule
     Layer(s) &  t-stat.  &  p-value & Layer(s) &  t-stat. & p-value &  Layer(s) & t-stat. &  p-value &  Layer(s) & t-stat.  & p-value  \\ 
     \midrule
(14, 15) & -1.947e+03 & 1.089e-228 & (12, 26) & -9.874e+02 & 1.696e-199 & (14, 27) & -inf & 0.000e+00 & (14, 30) & -1.036e+03 & 1.498e-201  \\
(14, 27) & -1.370e+03 & 1.458e-213 & (12, 14) & -1.159e+03 & 2.143e-206 & (14, 28) & -inf & 0.000e+00 & (14, 14) & -1.947e+03 & 1.089e-228  \\
(14, 26) & -9.995e+02 & 5.083e-200 & (11, 14) & -1.079e+03 & 2.581e-203 & (14, 29) & -9.854e+02 & 2.069e-199 & (14, 31) & -inf & 0.000e+00  \\
(9, 22) & -1.090e+03 & 9.535e-204 & (12, 27) & -1.008e+03 & 2.303e-200 & (14, 25) & -1.182e+03 & 3.077e-207 & (14, 28) & -1.389e+03 & 3.596e-214  \\
(14, 22) & -1.090e+03 & 9.535e-204 & (12, 29) & -1.628e+03 & 5.250e-221 & (14, 26) & -1.800e+03 & 2.590e-225 & (14, 29) & -1.003e+03 & 3.523e-200  \\
(14, 24) & -1.102e+03 & 3.288e-204 & (12, 30) & -1.202e+03 & 5.898e-208 & (14, 21) & -1.504e+03 & 1.377e-217 & (15, 16) & -2.537e+03 & 4.608e-240  \\
(14, 28) & -1.224e+03 & 1.002e-208 & (12, 24) & -9.850e+02 & 2.153e-199 & (12, 28) & -2.860e+03 & 3.234e-245 & (14, 27) & -1.069e+03 & 6.546e-203  \\
(14, 23) & -1.060e+03 & 1.561e-202 & (12, 28) & -9.854e+02 & 2.070e-199 & (12, 27) & -4.901e+03 & 2.232e-268 & (14, 19) & -1.389e+03 & 3.596e-214  \\
(9, 27) & -1.248e+03 & 1.493e-209 & (12, 25) & -1.012e+03 & 1.483e-200 & (14, 22) & -2.055e+03 & 5.115e-231 & (14, 26) & -1.007e+03 & 2.337e-200  \\
(9, 14) & -2.915e+03 & 4.935e-246 & (12, 21) & -1.430e+03 & 2.068e-215 & (12, 29) & -1.430e+03 & 2.068e-215 & (15, 17) & -1.947e+03 & 1.089e-228  \\

    \toprule
     \multicolumn{12}{c}{\textbf{OLMo-2 7B}} \\
     \midrule
     \multicolumn{3}{c}{metaphor\_boolean} & \multicolumn{3}{c}{implicatures} & \multicolumn{3}{c}{object\_counting} & \multicolumn{3}{c}{snarks}\\
     \midrule
     Layer(s) &  t-stat.  &  p-value & Layer(s) &  t-stat. & p-value &  Layer(s) & t-stat. &  p-value &  Layer(s) & t-stat.  & p-value  \\ 
     \midrule
(0, 2) & -9.934e+02 & 9.296e-200 & (1, 2) & -1.090e+03 & 9.535e-204 & (14, 27) & - & - & (2, 13) & -9.874e+02 & 1.697e-199  \\
(0, 29) & -3.484e+03 & 1.053e-253 & (1, 1) & -1.115e+03 & 1.055e-204 & (14, 28) & - & - & (2, 4) & -9.910e+02 & 1.184e-199  \\
(0, 10) & -1.081e+03 & 2.208e-203 & (1, 31) & -inf & 0.000e+00 & (14, 29) & - & - & (2, 25) & -1.914e+03 & 6.080e-228  \\
(0, 6) & -1.202e+03 & 5.898e-208 & (1, 30) & -inf & 0.000e+00 & (14, 25) & - & - & (2, 14) & -1.070e+03 & 5.744e-203  \\
(0, 7) & -9.890e+02 & 1.447e-199 & (1, 27) & -inf & 0.000e+00 & (14, 26) & - & - & (2, 12) & -1.132e+03 & 2.327e-205  \\
(0, 8) & -9.995e+02 & 5.083e-200 & (1, 25) & -1.504e+03 & 1.377e-217 & (14, 21) & - & - & (2, 23) & -4.901e+03 & 2.232e-268  \\
(3, 13) & -1.628e+03 & 5.250e-221 & (2, 25) & -2.239e+03 & 1.061e-234 & (12, 28) & - & - & (2, 24) & -4.901e+03 & 2.232e-268  \\
(0, 5) & -9.910e+02 & 1.184e-199 & (2, 2) & -1.102e+03 & 3.288e-204 & (12, 27) & - & - & (2, 26) & -1.914e+03 & 6.080e-228  \\
(0, 30) & -2.490e+03 & 2.901e-239 & (2, 31) & -inf & 0.000e+00 & (14, 22) & - & - & (2, 15) & -9.890e+02 & 1.446e-199  \\
(0, 1) & -9.934e+02 & 9.313e-200 & (1, 17) & -inf & 0.000e+00 & (12, 29) & - & - & (2, 8) & -1.008e+03 & 2.303e-200  \\
    \bottomrule
     \end{tabular}
     }
     \caption{BigBench tasks T-Test results for OLMo-2 7B models. \textsc{object\_counting} scores for OLMo-2 7B are missing due to a model error while running the task.}
     \label{tab:app:ttest-olmo7b-bigbench}
 \end{table*}

\begin{figure*}
    \centering
    \includegraphics[width=\linewidth,scale=0.7,trim={0 15cm 0cm 0cm},clip]{latex/norm_rank_olmo1b_contrastive.pdf}
    \caption{Effects of 1- and 2-layer patching configurations on the reciprocal rank of the target token for OLMo-2 1B models. Each square of the x- and y-coordinate grid represents the corresponding layers of the model being patched. Coordinates where x=y represent 1-layer patching.}
    \label{fig:1LP+2LP_logit}
\end{figure*}

 \begin{table*}
     \centering
     \small
     \begin{tabular}{cccccccc}

     \toprule
     \multicolumn{8}{c}{\textbf{Contrastive Tasks - Rank 1B}} \\
     \toprule
     \multicolumn{8}{c}{\textbf{OLMo-2 1B SFT}} \\
     \midrule
     \multicolumn{2}{c}{adj: ant} & \multicolumn{2}{c}{adj: comp} & \multicolumn{2}{c}{anim: color} & \multicolumn{2}{c}{anim: can\_fly}\\
     \midrule
     Layer(s) &  Rank contrib. & Layer(s) &  Rank contrib. & Layer(s) &  Rank contrib. &  Layer(s) &  Rank contrib.\\ 
     \midrule
(3, 3) & 0.14 & (0, 0) & 0.08  & (3, 13) & 0.11& (3, 7) & 0.05\\
(3, 14) & 0.14 & (0, 15) & 0.08 & (5, 13) & 0.11 & (1, 7) & 0.03\\
(3, 15) & 0.14 & (0, 14) & 0.07 &  (3, 12) & 0.10& (2, 7) & 0.03\\
(0, 0) & 0.13 & (4, 6) & 0.07 & (5, 12) & 0.10 & (3, 4) & 0.03\\
(0, 15) & 0.13 & (2, 3) & 0.06 & (0, 13) & 0.09 & (3, 8) & 0.03\\
(3, 11) & 0.11 & (3, 3) & 0.06  & (2, 13) & 0.09  &(3, 9) & 0.03\\
(3, 13) & 0.11 & (3, 14) & 0.06  & (3, 11) & 0.09 &(1, 6) & 0.02 \\
(0, 14) & 0.10 & (3, 15) & 0.06 & (3, 14) & 0.09  &(3, 5) & 0.02\\
(2, 3) & 0.10 & (2, 5) & 0.05 & (2, 12) & 0.08  &(3, 6) & 0.02\\
(0, 13) & 0.08 & (2, 6) & 0.05 & (6, 12) & 0.08 & (3, 11) & 0.02\\

    \toprule
     \multicolumn{8}{c}{\textbf{OLMo-2 1B DPO}} \\
     \midrule
     \multicolumn{2}{c}{adj: ant} & \multicolumn{2}{c}{adj: comp} & \multicolumn{2}{c}{anim: color} & \multicolumn{2}{c}{anim: can\_fly}\\
     \midrule
     Layer(s) &  Rank contrib. & Layer(s) &  Rank contrib. & Layer(s) &  Rank contrib. &  Layer(s) &  Rank contrib.\\ 
     \midrule
(3, 14) & 0.13 & (0, 0) & 0.07 & (0, 13) & 0.12 & (3, 7) & 0.04  \\
(3, 3) & 0.11  & (0, 14) & 0.07 & (5, 13) & 0.12 & (1, 7) & 0.03\\
(3, 13) & 0.11 & (0, 15) & 0.07 & (5, 12) & 0.11 & (3, 4) & 0.03 \\
(3, 15) & 0.11 & (2, 3) & 0.03 & (3, 12) & 0.09& (4, 7) & 0.03 \\
(0, 0) & 0.10  & (3, 14) & 0.03 & (3, 13) & 0.09 & (1, 6) & 0.02 \\
(0, 15) & 0.10 & (5, 6) & 0.03 & (3, 11) & 0.08&  (3, 5) & 0.02 \\
(0, 14) & 0.09 & (2, 2) & 0.02 & (7, 13) & 0.08& (3, 6) & 0.02  \\
(3, 11) & 0.08 & (2, 6) & 0.02 & (2, 13) & 0.07 & (3, 8) & 0.02 \\
(0, 3) & 0.07  & (2, 13) & 0.02 & (5, 11) & 0.07& (0, 5) & 0.01 \\
(0, 13) & 0.07 & (2, 14) & 0.02 & (5, 14) & 0.07 & (0, 7) & 0.01 \\

    \toprule
     \multicolumn{8}{c}{\textbf{OLMo-2 1B}} \\
     \midrule
     \multicolumn{2}{c}{adj: ant} & \multicolumn{2}{c}{adj: comp} & \multicolumn{2}{c}{anim: color} & \multicolumn{2}{c}{anim: can\_fly}\\
     \midrule
     Layer(s) &  Rank contrib. & Layer(s) &  Rank contrib. & Layer(s) &  Rank contrib. &  Layer(s) &  Rank contrib.\\ 
     \midrule
(1, 14) & 0.16 & (1, 1) & 0.08 & (6, 13) & 0.15 & (4, 8) & 0.17  \\
(1, 1) & 0.14 & (1, 14) & 0.08 & (6, 7) & 0.14 & (3, 8) & 0.15  \\
(1, 15) & 0.14 & (1, 15) & 0.08 & (6, 14) & 0.13 & (5, 8) & 0.13  \\
(0, 1) & 0.13 & (6, 14) & 0.07 & (7, 14) & 0.12 & (5, 9) & 0.11  \\
(1, 13) & 0.11 & (7, 9) & 0.07 & (6, 6) & 0.11 & (3, 10) & 0.08  \\
(1, 6) & 0.10 & (6, 6) & 0.06 & (6, 8) & 0.11 & (4, 9) & 0.08  \\
(3, 9) & 0.09 & (6, 15) & 0.06 & (6, 12) & 0.11 & (3, 9) & 0.07  \\
(6, 8) & 0.09 & (1, 13) & 0.05 & (6, 15) & 0.11 & (4, 10) & 0.07  \\
(6, 9) & 0.09 & (7, 10) & 0.05 & (6, 11) & 0.10 & (4, 7) & 0.06  \\
(7, 9) & 0.09 & (0, 1) & 0.04 & (7, 7) & 0.10 & (4, 11) & 0.05  \\

    \bottomrule
     \end{tabular}
     \caption{Numerical rank contributions of OLMo-2 1B models for various patching configurations. For each model, the 10 highest-contributing combinations are shown.}
     \label{tab:app:contrastive-rank-olmo1b-num}
 \end{table*}

 \begin{table*}
     \centering
     \small
     \begin{tabular}{cccccccc}

     \toprule
     \multicolumn{8}{c}{\textbf{BigBench Tasks - Logit 1B}} \\
     \toprule
     \multicolumn{8}{c}{\textbf{OLMo-2 1B SFT}} \\
     \midrule
     \multicolumn{2}{c}{metaphor\_boolean} & \multicolumn{2}{c}{implicatures} & \multicolumn{2}{c}{object\_counting} & \multicolumn{2}{c}{snarks}\\
     \midrule
     Layer(s) &  Logit contrib. & Layer(s) &  Logit contrib. & Layer(s) &  Logit contrib. &  Layer(s) &  Logit contrib.\\ 
     \midrule
(3, 13) & 3.16 & (13, 14) & 0.31 & (13, 13) & 0.17 & (13, 14) & 0.21  \\
(3, 14) & 3.02 & (13, 13) & 0.29 & (13, 14) & 0.17 & (13, 13) & 0.19  \\
(3, 3) & 2.8 & (13, 15) & 0.29 & (13, 15) & 0.17 & (13, 15) & 0.19  \\
(3, 15) & 2.8 & (14, 14) & 0.1 & (14, 14) & 0.13 & (15, 15) & 0.0  \\
(3, 12) & 2.79 & (14, 15) & 0.1 & (14, 15) & 0.13 & (14, 14) & -0.05  \\
(3, 11) & 2.67 & (15, 15) & 0.0 & (15, 15) & 0.0 & (14, 15) & -0.05  \\
(3, 10) & 2.45 & (12, 14) & -0.37 & (12, 13) & -0.34 & (12, 13) & -0.7  \\
(3, 8) & 2.34 & (12, 13) & -0.4 & (12, 14) & -0.34 & (12, 14) & -0.71  \\
(3, 9) & 2.23 & (12, 12) & -0.49 & (12, 12) & -0.38 & (12, 12) & -0.73  \\
(3, 4) & 1.49 & (12, 15) & -0.49 & (12, 15) & -0.38 & (12, 15) & -0.73  \\

    \toprule
     \multicolumn{8}{c}{\textbf{OLMo-2 1B DPO}} \\
     \midrule
     \multicolumn{2}{c}{metaphor\_boolean} & \multicolumn{2}{c}{implicatures} & \multicolumn{2}{c}{object\_counting} & \multicolumn{2}{c}{snarks}\\
     \midrule
     Layer(s) &  Logit contrib. & Layer(s) &  Logit contrib. & Layer(s) &  Logit contrib. &  Layer(s) &  Logit contrib.\\ 
     \midrule
(3, 13) & 2.77 & (13, 14) & 0.14 & (13, 13) & 0.18 & (13, 14) & 0.12  \\
(3, 14) & 2.6 & (13, 13) & 0.13 & (13, 14) & 0.18 & (13, 13) & 0.1  \\
(3, 10) & 2.44 & (13, 15) & 0.13 & (13, 15) & 0.18 & (13, 15) & 0.1  \\
(3, 3) & 2.43 & (14, 14) & 0.07 & (14, 14) & 0.12 & (15, 15) & 0.0  \\
(3, 15) & 2.43 & (14, 15) & 0.07 & (14, 15) & 0.12 & (14, 14) & -0.01  \\
(3, 11) & 2.41 & (15, 15) & 0.0 & (15, 15) & 0.0 & (14, 15) & -0.01  \\
(3, 12) & 2.41 & (12, 14) & -0.58 & (12, 13) & -0.35 & (12, 14) & -0.4  \\
(3, 8) & 2.07 & (12, 13) & -0.64 & (12, 14) & -0.35 & (12, 12) & -0.42  \\
(3, 9) & 1.9 & (12, 12) & -0.67 & (12, 12) & -0.4 & (12, 15) & -0.42  \\
(5, 6) & 1.46 & (12, 15) & -0.67 & (12, 15) & -0.4 & (12, 13) & -0.43  \\

    \toprule
     \multicolumn{8}{c}{\textbf{OLMo-2 1B}} \\
     \midrule
     \multicolumn{2}{c}{metaphor\_boolean} & \multicolumn{2}{c}{implicatures} & \multicolumn{2}{c}{object\_counting} & \multicolumn{2}{c}{snarks}\\
     \midrule
     Layer(s) &  Logit contrib. & Layer(s) &  Logit contrib. & Layer(s) &  Logit contrib. &  Layer(s) &  Logit contrib.\\ 
     \midrule
(0, 0) & 5.49 & (1, 14) & 1.05 & (2, 5) & 8.72 & (12, 13) & 2.5  \\
(0, 15) & 5.49 & (0, 1) & 1.02 & (3, 4) & 8.69 & (12, 14) & 2.47  \\
(0, 14) & 5.34 & (1, 1) & 0.98 & (3, 5) & 8.62 & (12, 12) & 2.46  \\
(0, 13) & 5.22 & (1, 15) & 0.98 & (2, 3) & 8.59 & (12, 15) & 2.46  \\
(1, 2) & 5.16 & (0, 14) & 0.97 & (3, 6) & 8.41 & (1, 7) & 2.31  \\
(0, 12) & 5.03 & (0, 0) & 0.93 & (2, 7) & 8.4 & (2, 7) & 2.15  \\
(0, 11) & 4.97 & (0, 15) & 0.93 & (3, 8) & 8.35 & (0, 3) & 1.85  \\
(1, 1) & 4.96 & (1, 11) & 0.92 & (2, 4) & 8.29 & (0, 7) & 1.85  \\
(1, 15) & 4.96 & (13, 14) & 0.89 & (3, 7) & 8.27 & (2, 4) & 1.81  \\
(1, 13) & 4.86 & (1, 9) & 0.87 & (3, 3) & 8.16 & (4, 7) & 1.8  \\

    \bottomrule
     \end{tabular}
     \caption{Numerical logit contributions of OLMo-2 1B models for various patching configurations. For each model, the 10 highest-contributing combinations are shown.}
     \label{tab:app:bb-logit-olmo1b-num}
 \end{table*}

 \begin{table*}
     \centering
     \small
     \begin{tabular}{cccccccc}

     \toprule
     \multicolumn{8}{c}{\textbf{BigBench Tasks - Logit 7B}} \\
     \toprule
     \multicolumn{8}{c}{\textbf{OLMo-2 7B SFT}} \\
     \midrule
     \multicolumn{2}{c}{metaphor\_boolean} & \multicolumn{2}{c}{implicatures} & \multicolumn{2}{c}{object\_counting} & \multicolumn{2}{c}{snarks}\\
     \midrule
     Layer(s) &  Logit contrib. & Layer(s) &  Logit contrib. & Layer(s) &  Logit contrib. &  Layer(s) &  Logit contrib.\\ 
     \midrule
(15, 24) & 2.6 & (12, 26) & 1.27 & (1, 25) & 0.79 & (15, 16) & 3.74  \\
(15, 22) & 2.59 & (12, 14) & 1.22 & (1, 24) & 0.66 & (15, 28) & 3.48  \\
(15, 27) & 2.56 & (12, 27) & 1.19 & (0, 2) & 0.61 & (15, 30) & 3.48  \\
(15, 23) & 2.55 & (12, 28) & 1.18 & (14, 27) & 0.61 & (15, 15) & 3.44  \\
(15, 26) & 2.54 & (12, 29) & 1.15 & (22, 27) & 0.61 & (15, 31) & 3.44  \\
(15, 28) & 2.51 & (12, 30) & 1.08 & (1, 27) & 0.6 & (15, 29) & 3.42  \\
(15, 21) & 2.47 & (12, 25) & 1.05 & (1, 23) & 0.59 & (15, 18) & 3.4  \\
(15, 29) & 2.45 & (12, 12) & 1.0 & (1, 28) & 0.59 & (15, 26) & 3.4  \\
(15, 25) & 2.43 & (12, 24) & 1.0 & (22, 26) & 0.59 & (15, 19) & 3.37  \\
(15, 30) & 2.4 & (12, 31) & 1.0 & (14, 21) & 0.58 & (15, 27) & 3.37  \\
    \toprule
     \multicolumn{8}{c}{\textbf{OLMo-2 7B DPO}} \\
     \midrule
     \multicolumn{2}{c}{metaphor\_boolean} & \multicolumn{2}{c}{implicatures} & \multicolumn{2}{c}{object\_counting} & \multicolumn{2}{c}{snarks}\\
     \midrule
     Layer(s) &  Logit contrib. & Layer(s) &  Logit contrib. & Layer(s) &  Logit contrib. &  Layer(s) &  Logit contrib.\\ 
     \midrule
(14, 15) & 4.11 & (12, 26) & 2.17 & (14, 27) & 1.72 & (14, 30) & 4.47  \\
(14, 27) & 3.78 & (12, 14) & 2.16 & (14, 28) & 1.67 & (14, 14) & 4.44  \\
(14, 26) & 3.75 & (11, 14) & 2.15 & (14, 29) & 1.52 & (14, 31) & 4.44  \\
(9, 22) & 3.72 & (12, 27) & 2.07 & (14, 25) & 1.51 & (14, 28) & 4.43  \\
(14, 22) & 3.71 & (12, 29) & 2.0 & (14, 26) & 1.5 & (14, 29) & 4.4  \\
(14, 24) & 3.7 & (12, 30) & 2.0 & (14, 21) & 1.46 & (15, 16) & 4.36  \\
(14, 28) & 3.7 & (12, 24) & 1.95 & (12, 28) & 1.39 & (14, 27) & 4.34  \\
(14, 23) & 3.69 & (12, 28) & 1.95 & (12, 27) & 1.31 & (14, 19) & 4.31  \\
(9, 27) & 3.62 & (12, 25) & 1.93 & (14, 22) & 1.28 & (14, 26) & 4.3  \\
(9, 14) & 3.6 & (12, 21) & 1.92 & (12, 29) & 1.19 & (15, 17) & 4.3  \\

    \toprule
     \multicolumn{8}{c}{\textbf{OLMo-2 7B}} \\
     \midrule
     \multicolumn{2}{c}{metaphor\_boolean} & \multicolumn{2}{c}{implicatures} & \multicolumn{2}{c}{object\_counting} & \multicolumn{2}{c}{snarks}\\
     \midrule
     Layer(s) &  Logit contrib. & Layer(s) &  Logit contrib. & Layer(s) &  Logit contrib. &  Layer(s) &  Logit contrib.\\ 
     \midrule
(0, 2) & 0.55 & (1, 2) & 0.5 & (14, 27) & 1.72 & (2, 13) & 1.18  \\
(0, 29) & 0.38 & (1, 1) & 0.48 & (14, 28) & 1.67 & (2, 4) & 1.07  \\
(0, 10) & 0.37 & (1, 31) & 0.48 & (14, 29) & 1.52 & (2, 25) & 1.06  \\
(0, 6) & 0.36 & (1, 30) & 0.44 & (14, 25) & 1.51 & (2, 14) & 1.05  \\
(0, 7) & 0.34 & (1, 27) & 0.4 & (14, 26) & 1.5 & (2, 12) & 1.04  \\
(0, 8) & 0.34 & (1, 25) & 0.39 & (14, 21) & 1.46 & (2, 23) & 1.04  \\
(3, 13) & 0.34 & (2, 25) & 0.38 & (12, 28) & 1.39 & (2, 24) & 1.04  \\
(0, 5) & 0.32 & (2, 2) & 0.36 & (12, 27) & 1.31 & (2, 26) & 1.01  \\
(0, 30) & 0.31 & (2, 31) & 0.36 & (14, 22) & 1.28 & (2, 15) & 1.0  \\
(0, 1) & 0.3 & (1, 17) & 0.34 & (12, 29) & 1.19 & (2, 8) & 0.98  \\

    \bottomrule
     \end{tabular}
     \caption{Numerical logit contributions of OLMo-2 7B models for various patching configurations. For each model, the 10 highest-contributing combinations are shown.}
     \label{tab:app:bb-logit-olmo7b-num}
 \end{table*}

 \begin{table*}
     \centering
     \small
     \begin{tabular}{cccccccc}

     \toprule
     \multicolumn{8}{c}{\textbf{BigBench Tasks - Rank 7B}} \\
     \toprule
     \multicolumn{8}{c}{\textbf{OLMo-2 7B SFT}} \\
     \midrule
     \multicolumn{2}{c}{metaphor\_boolean} & \multicolumn{2}{c}{implicatures} & \multicolumn{2}{c}{object\_counting} & \multicolumn{2}{c}{snarks}\\
     \midrule
     Layer(s) &  Rank contrib. & Layer(s) &  Rank contrib. & Layer(s) &  Rank contrib. &  Layer(s) &  Rank contrib.\\ 
     \midrule
(6, 28) & 0.04 & (21, 24) & 0.01 & (0, 0) & 0.0 & (3, 11) & 0.14  \\
(6, 29) & 0.04 & (21, 25) & 0.01 & (0, 1) & 0.0 & (8, 10) & 0.12  \\
(6, 6) & 0.03 & (0, 0) & 0.0 & (0, 2) & 0.0 & (10, 24) & 0.11  \\
(6, 9) & 0.03 & (0, 1) & 0.0 & (0, 3) & 0.0 & (10, 25) & 0.11  \\
(6, 23) & 0.03 & (0, 2) & 0.0 & (0, 4) & 0.0 & (10, 10) & 0.1  \\
(6, 27) & 0.03 & (0, 3) & 0.0 & (0, 5) & 0.0 & (10, 26) & 0.1  \\
(6, 30) & 0.03 & (0, 4) & 0.0 & (0, 6) & 0.0 & (10, 27) & 0.1  \\
(6, 31) & 0.03 & (0, 5) & 0.0 & (0, 7) & 0.0 & (10, 28) & 0.1  \\
(13, 14) & 0.03 & (0, 6) & 0.0 & (0, 8) & 0.0 & (10, 29) & 0.1  \\
(21, 21) & 0.03 & (0, 7) & 0.0 & (0, 9) & 0.0 & (10, 30) & 0.1  \\
    \toprule
     \multicolumn{8}{c}{\textbf{OLMo-2 7B DPO}} \\
     \midrule
     \multicolumn{2}{c}{metaphor\_boolean} & \multicolumn{2}{c}{implicatures} & \multicolumn{2}{c}{object\_counting} & \multicolumn{2}{c}{snarks}\\
     \midrule
     Layer(s) &  Rank contrib. & Layer(s) &  Rank contrib. & Layer(s) &  Rank contrib. &  Layer(s) &  Rank contrib.\\ 
     \midrule
(13, 15) & 0.03 & (21, 24) & 0.01 & (0, 12) & 0.01 & (15, 19) & 0.07  \\
(1, 16) & 0.02 & (0, 0) & 0.0 & (2, 3) & 0.01 & (16, 17) & 0.07  \\
(6, 9) & 0.02 & (0, 1) & 0.0 & (2, 5) & 0.01 & (16, 18) & 0.07  \\
(8, 16) & 0.02 & (0, 2) & 0.0 & (2, 6) & 0.01 & (17, 17) & 0.07  \\
(13, 14) & 0.02 & (0, 3) & 0.0 & (2, 8) & 0.01 & (17, 29) & 0.07  \\
(0, 16) & 0.01 & (0, 4) & 0.0 & (2, 12) & 0.01 & (17, 30) & 0.07  \\
(1, 8) & 0.01 & (0, 5) & 0.0 & (2, 13) & 0.01 & (17, 31) & 0.07  \\
(1, 9) & 0.01 & (0, 6) & 0.0 & (5, 8) & 0.01 & (15, 20) & 0.06  \\
(1, 10) & 0.01 & (0, 7) & 0.0 & (0, 0) & 0.0 & (15, 15) & 0.05  \\
(1, 11) & 0.01 & (0, 8) & 0.0 & (0, 1) & 0.0 & (15, 18) & 0.05  \\

    \toprule
     \multicolumn{8}{c}{\textbf{OLMo-2 7B}} \\
     \midrule
     \multicolumn{2}{c}{metaphor\_boolean} & \multicolumn{2}{c}{implicatures} & \multicolumn{2}{c}{object\_counting} & \multicolumn{2}{c}{snarks}\\
     \midrule
     Layer(s) &  Rank contrib. & Layer(s) &  Rank contrib. & Layer(s) &  Rank contrib. &  Layer(s) &  Rank contrib.\\ 
     \midrule
(6, 11) & 0.05 & (3, 6) & 0.01 & (0, 12) & 0.01 & (7, 19) & 0.09  \\
(6, 9) & 0.04 & (13, 16) & 0.01 & (2, 3) & 0.01 & (6, 29) & 0.08  \\
(1, 15) & 0.03 & (0, 0) & 0.0 & (2, 5) & 0.01 & (7, 18) & 0.08  \\
(1, 16) & 0.03 & (0, 1) & 0.0 & (2, 6) & 0.01 & (7, 20) & 0.08  \\
(1, 17) & 0.03 & (0, 2) & 0.0 & (2, 8) & 0.01 & (7, 22) & 0.08  \\
(1, 18) & 0.03 & (0, 3) & 0.0 & (2, 12) & 0.01 & (16, 17) & 0.08  \\
(2, 3) & 0.03 & (0, 4) & 0.0 & (2, 13) & 0.01 & (6, 6) & 0.07  \\
(5, 11) & 0.03 & (0, 5) & 0.0 & (5, 8) & 0.01 & (6, 17) & 0.07  \\
(9, 22) & 0.03 & (0, 6) & 0.0 & (0, 0) & 0.0 & (6, 21) & 0.07  \\
(1, 1) & 0.02 & (0, 7) & 0.0 & (0, 1) & 0.0 & (6, 30) & 0.07  \\

    \bottomrule
     \end{tabular}
     \caption{Numerical rank contributions of OLMo-2 7B models for various patching configurations. For each model, the 10 highest-contributing combinations are shown.}
     \label{tab:app:bb-rank-olmo7b-num}
 \end{table*}

 \begin{table*}
     \centering
     \small
     \begin{tabular}{cccccccc}

     \toprule
     \multicolumn{8}{c}{\textbf{BigBench Tasks - Rank 1B}} \\
     \toprule
     \multicolumn{8}{c}{\textbf{OLMo-2 1B SFT}} \\
     \midrule
     \multicolumn{2}{c}{metaphor\_boolean} & \multicolumn{2}{c}{implicatures} & \multicolumn{2}{c}{object\_counting} & \multicolumn{2}{c}{snarks}\\
     \midrule
     Layer(s) &  Rank contrib. & Layer(s) &  Rank contrib. & Layer(s) &  Rank contrib. &  Layer(s) &  Rank contrib.\\ 
     \midrule
(0, 0) & 0.0 & (0, 0) & 0.0 & (0, 0) & 0.0 & (13, 13) & 0.0  \\
(0, 1) & 0.0 & (0, 1) & 0.0 & (0, 1) & 0.0 & (13, 14) & 0.0  \\
(0, 2) & 0.0 & (0, 2) & 0.0 & (0, 2) & 0.0 & (13, 15) & 0.0  \\
(0, 3) & 0.0 & (0, 3) & 0.0 & (0, 3) & 0.0 & (15, 15) & 0.0  \\
(0, 4) & 0.0 & (0, 4) & 0.0 & (0, 4) & 0.0 & (14, 14) & -0.01  \\
(0, 5) & 0.0 & (0, 5) & 0.0 & (0, 5) & 0.0 & (14, 15) & -0.01  \\
(0, 6) & 0.0 & (0, 6) & 0.0 & (0, 6) & 0.0 & (9, 9) & -0.02  \\
(0, 7) & 0.0 & (0, 7) & 0.0 & (0, 7) & 0.0 & (9, 13) & -0.02  \\
(0, 8) & 0.0 & (0, 8) & 0.0 & (0, 8) & 0.0 & (9, 15) & -0.02  \\
(0, 9) & 0.0 & (0, 9) & 0.0 & (0, 9) & 0.0 & (9, 10) & -0.03  \\
    \toprule
     \multicolumn{8}{c}{\textbf{OLMo-2 1B DPO}} \\
     \midrule
     \multicolumn{2}{c}{metaphor\_boolean} & \multicolumn{2}{c}{implicatures} & \multicolumn{2}{c}{object\_counting} & \multicolumn{2}{c}{snarks}\\
     \midrule
     Layer(s) &  Rank contrib. & Layer(s) &  Rank contrib. & Layer(s) &  Rank contrib. &  Layer(s) &  Rank contrib.\\ 
     \midrule
(1, 3) & 0.01 & (0, 0) & 0.0 & (0, 0) & 0.0 & (10, 10) & 0.01  \\
(0, 0) & 0.0 & (0, 1) & 0.0 & (0, 1) & 0.0 & (10, 11) & 0.01  \\
(0, 1) & 0.0 & (0, 2) & 0.0 & (0, 2) & 0.0 & (10, 12) & 0.01  \\
(0, 2) & 0.0 & (0, 3) & 0.0 & (0, 3) & 0.0 & (10, 13) & 0.01  \\
(0, 3) & 0.0 & (0, 4) & 0.0 & (0, 4) & 0.0 & (10, 14) & 0.01  \\
(0, 4) & 0.0 & (0, 5) & 0.0 & (0, 5) & 0.0 & (10, 15) & 0.01  \\
(0, 5) & 0.0 & (0, 6) & 0.0 & (0, 6) & 0.0 & (9, 9) & 0.0  \\
(0, 6) & 0.0 & (0, 7) & 0.0 & (0, 7) & 0.0 & (9, 10) & 0.0  \\
(0, 7) & 0.0 & (0, 8) & 0.0 & (0, 8) & 0.0 & (9, 11) & 0.0  \\
(0, 8) & 0.0 & (0, 9) & 0.0 & (0, 9) & 0.0 & (9, 12) & 0.0  \\

    \toprule
     \multicolumn{8}{c}{\textbf{OLMo-2 1B}} \\
     \midrule
     \multicolumn{2}{c}{metaphor\_boolean} & \multicolumn{2}{c}{implicatures} & \multicolumn{2}{c}{object\_counting} & \multicolumn{2}{c}{snarks}\\
     \midrule
     Layer(s) &  Rank contrib. & Layer(s) &  Rank contrib. & Layer(s) &  Rank contrib. &  Layer(s) &  Rank contrib.\\ 
     \midrule
(4, 4) & 0.13 & (0, 0) & 0.0 & (0, 0) & 0.0 & (10, 14) & 0.1  \\
(4, 5) & 0.13 & (0, 1) & 0.0 & (0, 1) & 0.0 & (3, 6) & 0.08  \\
(4, 11) & 0.13 & (0, 2) & 0.0 & (0, 2) & 0.0 & (10, 10) & 0.08  \\
(4, 14) & 0.13 & (0, 3) & 0.0 & (0, 3) & 0.0 & (10, 15) & 0.08  \\
(4, 15) & 0.13 & (0, 4) & 0.0 & (0, 4) & 0.0 & (2, 9) & 0.07  \\
(4, 8) & 0.12 & (0, 5) & 0.0 & (0, 5) & 0.0 & (2, 14) & 0.07  \\
(4, 12) & 0.12 & (0, 6) & 0.0 & (0, 6) & 0.0 & (10, 11) & 0.07  \\
(4, 13) & 0.12 & (0, 7) & 0.0 & (0, 7) & 0.0 & (10, 12) & 0.07  \\
(4, 10) & 0.11 & (0, 8) & 0.0 & (0, 8) & 0.0 & (10, 13) & 0.07  \\
(4, 9) & 0.1 & (0, 9) & 0.0 & (0, 9) & 0.0 & (2, 2) & 0.06  \\

    \bottomrule
     \end{tabular}
     \caption{Numerical rank contributions of OLMo-2 1B models for various patching configurations. For each model, the 10 highest-contributing combinations are shown.}
     \label{tab:app:bb-rank-olmo1b-num}
 \end{table*}

 \begin{table*}
     \centering
     \small
     \begin{tabular}{cccccccc}

     \toprule
     \multicolumn{8}{c}{\textbf{Contrastive Tasks - Rank 7B}} \\
     \toprule
     \multicolumn{8}{c}{\textbf{OLMo-2 7B SFT}} \\
     \midrule
     \multicolumn{2}{c}{adj: ant} & \multicolumn{2}{c}{adj: comp} & \multicolumn{2}{c}{anim: color} & \multicolumn{2}{c}{anim: can\_fly}\\
     \midrule
     Layer(s) &  Rank contrib. & Layer(s) &  Rank contrib. & Layer(s) &  Rank contrib. &  Layer(s) &  Rank contrib.\\ 
     \midrule
 (7, 18) & 0.08 & (6, 8) & 0.13 & (6, 8) & 0.01 & (0, 13) & 0.06  \\
 (8, 15) & 0.08 & (7, 8) & 0.13 & (9, 13) & 0.01 & (0, 4) & 0.03  \\
 (8, 16) & 0.08 & (5, 9) & 0.12 & (9, 14) & 0.01 & (0, 5) & 0.03  \\
(8, 17) & 0.08 & (6, 9) & 0.12 & (2, 16) & -0.0 & (0, 6) & 0.03  \\
(9, 24) & 0.08 & (7, 9) & 0.12 & (2, 17) & -0.0 & (0, 0) & 0.02  \\
(4, 14) & 0.07 & (8, 8) & 0.12 & (3, 15) & -0.0 & (0, 7) & 0.02  \\
(4, 19) & 0.07 & (8, 27) & 0.12 & (4, 13) & -0.0 & (0, 12) & 0.02  \\
(7, 17) & 0.07 & (8, 28) & 0.12 & (4, 14) & -0.0 & (0, 31) & 0.02  \\
(8, 18) & 0.07 & (8, 30) & 0.12 & (5, 16) & -0.0 & (1, 13) & 0.02  \\
(4, 15) & 0.06 & (8, 31) & 0.12 & (6, 6) & -0.0 & (0, 1) & 0.01  \\

    \toprule
     \multicolumn{8}{c}{\textbf{OLMo-2 7B DPO}} \\
     \midrule
     \multicolumn{2}{c}{adj: ant} & \multicolumn{2}{c}{adj: comp} & \multicolumn{2}{c}{anim: color} & \multicolumn{2}{c}{anim: can\_fly}\\
     \midrule
     Layer(s) &  Rank contrib. & Layer(s) &  Rank contrib. & Layer(s) &  Rank contrib. &  Layer(s) &  Rank contrib.\\ 
     \midrule
 (8, 26) & 0.1 & (6, 9) & 0.12 & (8, 14) & 0.02 & (0, 13) & 0.04  \\
(8, 25) & 0.09 & (7, 9) & 0.12 & (9, 13) & 0.02 & (0, 5) & 0.03  \\
(1, 17) & 0.07 & (5, 9) & 0.11 & (9, 14) & 0.02 & (0, 0) & 0.02  \\
(7, 27) & 0.07 & (8, 8) & 0.11 & (5, 18) & 0.01 & (0, 4) & 0.02  \\
(7, 28) & 0.07 & (8, 31) & 0.11 & (6, 8) & 0.01 & (0, 12) & 0.02  \\
(8, 15) & 0.07 & (4, 19) & 0.1 & (6, 14) & 0.01 & (0, 26) & 0.02  \\
(8, 16) & 0.07 & (7, 8) & 0.1 & (6, 16) & 0.01 & (0, 28) & 0.02  \\
(8, 17) & 0.07 & (8, 9) & 0.1 & (6, 18) & 0.01 & (0, 31) & 0.02  \\
(8, 18) & 0.07 & (9, 9) & 0.1 & (9, 11) & 0.01 & (0, 1) & 0.01  \\
(8, 24) & 0.07 & (9, 28) & 0.1 & (9, 12) & 0.01 & (0, 3) & 0.01  \\

    \toprule
     \multicolumn{8}{c}{\textbf{OLMo-2 7B}} \\
     \midrule
     \multicolumn{2}{c}{adj: ant} & \multicolumn{2}{c}{adj: comp} & \multicolumn{2}{c}{anim: color} & \multicolumn{2}{c}{anim: can\_fly}\\
     \midrule
     Layer(s) &  Rank contrib. & Layer(s) &  Rank contrib. & Layer(s) &  Rank contrib. &  Layer(s) &  Rank contrib.\\ 
     \midrule
 (7, 18) & 0.1 & (3, 18) & 0.11 & (9, 12) & 0.01 & (0, 11) & 0.06  \\
 (2, 24) & 0.08 & (7, 10) & 0.11 & (9, 13) & 0.01 & (0, 10) & 0.03  \\
(2, 25) & 0.08 & (3, 12) & 0.1 & (9, 14) & 0.01 & (0, 12) & 0.03  \\
(4, 18) & 0.08 & (4, 12) & 0.1 & (2, 16) & 0.0 & (0, 6) & 0.02  \\
(7, 17) & 0.08 & (4, 18) & 0.1 & (4, 5) & 0.0 & (0, 0) & 0.01  \\
(1, 3) & 0.07 & (5, 11) & 0.1 & (5, 5) & 0.0 & (0, 5) & 0.01  \\
(2, 16) & 0.07 & (5, 17) & 0.1 & (5, 12) & 0.0 & (0, 7) & 0.01  \\
(2, 27) & 0.07 & (5, 18) & 0.1 & (5, 28) & 0.0 & (0, 9) & 0.01  \\
(7, 25) & 0.07 & (5, 27) & 0.1 & (5, 31) & 0.0 & (0, 13) & 0.01  \\
(7, 29) & 0.07 & (6, 18) & 0.1 & (6, 6) & 0.0 & (0, 14) & 0.01  \\
    \bottomrule
     \end{tabular}
     \caption{Numerical rank contributions of OLMo-2 7B models for various patching configurations. For each model, the 10 highest-contributing combinations are shown.}
     \label{tab:app:contrastive-rank-olmo7b-num}
 \end{table*}

 \begin{table*}
     \centering
     \small
     \begin{tabular}{cccccccc}

     \toprule
     \multicolumn{8}{c}{\textbf{Contrastive Tasks - Logit 1B}} \\
     \toprule
     \multicolumn{8}{c}{\textbf{OLMo-2 1B SFT}} \\
     \midrule
     \multicolumn{2}{c}{adj: ant} & \multicolumn{2}{c}{adj: comp} & \multicolumn{2}{c}{anim: color} & \multicolumn{2}{c}{anim: can\_fly}\\
     \midrule
     Layer(s) &  Logit contrib. & Layer(s) &  Logit contrib. & Layer(s) &  Logit contrib. &  Layer(s) &  Logit contrib.\\ 
     \midrule
(3, 15) & 1.73 & (5, 11) & 1.46 & (3, 11) & 3.15 & (0, 11) & 4.18  \\
(3, 14) & 1.47 & (4, 11) & 1.42 & (5, 13) & 3.09 & (3, 11) & 3.95  \\
(7, 15) & 1.44 & (7, 11) & 1.39 & (5, 12) & 3.02 & (0, 14) & 3.8  \\
(3, 13) & 1.39 & (3, 13) & 1.37 & (7, 13) & 2.99 & (0, 13) & 3.7  \\
(5, 15) & 1.29 & (6, 11) & 1.37 & (6, 13) & 2.93 & (0, 15) & 3.45  \\
(8, 15) & 1.13 & (5, 15) & 1.36 & (6, 12) & 2.86 & (0, 12) & 3.34  \\
(6, 15) & 1.08 & (3, 11) & 1.35 & (7, 12) & 2.85 & (2, 11) & 3.32  \\
(7, 13) & 1.02 & (8, 11) & 1.32 & (3, 13) & 2.84 & (4, 11) & 3.3  \\
(7, 14) & 0.95 & (9, 11) & 1.32 & (4, 11) & 2.66 & (0, 10) & 3.22  \\
(9, 15) & 0.94 & (3, 15) & 1.29 & (3, 12) & 2.63 & (5, 11) & 3.13  \\

    \toprule
     \multicolumn{8}{c}{\textbf{OLMo-2 1B DPO}} \\
     \midrule
     \multicolumn{2}{c}{adj: ant} & \multicolumn{2}{c}{adj: comp} & \multicolumn{2}{c}{anim: color} & \multicolumn{2}{c}{anim: can\_fly}\\
     \midrule
     Layer(s) &  Logit contrib. & Layer(s) &  Logit contrib. & Layer(s) &  Logit contrib. &  Layer(s) &  Logit contrib.\\ 
     \midrule
(3, 15) & 2.24 & (3, 13) & 2.11 & (5, 13) & 3.04 & (0, 11) & 3.73  \\
(3, 13) & 1.84 & (3, 12) & 2.09 & (7, 13) & 2.99 & (3, 11) & 3.44  \\
(0, 15) & 1.79 & (4, 11) & 1.99 & (6, 13) & 2.93 & (0, 13) & 3.39  \\
(3, 14) & 1.79 & (5, 11) & 1.99 & (5, 12) & 2.77 & (0, 14) & 3.31  \\
(7, 15) & 1.7 & (0, 11) & 1.98 & (3, 11) & 2.73 & (0, 15) & 3.08  \\
(5, 15) & 1.49 & (9, 11) & 1.95 & (6, 12) & 2.64 & (0, 12) & 2.97  \\
(8, 15) & 1.44 & (6, 11) & 1.94 & (7, 12) & 2.59 & (5, 11) & 2.84  \\
(9, 15) & 1.3 & (3, 11) & 1.93 & (8, 13) & 2.53 & (4, 11) & 2.83  \\
(4, 15) & 1.23 & (2, 11) & 1.91 & (7, 14) & 2.45 & (2, 11) & 2.8  \\
(6, 15) & 1.2 & (7, 11) & 1.91 & (3, 13) & 2.38 & (6, 11) & 2.76  \\

    \toprule
     \multicolumn{8}{c}{\textbf{OLMo-2 1B}} \\
     \midrule
     \multicolumn{2}{c}{adj: ant} & \multicolumn{2}{c}{adj: comp} & \multicolumn{2}{c}{anim: color} & \multicolumn{2}{c}{anim: can\_fly}\\
     \midrule
     Layer(s) &  Logit contrib. & Layer(s) &  Logit contrib. & Layer(s) &  Logit contrib. &  Layer(s) &  Logit contrib.\\ 
     \midrule
(3, 3) & -5.42 & (3, 5) & -5.94 & (6, 14) & 1.15 & (13, 14) & -0.4  \\
(1, 3) & -5.6 & (6, 15) & -5.96 & (6, 13) & 1.08 & (13, 15) & -0.4  \\
(0, 3) & -6.0 & (1, 5) & -6.02 & (6, 15) & 1.08 & (11, 15) & -0.59  \\
(2, 3) & -6.15 & (5, 5) & -6.05 & (6, 12) & 0.75 & (12, 15) & -0.64  \\
(5, 5) & -6.17 & (2, 5) & -6.22 & (6, 7) & 0.67 & (12, 14) & -0.68  \\
(1, 5) & -6.26 & (9, 15) & -6.22 & (6, 6) & 0.55 & (6, 9) & -0.71  \\
(3, 5) & -6.29 & (10, 15) & -6.22 & (6, 11) & 0.5 & (13, 13) & -0.71  \\
(4, 5) & -6.33 & (11, 15) & -6.24 & (6, 10) & 0.46 & (15, 15) & -0.74  \\
(11, 15) & -6.46 & (4, 5) & -6.26 & (7, 15) & 0.44 & (11, 14) & -0.76  \\
(13, 15) & -6.49 & (7, 15) & -6.29 & (7, 14) & 0.37 & (12, 12) & -0.77  \\
    \bottomrule
     \end{tabular}
     \caption{Numerical logit contributions of OLMo-2 1B models for various patching configurations. For each model, the 10 highest-contributing combinations are shown.}
     \label{tab:app:contrastive-logit-olmo1b-num}
 \end{table*}

\begin{table*}
     \centering
     \small
     \begin{tabular}{cccccccc}

     \toprule
     \multicolumn{8}{c}{\textbf{Contrastive Tasks - Logit 7B}} \\
     \toprule
     \multicolumn{8}{c}{\textbf{OLMo-2 7B SFT}} \\
     \midrule
     \multicolumn{2}{c}{adj: ant} & \multicolumn{2}{c}{adj: comp} & \multicolumn{2}{c}{anim: color} & \multicolumn{2}{c}{anim: can\_fly}\\
     \midrule
     Layer(s) &  Logit contrib. & Layer(s) &  Logit contrib. & Layer(s) &  Logit contrib. &  Layer(s) &  Logit contrib.\\ 
     \midrule
(1, 17) & -0.84 & (4, 16) & -1.75 & (0, 31) & 0.75 & (0, 14) & 4.11  \\
(1, 16) & -0.91 & (6, 16) & -1.93 & (0, 0) & 0.62 & (1, 14) & 3.64  \\
(3, 16) & -1.01 & (7, 16) & -1.97 & (1, 31) & 0.19 & (0, 1) & 2.66  \\
(3, 17) & -1.07 & (8, 16) & -2.0 & (0, 1) & 0.11 & (0, 31) & 2.6  \\
(8, 17) & -1.07 & (1, 16) & -2.01 & (0, 27) & 0.05 & (1, 15) & 2.46  \\
(4, 16) & -1.11 & (3, 16) & -2.04 & (0, 4) & -0.02 & (0, 6) & 2.44  \\
(8, 16) & -1.13 & (5, 16) & -2.29 & (1, 1) & -0.04 & (0, 28) & 2.35  \\
(0, 17) & -1.16 & (9, 16) & -2.29 & (0, 28) & -0.1 & (0, 13) & 2.26  \\
(4, 17) & -1.16 & (4, 14) & -2.3 & (0, 30) & -0.34 & (0, 15) & 2.26  \\
(6, 16) & -1.2 & (11, 16) & -2.3 & (0, 3) & -0.35 & (0, 0) & 2.25  \\

    \toprule
     \multicolumn{8}{c}{\textbf{OLMo-2 7B DPO}} \\
     \midrule
     \multicolumn{2}{c}{adj: ant} & \multicolumn{2}{c}{adj: comp} & \multicolumn{2}{c}{anim: color} & \multicolumn{2}{c}{anim: can\_fly}\\
     \midrule
     Layer(s) &  Logit contrib. & Layer(s) &  Logit contrib. & Layer(s) &  Logit contrib. &  Layer(s) &  Logit contrib.\\ 
     \midrule
(1, 16) & 0.87 & (4, 16) & -0.12 & (0, 0) & 1.91 & (0, 14) & 5.36  \\
(3, 16) & 0.81 & (1, 16) & -0.27 & (0, 28) & 1.85 & (1, 14) & 4.89  \\
(1, 17) & 0.7 & (3, 16) & -0.32 & (0, 31) & 1.76 & (0, 1) & 4.41  \\
(8, 16) & 0.64 & (8, 16) & -0.33 & (0, 27) & 1.74 & (0, 0) & 4.05  \\
(4, 16) & 0.56 & (7, 16) & -0.35 & (0, 29) & 1.26 & (1, 15) & 3.88  \\
(3, 17) & 0.46 & (6, 16) & -0.44 & (0, 1) & 1.21 & (0, 15) & 3.8  \\
(0, 16) & 0.45 & (0, 16) & -0.58 & (0, 26) & 1.11 & (0, 6) & 3.63  \\
(0, 17) & 0.45 & (11, 16) & -0.67 & (1, 31) & 1.07 & (0, 31) & 3.59  \\
(8, 17) & 0.4 & (9, 16) & -0.7 & (0, 30) & 0.97 & (1, 1) & 3.44  \\
(7, 16) & 0.36 & (4, 14) & -0.75 & (0, 9) & 0.94 & (0, 5) & 3.4  \\

    \toprule
     \multicolumn{8}{c}{\textbf{OLMo-2 7B}} \\
     \midrule
     \multicolumn{2}{c}{adj: ant} & \multicolumn{2}{c}{adj: comp} & \multicolumn{2}{c}{anim: color} & \multicolumn{2}{c}{anim: can\_fly}\\
     \midrule
     Layer(s) &  Logit contrib. & Layer(s) &  Logit contrib. & Layer(s) &  Logit contrib. &  Layer(s) &  Logit contrib.\\ 
     \midrule
(0, 0) & -2.3 & (0, 0) & -3.0 & (31, 31) & -0.43 & (31, 31) & 0.66  \\
(1, 3) & -6.21 & (7, 19) & -4.92 & (2, 31) & -1.97 & (0, 6) & 0.18  \\
(3, 3) & -6.23 & (7, 20) & -4.93 & (0, 1) & -2.16 & (0, 10) & -0.18  \\
(2, 3) & -6.34 & (7, 21) & -4.93 & (5, 31) & -2.23 & (0, 7) & -0.2  \\
(0, 3) & -6.51 & (9, 21) & -5.03 & (0, 0) & -2.4 & (0, 8) & -0.28  \\
(5, 7) & -6.51 & (8, 21) & -5.04 & (30, 31) & -2.41 & (0, 9) & -0.28  \\
(6, 7) & -6.51 & (0, 1) & -5.07 & (0, 31) & -2.43 & (0, 11) & -0.36  \\
(4, 7) & -6.53 & (6, 21) & -5.07 & (1, 31) & -2.5 & (0, 1) & -0.41  \\
(5, 5) & -6.54 & (11, 20) & -5.08 & (30, 30) & -2.68 & (0, 31) & -0.59  \\
(7, 7) & -6.58 & (11, 21) & -5.08 & (1, 1) & -2.79 & (0, 5) & -0.69  \\
    \bottomrule
     \end{tabular}
     \caption{Numerical logit contributions of OLMo-2 7B models for various patching configurations. For each model, the 10 highest-contributing combinations are shown.}
     \label{tab:app:contrastive-logit-olmo7b-num}
 \end{table*}

\begin{table*}
\centering
\small
\begin{tabular}{cccccccc}
\toprule
\multicolumn{8}{c}{\textbf{Contrastive Tasks - Logits 1B}} \\
\toprule\multicolumn{8}{c}{\textbf{OLMo-2 1B SFT}} \\
\midrule\multicolumn{2}{c}{adj: ant} & \multicolumn{2}{c}{adj: comp} & \multicolumn{2}{c}{anim: color} & \multicolumn{2}{c}{anim: can\_fly}\\
\midrule
Layer(s) &  Logits contrib. & Layer(s) &  Logits contrib. & Layer(s) &  Logits contrib. &  Layer(s) &  Logits contrib.\\
\midrule
(3, 14, 15) & 1.97 & (2, 5, 11) & 1.69 & (3, 6, 12) & 4.4 & (0, 11, 14) & 4.42  \\
(3, 13, 15) & 1.87 & (4, 5, 11) & 1.62 & (3, 11, 13) & 4.36 & (0, 6, 11) & 4.29  \\
(3, 11, 15) & 1.67 & (3, 13, 15) & 1.61 & (3, 6, 13) & 4.25 & (0, 11, 15) & 4.25  \\
(3, 12, 15) & 1.63 & (2, 4, 11) & 1.58 & (1, 6, 12) & 4.09 & (3, 11, 14) & 4.22  \\
(7, 13, 15) & 1.5 & (3, 14, 15) & 1.57 & (3, 11, 14) & 4.09 & (0, 7, 11) & 4.16  \\
(5, 7, 15) & 1.49 & (5, 9, 11) & 1.53 & (3, 11, 12) & 4.04 & (3, 5, 11) & 4.14  \\
(7, 14, 15) & 1.45 & (5, 10, 11) & 1.5 & (3, 7, 12) & 4.02 & (0, 5, 11) & 4.06  \\
(5, 8, 15) & 1.41 & (1, 6, 11) & 1.49 & (3, 7, 13) & 3.98 & (0, 10, 11) & 4.04  \\
(3, 13, 14) & 1.39 & (4, 6, 11) & 1.48 & (3, 10, 13) & 3.98 & (0, 11, 13) & 4.04  \\
(7, 11, 15) & 1.36 & (7, 9, 11) & 1.48 & (4, 6, 12) & 3.93 & (3, 4, 11) & 4.02  \\
\toprule\multicolumn{8}{c}{\textbf{OLMo-2 1B DPO}} \\
\midrule\multicolumn{2}{c}{adj: ant} & \multicolumn{2}{c}{adj: comp} & \multicolumn{2}{c}{anim: color} & \multicolumn{2}{c}{anim: can\_fly}\\
\midrule
Layer(s) &  Logits contrib. & Layer(s) &  Logits contrib. & Layer(s) &  Logits contrib. &  Layer(s) &  Logits contrib.\\
\midrule
(3, 14, 15) & 2.51 & (3, 13, 15) & 2.19 & (3, 11, 13) & 4.56 & (3, 5, 11) & 3.76  \\
(3, 13, 15) & 2.44 & (2, 5, 11) & 2.11 & (3, 6, 13) & 4.14 & (0, 11, 14) & 3.74  \\
(3, 11, 15) & 2.22 & (2, 4, 11) & 2.09 & (3, 11, 14) & 4.13 & (0, 11, 15) & 3.73  \\
(3, 12, 15) & 2.21 & (3, 14, 15) & 2.07 & (3, 6, 12) & 4.07 & (0, 6, 11) & 3.69  \\
(0, 14, 15) & 1.88 & (4, 5, 11) & 2.06 & (2, 11, 13) & 3.95 & (3, 4, 11) & 3.68  \\
(2, 3, 15) & 1.85 & (5, 9, 11) & 2.06 & (4, 6, 13) & 3.93 & (0, 11, 13) & 3.64  \\
(7, 13, 15) & 1.8 & (4, 6, 11) & 2.02 & (3, 11, 12) & 3.92 & (2, 4, 11) & 3.56  \\
(3, 13, 14) & 1.73 & (5, 10, 11) & 2.01 & (3, 10, 13) & 3.9 & (0, 7, 11) & 3.55  \\
(7, 14, 15) & 1.7 & (7, 9, 11) & 2.01 & (3, 7, 13) & 3.89 & (3, 11, 14) & 3.48  \\
(0, 13, 15) & 1.69 & (5, 6, 11) & 2.0 & (0, 11, 13) & 3.86 & (0, 5, 11) & 3.47  \\
\toprule\multicolumn{8}{c}{\textbf{OLMo-2 1B}} \\
\midrule\multicolumn{2}{c}{adj: ant} & \multicolumn{2}{c}{adj: comp} & \multicolumn{2}{c}{anim: color} & \multicolumn{2}{c}{anim: can\_fly}\\
\midrule
Layer(s) &  Logits contrib. & Layer(s) &  Logits contrib. & Layer(s) &  Logits contrib. &  Layer(s) &  Logits contrib.\\
\midrule
(1, 2, 3) & -5.87 & (3, 4, 5) & -5.83 & (6, 7, 14) & 1.51 & (6, 8, 11) & 0.13  \\
(0, 1, 3) & -6.0 & (7, 9, 15) & -6.02 & (6, 7, 13) & 1.48 & (6, 7, 11) & 0.03  \\
(0, 2, 3) & -6.2 & (7, 10, 15) & -6.08 & (6, 7, 15) & 1.45 & (6, 8, 10) & 0.01  \\
(3, 4, 5) & -6.21 & (6, 14, 15) & -6.12 & (6, 7, 12) & 1.21 & (6, 9, 11) & -0.15  \\
(6, 9, 15) & -6.4 & (7, 11, 15) & -6.16 & (6, 10, 13) & 1.2 & (6, 9, 15) & -0.27  \\
(0, 1, 5) & -6.41 & (1, 2, 5) & -6.18 & (6, 10, 14) & 1.16 & (6, 7, 10) & -0.34  \\
(11, 12, 15) & -6.41 & (6, 9, 15) & -6.18 & (6, 12, 14) & 1.14 & (6, 9, 12) & -0.34  \\
(7, 9, 15) & -6.42 & (10, 11, 15) & -6.19 & (6, 13, 14) & 1.14 & (6, 9, 10) & -0.35  \\
(1, 4, 5) & -6.45 & (9, 11, 15) & -6.2 & (6, 13, 15) & 1.13 & (6, 9, 14) & -0.38  \\
(1, 2, 5) & -6.48 & (2, 4, 5) & -6.21 & (6, 14, 15) & 1.12 & (13, 14, 15) & -0.4  \\
\bottomrule
\end{tabular}
\caption{Numerical logit contributions of OLMo-2 1B models for various patching configurations. For each model, the 10 highest-contributing combinations are shown.}
\label{tab:app:logits-3-layer-triples}
\end{table*}

\begin{table*}
\centering
\small
\begin{tabular}{cccccccc}
\toprule
\multicolumn{8}{c}{\textbf{Contrastive Tasks - Rank 1B}} \\
\toprule\multicolumn{8}{c}{\textbf{OLMo-2 1B SFT}} \\
\midrule\multicolumn{2}{c}{adj: ant} & \multicolumn{2}{c}{adj: comp} & \multicolumn{2}{c}{anim: color} & \multicolumn{2}{c}{anim: can\_fly}\\
\midrule
Layer(s) &  Rank contrib. & Layer(s) &  Rank contrib. & Layer(s) &  Rank contrib. &  Layer(s) &  Rank contrib.\\
\midrule
(3, 14, 15) & 0.14 & (0, 14, 15) & 0.07 & (3, 5, 13) & 0.18 & (2, 3, 7) & 0.06  \\
(3, 11, 13) & 0.11 & (2, 3, 14) & 0.07 & (3, 6, 12) & 0.18 & (2, 6, 9) & 0.05  \\
(3, 11, 15) & 0.11 & (4, 6, 15) & 0.07 & (4, 5, 13) & 0.17 & (3, 6, 7) & 0.05  \\
(3, 13, 14) & 0.11 & (2, 3, 15) & 0.06 & (3, 5, 12) & 0.16 & (3, 7, 14) & 0.05  \\
(3, 13, 15) & 0.11 & (3, 14, 15) & 0.06 & (3, 6, 13) & 0.16 & (3, 7, 15) & 0.05  \\
(0, 14, 15) & 0.1 & (4, 5, 6) & 0.06 & (4, 6, 12) & 0.16 & (0, 3, 7) & 0.04  \\
(2, 3, 15) & 0.1 & (4, 6, 14) & 0.06 & (3, 11, 13) & 0.15 & (1, 6, 7) & 0.04  \\
(3, 11, 14) & 0.1 & (2, 3, 13) & 0.05 & (4, 6, 13) & 0.15 & (2, 4, 7) & 0.04  \\
(2, 3, 14) & 0.09 & (2, 5, 13) & 0.05 & (4, 7, 13) & 0.15 & (2, 4, 9) & 0.04  \\
(3, 5, 11) & 0.09 & (2, 5, 14) & 0.05 & (2, 6, 12) & 0.14 & (3, 4, 7) & 0.04  \\
\toprule\multicolumn{8}{c}{\textbf{OLMo-2 1B DPO}} \\
\midrule\multicolumn{2}{c}{adj: ant} & \multicolumn{2}{c}{adj: comp} & \multicolumn{2}{c}{anim: color} & \multicolumn{2}{c}{anim: can\_fly}\\
\midrule
Layer(s) &  Rank contrib. & Layer(s) &  Rank contrib. & Layer(s) &  Rank contrib. &  Layer(s) &  Rank contrib.\\
\midrule
(3, 14, 15) & 0.13 & (0, 14, 15) & 0.07 & (3, 5, 13) & 0.17 & (3, 4, 6) & 0.04  \\
(3, 13, 14) & 0.11 & (4, 6, 14) & 0.04 & (3, 6, 13) & 0.16 & (3, 4, 7) & 0.04  \\
(3, 13, 15) & 0.11 & (2, 3, 14) & 0.03 & (3, 11, 13) & 0.16 & (3, 7, 15) & 0.04  \\
(0, 14, 15) & 0.09 & (2, 3, 15) & 0.03 & (4, 6, 12) & 0.15 & (0, 5, 7) & 0.03  \\
(3, 11, 13) & 0.08 & (3, 14, 15) & 0.03 & (4, 6, 13) & 0.15 & (1, 6, 7) & 0.03  \\
(3, 11, 14) & 0.08 & (4, 5, 6) & 0.03 & (3, 5, 12) & 0.14 & (1, 7, 15) & 0.03  \\
(3, 11, 15) & 0.08 & (4, 6, 13) & 0.03 & (3, 6, 12) & 0.14 & (2, 3, 7) & 0.03  \\
(0, 3, 15) & 0.07 & (5, 6, 14) & 0.03 & (4, 5, 13) & 0.14 & (2, 4, 7) & 0.03  \\
(0, 13, 14) & 0.07 & (5, 6, 15) & 0.03 & (0, 13, 14) & 0.13 & (2, 4, 8) & 0.03  \\
(0, 13, 15) & 0.07 & (0, 1, 3) & 0.02 & (2, 5, 12) & 0.13 & (3, 4, 8) & 0.03  \\
\toprule\multicolumn{8}{c}{\textbf{OLMo-2 1B}} \\
\midrule\multicolumn{2}{c}{adj: ant} & \multicolumn{2}{c}{adj: comp} & \multicolumn{2}{c}{anim: color} & \multicolumn{2}{c}{anim: can\_fly}\\
\midrule
Layer(s) &  Rank contrib. & Layer(s) &  Rank contrib. & Layer(s) &  Rank contrib. &  Layer(s) &  Rank contrib.\\
\midrule
(1, 14, 15) & 0.16 & (1, 14, 15) & 0.08 & (6, 7, 13) & 0.16 & (5, 8, 10) & 0.22  \\
(0, 1, 14) & 0.14 & (6, 14, 15) & 0.07 & (6, 7, 14) & 0.16 & (4, 8, 10) & 0.21  \\
(0, 1, 13) & 0.13 & (7, 8, 9) & 0.07 & (6, 13, 14) & 0.16 & (4, 8, 11) & 0.21  \\
(0, 1, 15) & 0.13 & (7, 9, 14) & 0.07 & (6, 7, 12) & 0.15 & (4, 7, 10) & 0.19  \\
(6, 7, 9) & 0.12 & (7, 9, 15) & 0.07 & (6, 13, 15) & 0.15 & (5, 7, 10) & 0.19  \\
(6, 9, 14) & 0.12 & (7, 9, 10) & 0.06 & (6, 7, 8) & 0.14 & (3, 8, 11) & 0.18  \\
(1, 13, 15) & 0.11 & (7, 9, 11) & 0.06 & (6, 7, 11) & 0.14 & (4, 8, 14) & 0.18  \\
(7, 9, 14) & 0.11 & (7, 10, 11) & 0.06 & (6, 7, 15) & 0.14 & (5, 8, 14) & 0.18  \\
(0, 1, 6) & 0.1 & (0, 1, 14) & 0.05 & (6, 14, 15) & 0.13 & (3, 8, 14) & 0.17  \\
(1, 6, 15) & 0.1 & (1, 13, 14) & 0.05 & (6, 7, 9) & 0.12 & (4, 7, 8) & 0.17  \\
\bottomrule
\end{tabular}
\caption{Numerical rank contributions of OLMo-2 1B models for various patching configurations. For each model, the 10 highest-contributing combinations are shown.}
\label{tab:contrastive-tasks-rank-1b}
\end{table*}

\section{Task Inference Results}
\label{app:sec:instr_acc}

\begin{figure*}[htb!]
    \centering
    \includegraphics[width=0.9\linewidth,scale=0.8,trim={0 0cm 0 0cm},clip]{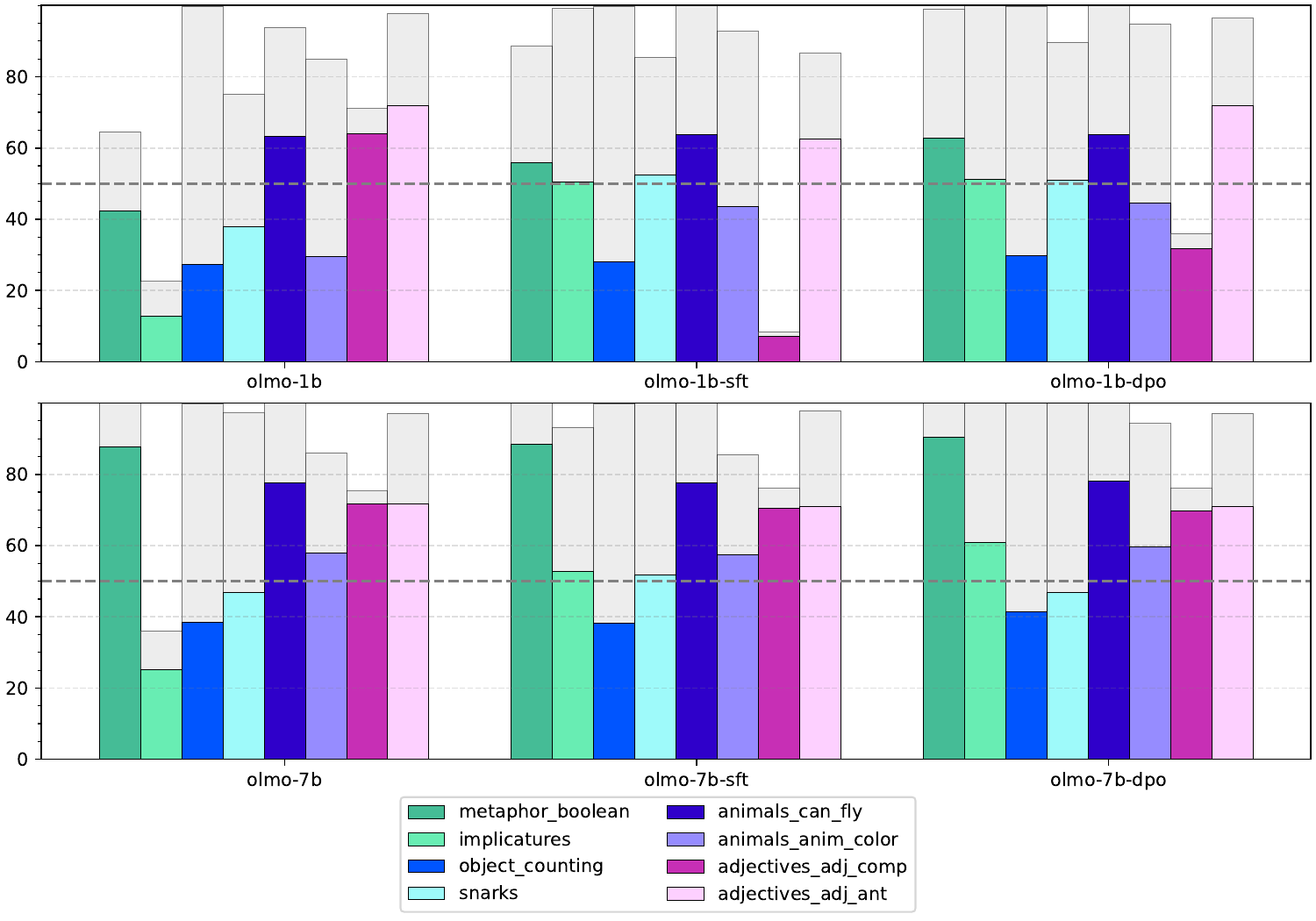}
    \caption{EMA scores (in color) for OLMo-1B and OLMo-7B models. Instruction Accuracy scores (gray) show that instruction-following abilities are present for most models and tasks (IA>50\%)} 
    \label{fig:basic_acc}
\end{figure*}

Some of our tasks are not easily evaluated by EMA metrics due to having multiple valid answer options. For example, the \textsc{Adj: Antonym} task expects the model to output the antonym of an adjective. However, multiple antonyms can exist for a given word, and oftentimes cannot be exhaustively listed. While the correctness of an output can still be measured through similarity metrics, this would not capture instruction following -- i.e. whether the model produced an appropriate response to the instruction, even if the answer is wrong.

Thus, for such tasks, we manually define a scope of accepted response types and implement an LLM-as-a-Judge strategy using GPT-5-nano \cite{chatgpt5-nano}. We prompt the judge to answer `yes' or `no' as to whether or not the response is appropriate given the instruction, regardless of answer correctness. If the judge answers `yes', we mark the answer as correct. We refer to this metric as \textsc{instructional accuracy} (IA). 

Table \ref{app:tab:llm_judge_prompts} shows the prompts provided for the judge for each task, with the manually-defined criteria in bold. We report the EMA and IA scores of our models in Figure \ref{fig:basic_acc}.

\begin{table*}
    \centering
    \small
    \begin{tabular}{p{5cm}p{10cm}}
    \toprule
         Task & Prompt \\
         \midrule
         adjective: comparative & `You will be given a task instruction and an answer someone wrote in response. Regardless of whether the answer is correct, determine whether the answer is \textbf{an English adjective in the comparative form.} Output either `yes' or `no'. Instruction: `\{\}' Answer: `\{\}'. Your decision:' \\
         \\ \hline \\
         adjective: antonym & `You will be given a task instruction and an answer someone wrote in response. Regardless of whether the answer is correct, determine whether the answer is \textbf{an English adjective in its standard declarative form.} Output either `yes' or `no'. Instruction: `\{\}' Answer: `\{\}'. Your decision:' \\
         \\ \hline \\
         animal: color & `You will be given a task instruction and an answer someone wrote in response. Regardless of whether the answer is correct, determine whether the answer is \textbf{the name of a color.} Output either `yes' or `no'. Instruction: `\{\}' Answer: `\{\}'. Your decision:' \\
         \bottomrule
    \end{tabular}
    \caption{LLM judge prompts for instructional accuracy (IA). The `instruction' field is filled with the query-independent task instruction and the `answer' field is filled with the inference model's output given a query. Thus, IA measures the appropriateness of a response to the instruction rather than the correctness of the answer.}
    \label{app:tab:llm_judge_prompts}
\end{table*}

\section{Vector Space Analysis Results}
\label{app:sec:vector_space_graphs}
\begin{figure*}
    \centering
    \includegraphics[width=\linewidth,scale=0.9,trim={0 2cm 0cm 2cm},clip]{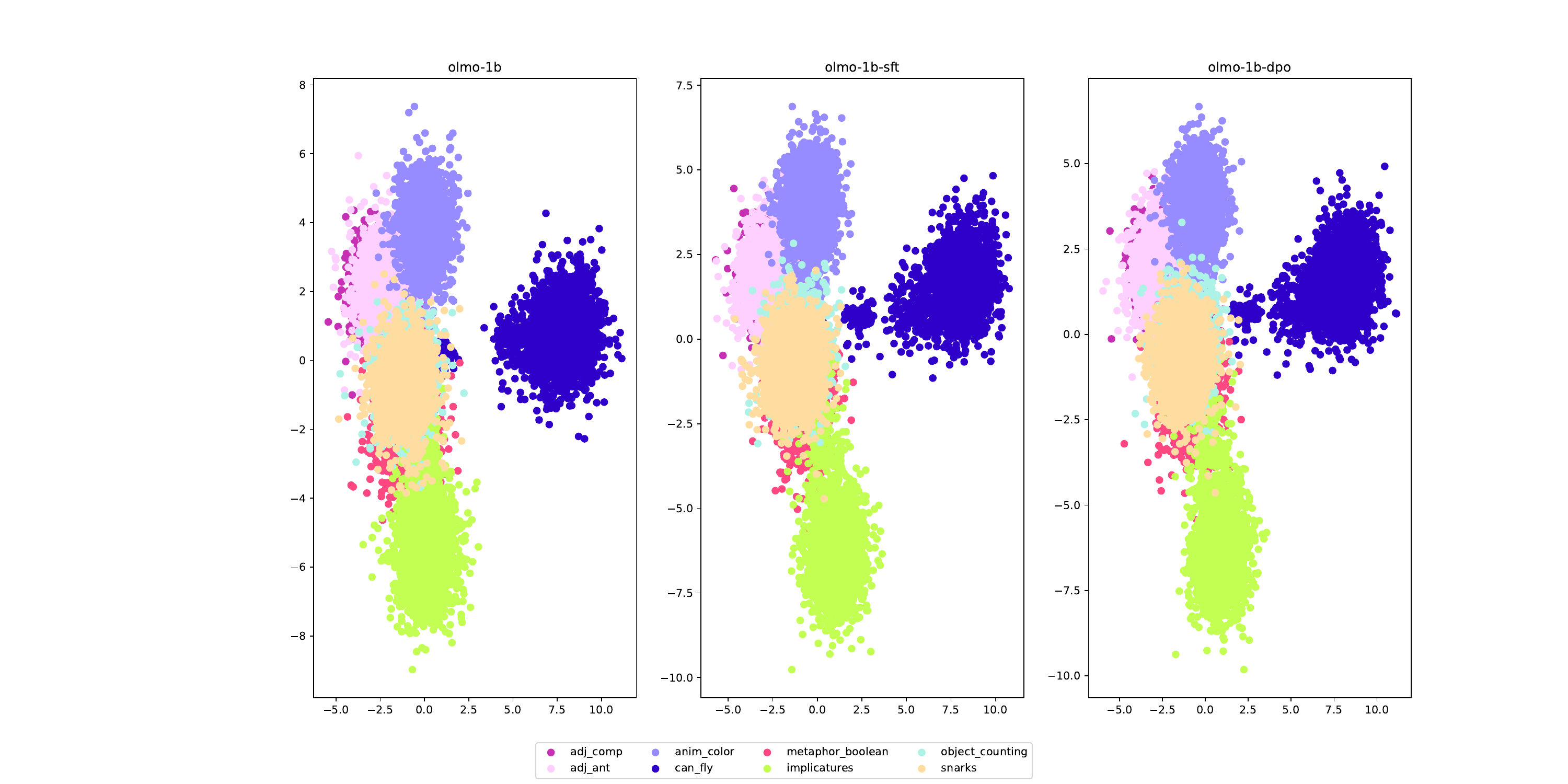}
    
    \includegraphics[width=\linewidth,scale=0.9,trim={0 0cm 0cm 2cm},clip]{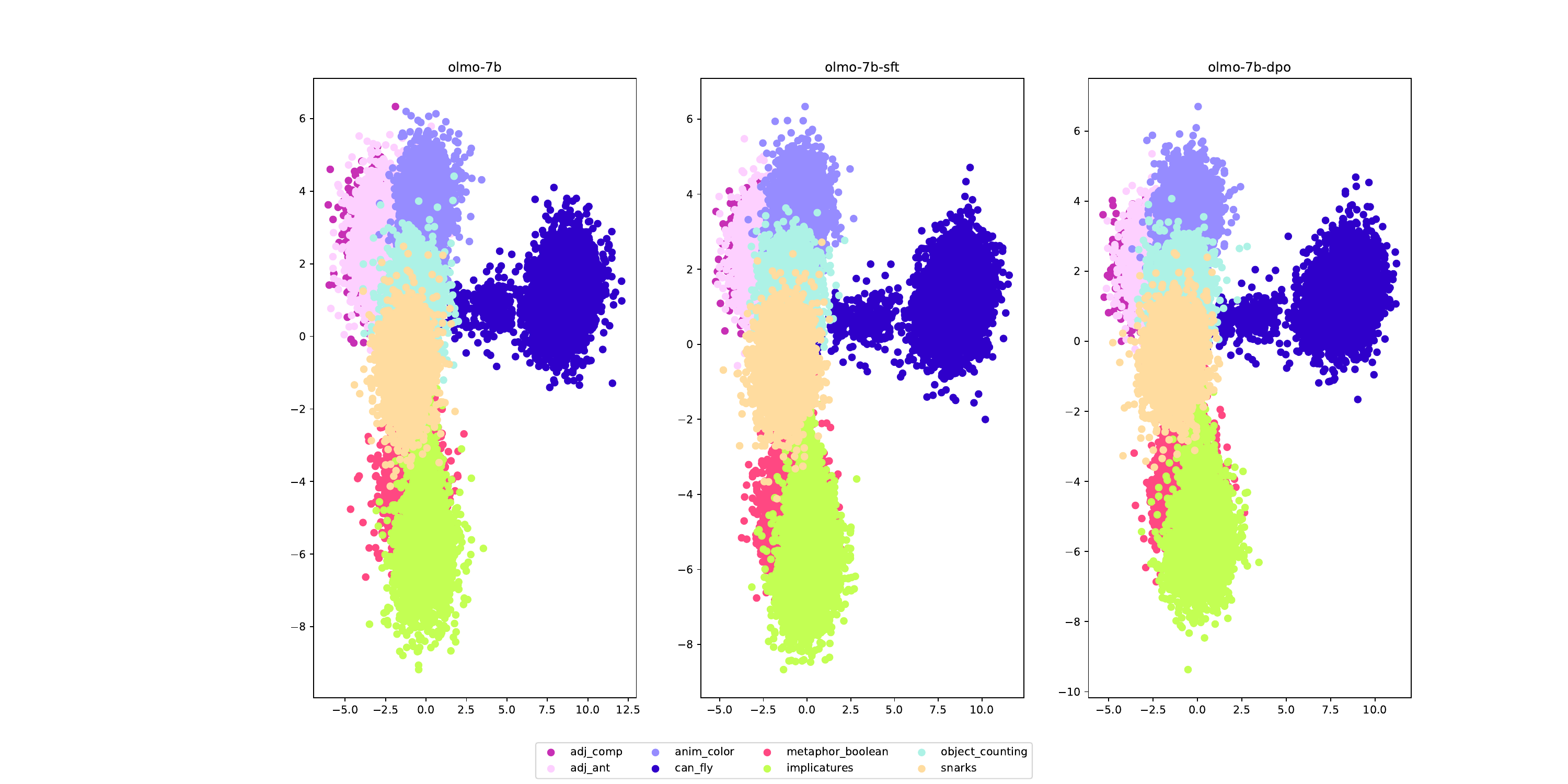}
    
    \caption{Our LDA analysis of 200 instruction rephrasals on OLMo-2 models shows that tasks form organized clusters, in line with previous results. However, they are not always linearly separable (e.g. "adj\_comp" and "adj\_ant").}
    \label{fig:lda-plots}
\end{figure*}

In Figure \ref{fig:lda-plots}, we present the LDA task representations for each of our 8 tasks. For each task, we produce 200 rephrasings of the instruction as described in Section \ref{subsec:spatial_analysis}.

We prompt ChatGPT-5-nano \cite{chatgpt5-nano} using the following prompt:

\begin{quote}
This is a task instruction: `\{\}' Please rephrase this instruction in 200 different ways without changing the essential meaning and without specifying any particular statements the speakers made. Please vary the rephrasals in terms of syntax, linguistic style, and complexity.
\end{quote}

where we fill in the blank using the standard instruction in Table \ref{tab:contrastive_tasks}.

\section{Path Analysis Results with top-$k$ Attention}
\label{app:sec:path_analysis_topk}
In this section, we present supplementary results for our path analysis experiments in Section \ref{sec:causal_experiments}. These results reinforce the role of the $\emT_\text{inst}$ as a primary source of high-ranking paths, particularly in comparison to later, non-instructional tokens.

\begin{figure}
    \centering
    \includegraphics[width=\linewidth,scale=0.7,trim={0 0cm 0cm 0cm},clip]{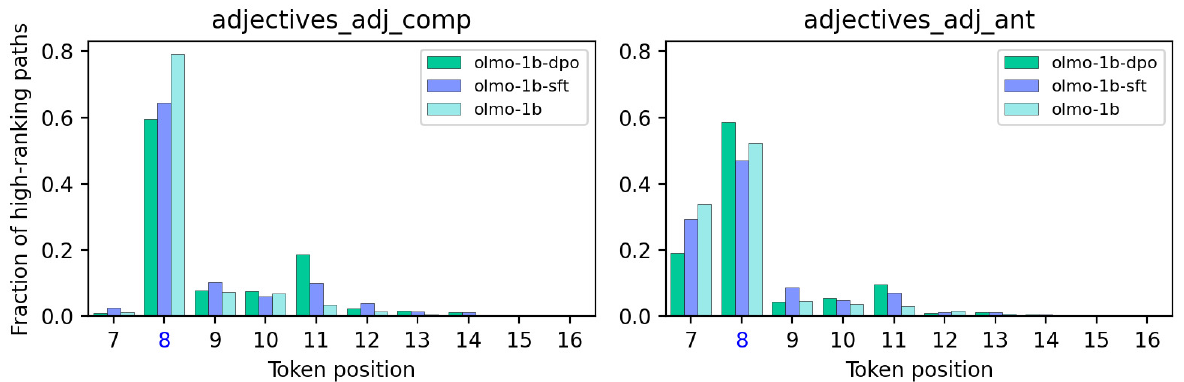}

    \caption{Fraction of high-ranking paths contributed by each token position, across all observed high-ranking paths. Top-$k$ attention was used, with $k{=}1$.}
    \label{app:fig:path_k1}
\end{figure}

\begin{figure}
    \centering
    \includegraphics[width=\linewidth,scale=0.7,trim={0 0cm 0cm 0cm},clip]{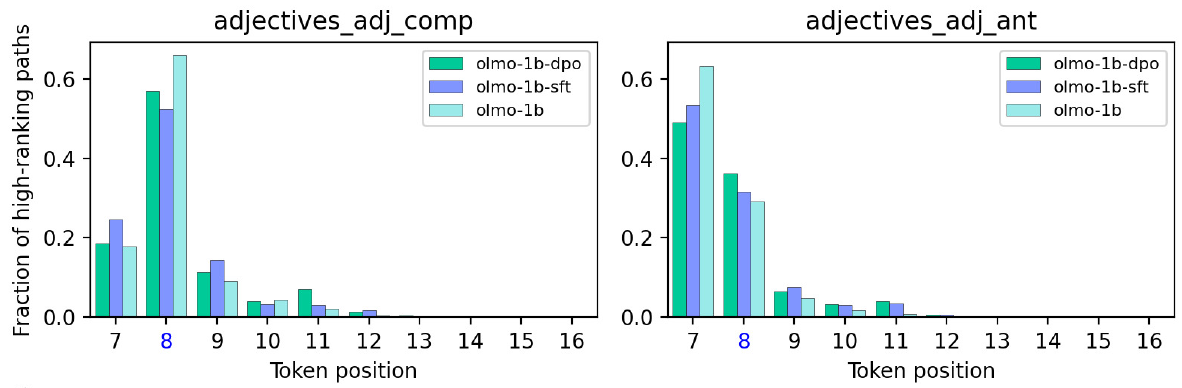}

    \caption{Fraction of high-ranking paths contributed by each token position, across all observed high-ranking paths. Top-$k$ attention was used, with $k{=}2$.}
    \label{app:fig:path_k2}
\end{figure}

\begin{figure}
    \centering
    \includegraphics[width=\linewidth,scale=0.7,trim={0 0cm 0cm 0cm},clip]{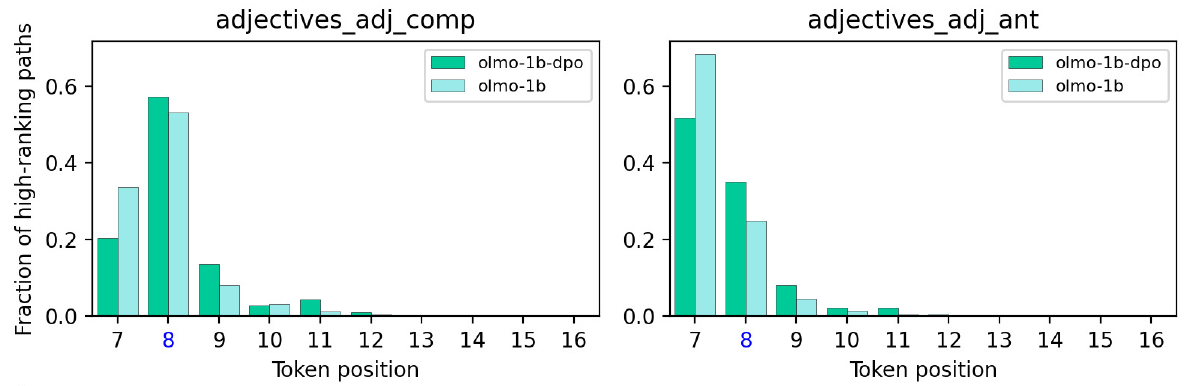}

    \caption{Fraction of high-ranking paths contributed by each token position, across all observed high-ranking paths. Top-$k$ attention was used, with $k{=}3$. Due to constraints, the SFT model was not analyzed.}
    \label{app:fig:path_k3}
\end{figure}

Figures \ref{app:fig:path_k1} - \ref{app:fig:path_k3} show the fraction of high-ranking paths contributed by each token position, across all examined tokens (positions 7 - 16). Due to long computation times and memory requirements, we do not analyze the full prompt. We notice that the $\emT_\text{inst}$ token tends to yield the greatest portion of high-ranking paths across model checkpoints for \textsc{Adjectives: Comparative}, and consistently more high-ranking paths than later tokens. For \textsc{Adjectives: Antonym}, $\emT_\text{inst}-1$ (Token 7) yields more high-ranking paths for all models in $k{>}1$\footnote{We note that, due to experiment constraints, we were unable to run the SFT model in the $k{=}3$ setting.}. This could be due to the greater open-endedness of the antonyms task, resulting in a wider pool of possible answers that must be sorted through by the model.

\section{Attention Head Activity Experiments}
\label{app:mainsec:attention_head_activity}
In this section, we present the full details of our experiments regarding attention head activity.

\subsection{Confidence and Significance Tests}
\label{app:sec:attention_head_activity_stat_tests}
\begin{figure}
    \centering
    \includegraphics[width=\linewidth,scale=0.7,trim={0 0cm 0cm 0cm},clip]{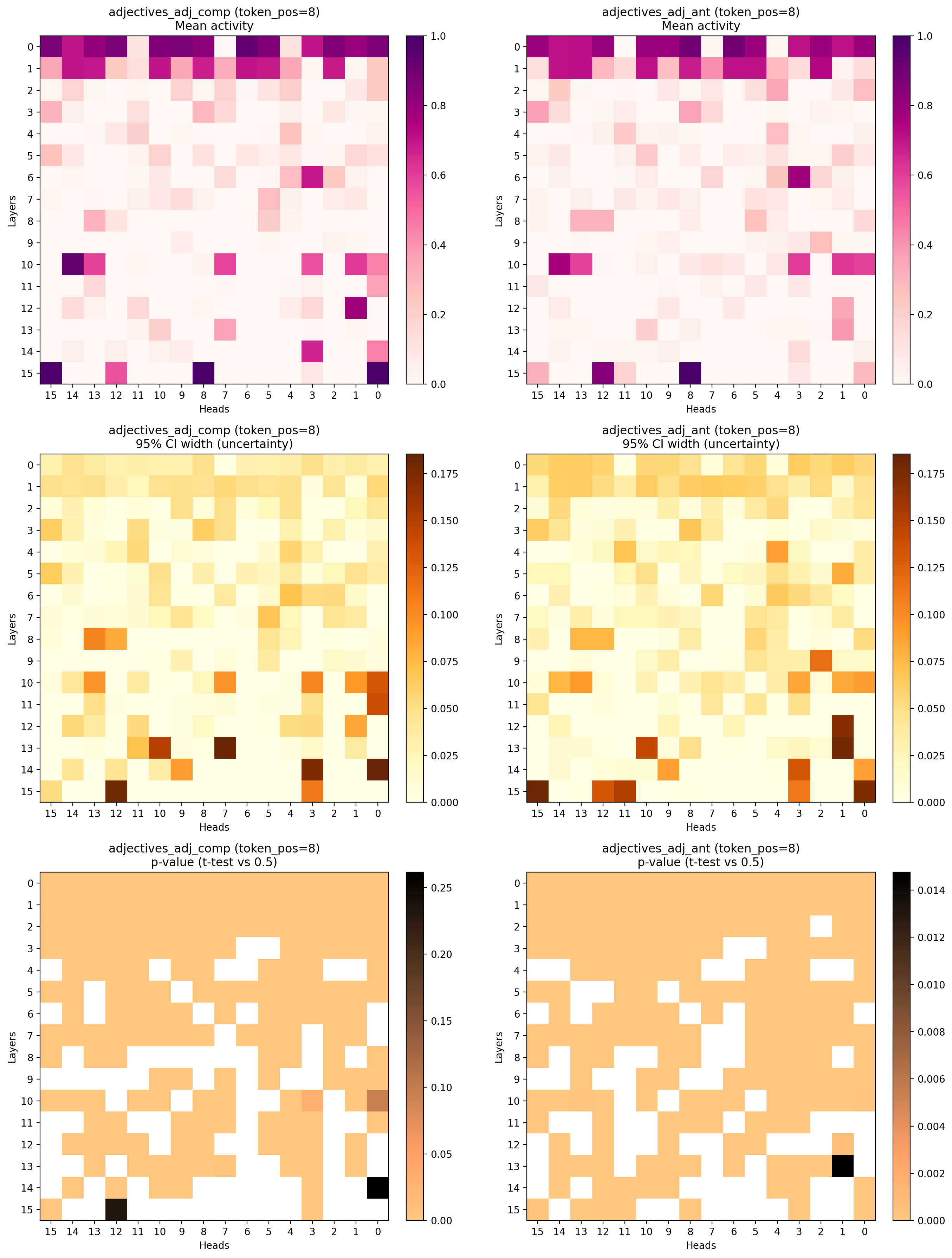}

    \caption{Results of our statistical tests of attention head activity, for OLMo-1B, $k{=}1$. Top: average activity across 100 samples. Middle: boostrapped confidence intervals, with lighter=narrower. Bottom: p-values showing significance of head variance against Gaussian noise, with lighter=lower p-value.}
    \label{fig:bootstrapped-app}
\end{figure}

\begin{figure}
    \centering
    \includegraphics[width=\linewidth,scale=0.7,trim={0 0cm 0cm 0cm},clip]{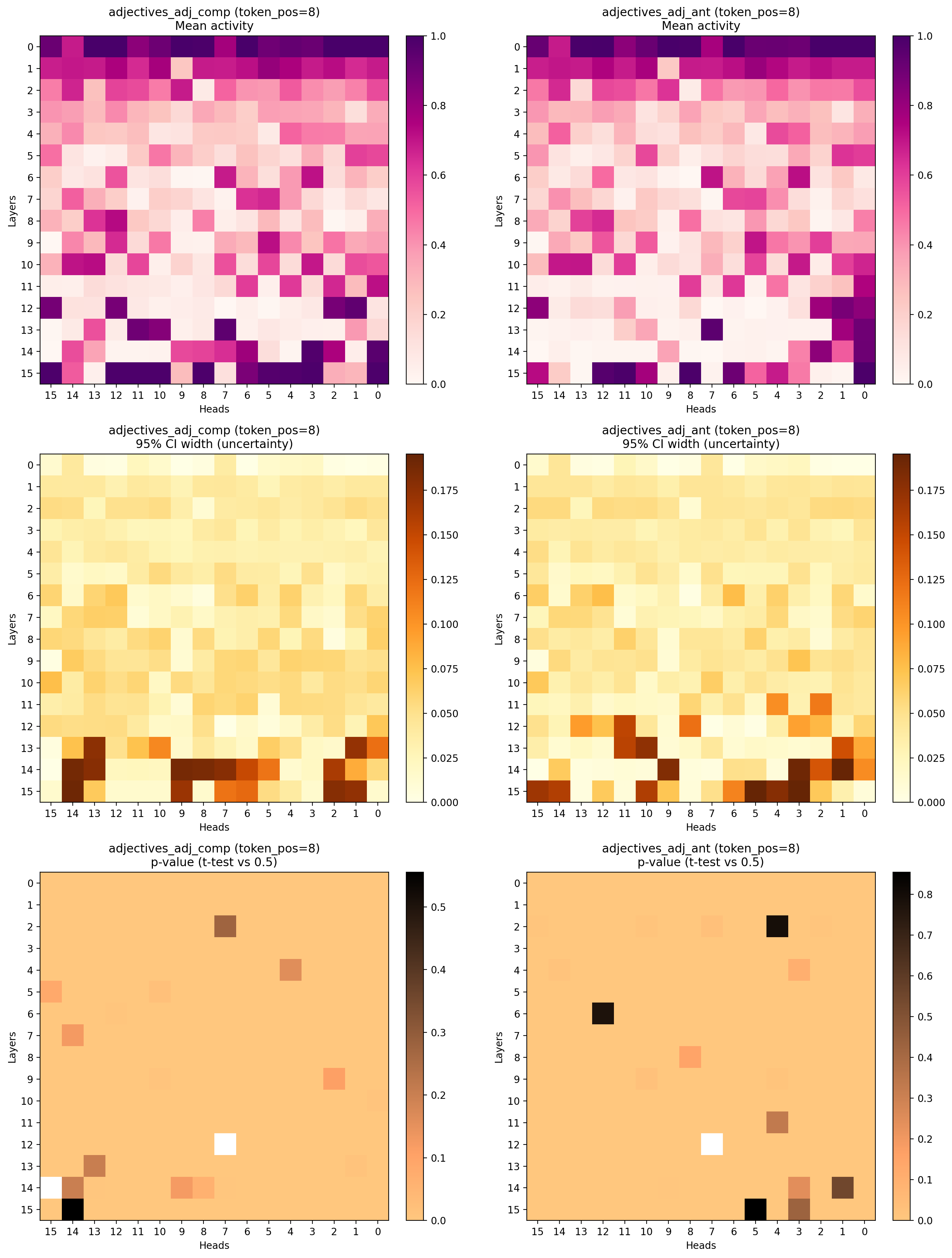}

    \caption{Results of our statistical tests of attention head activity, for OLMo-1B, $k{=}1$. Top: average activity across 100 samples. Middle: boostrapped confidence intervals, with lighter=narrower. Bottom: p-values showing significance of head variance against Gaussian noise, with lighter=lower p-value.}
    \label{fig:bootstrapped-2}
\end{figure}

\begin{figure}
    \centering
    \includegraphics[width=\linewidth,scale=0.7,trim={0 0cm 0cm 0cm},clip]{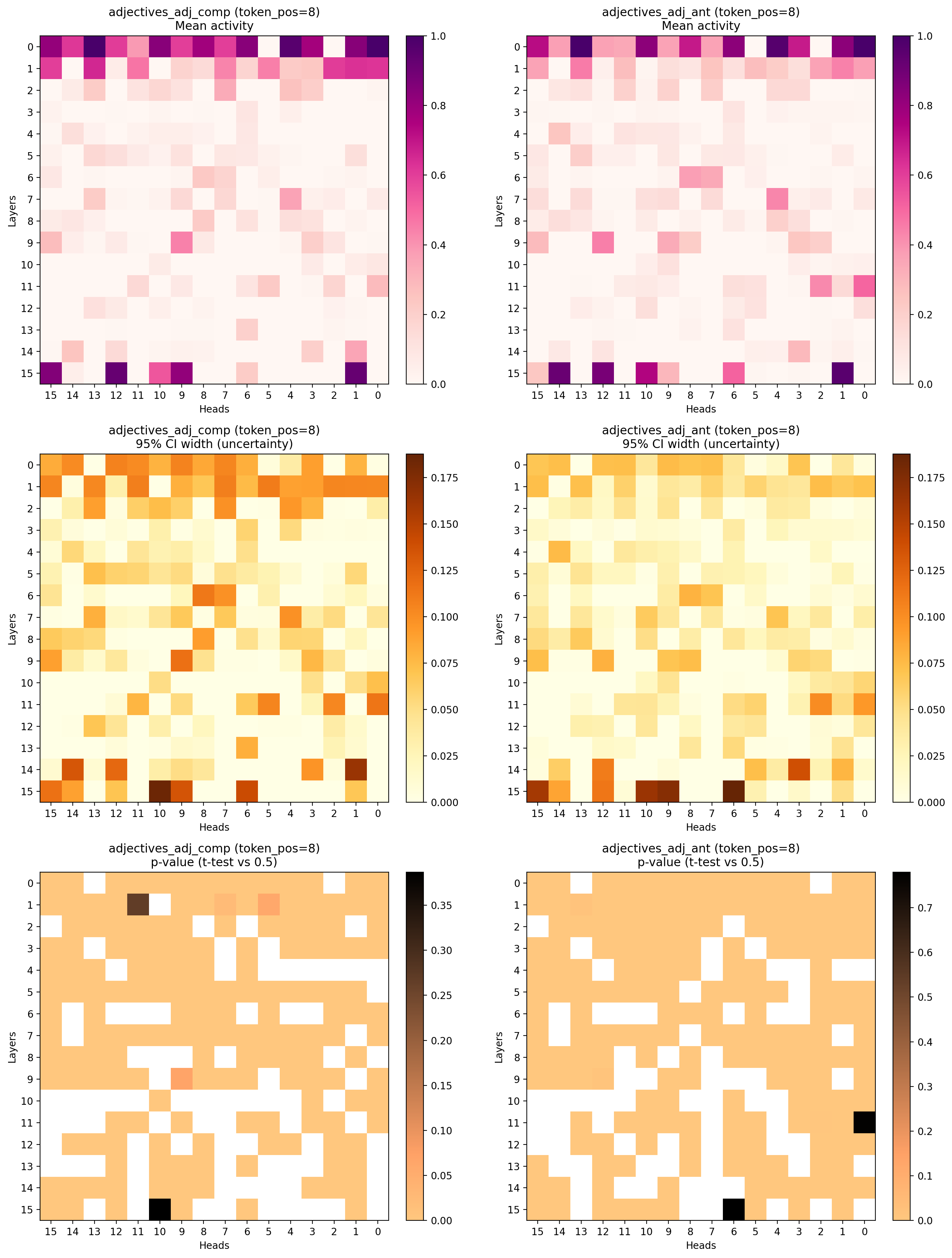}

    \caption{Results of our statistical tests of attention head activity, for OLMo-1B-SFT, $k{=}1$. Top: average activity across 100 samples. Middle: boostrapped confidence intervals, with lighter=narrower. Bottom: p-values showing significance of head variance against Gaussian noise, with lighter=lower p-value.}
    \label{fig:bootstrapped-3}
\end{figure}

\begin{figure}
    \centering
    \includegraphics[width=\linewidth,scale=0.7,trim={0 0cm 0cm 0cm},clip]{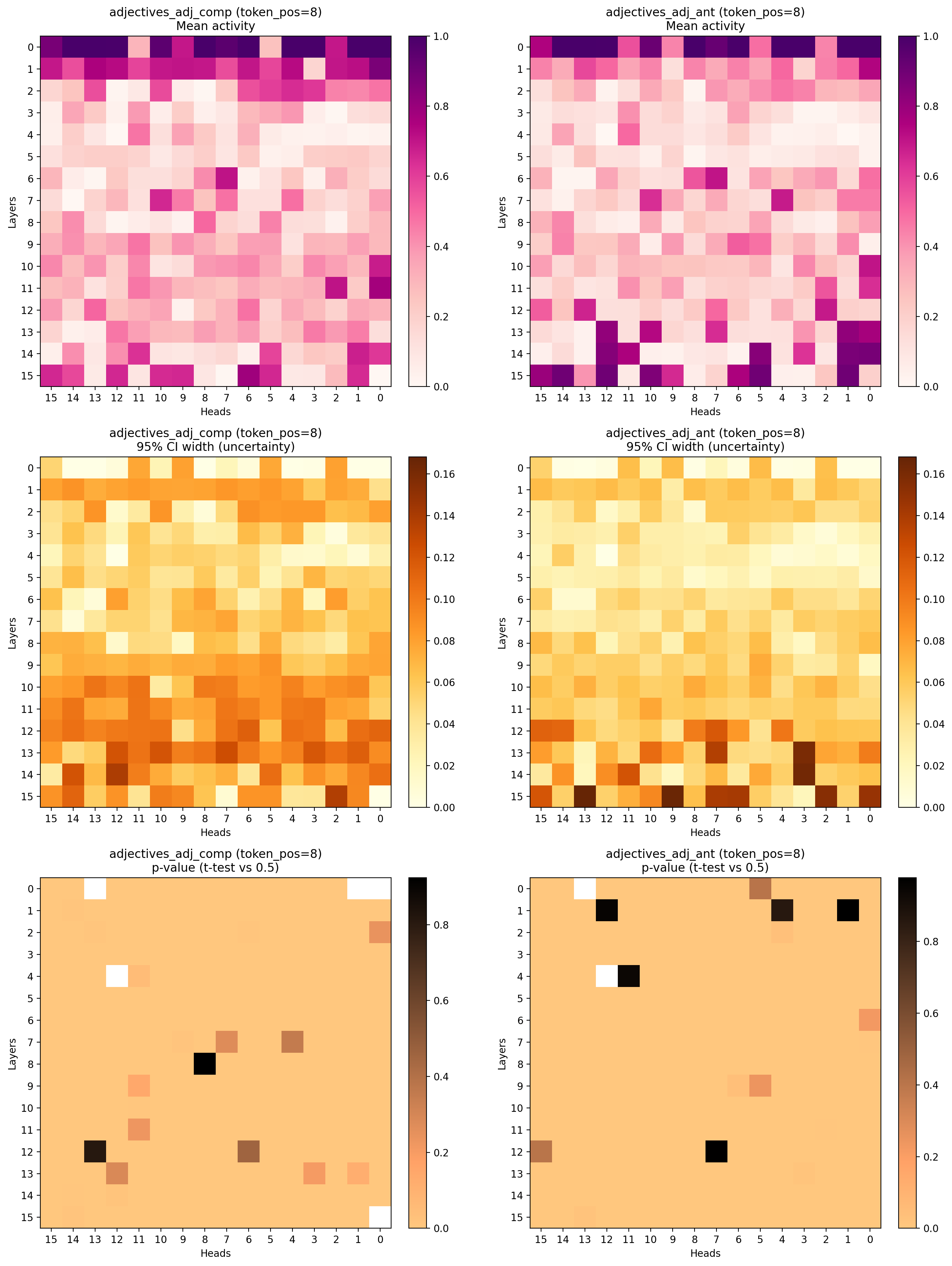}

    \caption{Results of our statistical tests of attention head activity, for OLMo-1B-SFT, $k{=}1$. Top: average activity across 100 samples. Middle: boostrapped confidence intervals, with lighter=narrower. Bottom: p-values showing significance of head variance against Gaussian noise, with lighter=lower p-value.}
    \label{fig:bootstrapped-4}
\end{figure}

\begin{figure}
    \centering
    \includegraphics[width=\linewidth,scale=0.7,trim={0 0cm 0cm 0cm},clip]{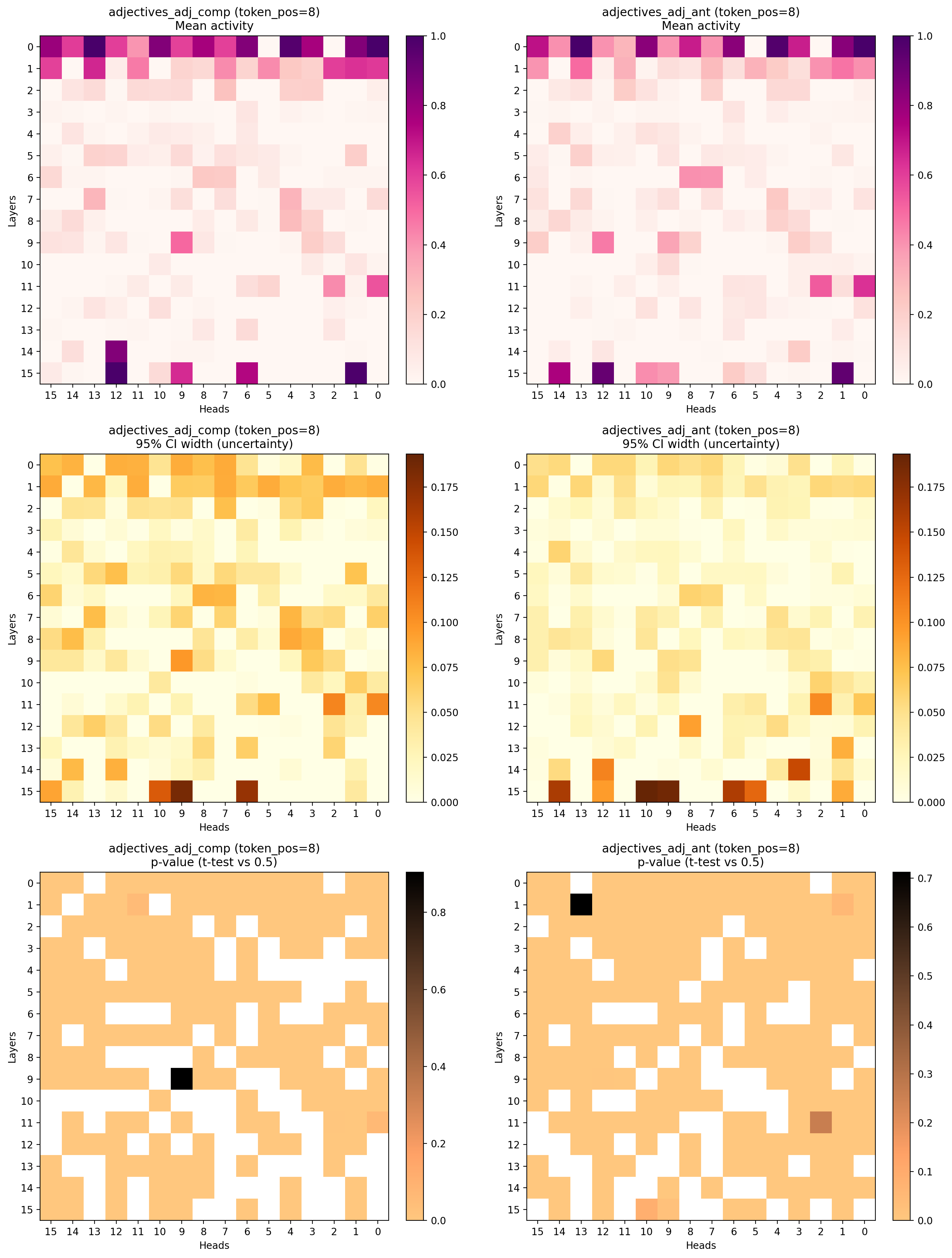}

    \caption{Results of our statistical tests of attention head activity, for OLMo-1B-DPO, $k{=}1$. Top: average activity across 100 samples. Middle: boostrapped confidence intervals, with lighter=narrower. Bottom: p-values showing significance of head variance against Gaussian noise, with lighter=lower p-value.}
    \label{fig:bootstrapped-5}
\end{figure}

\begin{figure}
    \centering
    \includegraphics[width=\linewidth,scale=0.7,trim={0 0cm 0cm 0cm},clip]{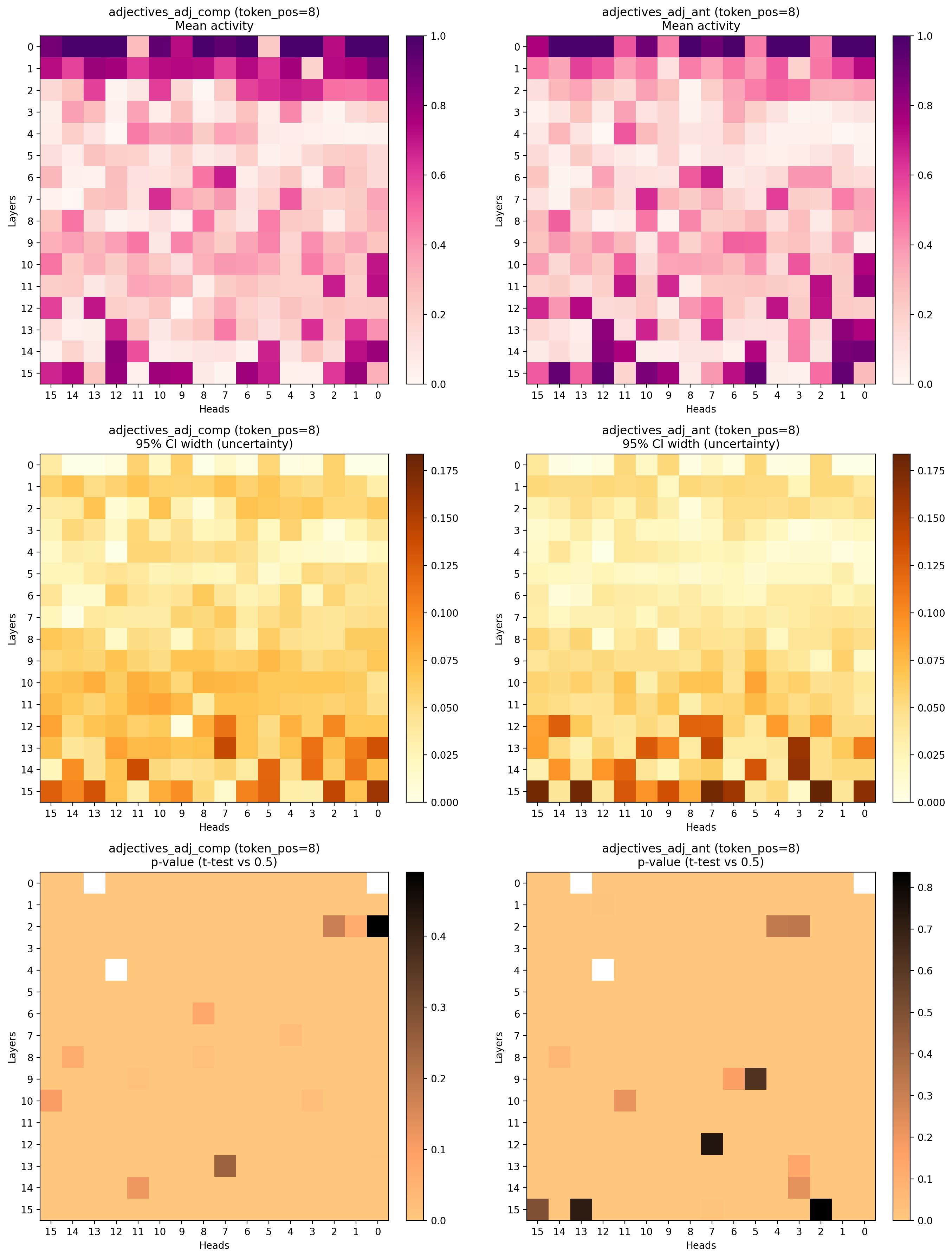}

    \caption{Results of our statistical tests of attention head activity, for OLMo-1B-DPO, $k{=}1$. Top: average activity across 100 samples. Middle: boostrapped confidence intervals, with lighter=narrower. Bottom: p-values showing significance of head variance against Gaussian noise, with lighter=lower p-value.}
    \label{fig:bootstrapped-6}
\end{figure}

Figures \ref{fig:bootstrapped-app} - \ref{fig:bootstrapped-6} show the mean activity rates for attention heads across the adjectives tasks, along with the bootstrapped confidence intervals and p-values.

\begin{figure}
    \centering
    \includegraphics[width=\linewidth,scale=0.7,trim={0 0cm 0cm 0cm},clip]{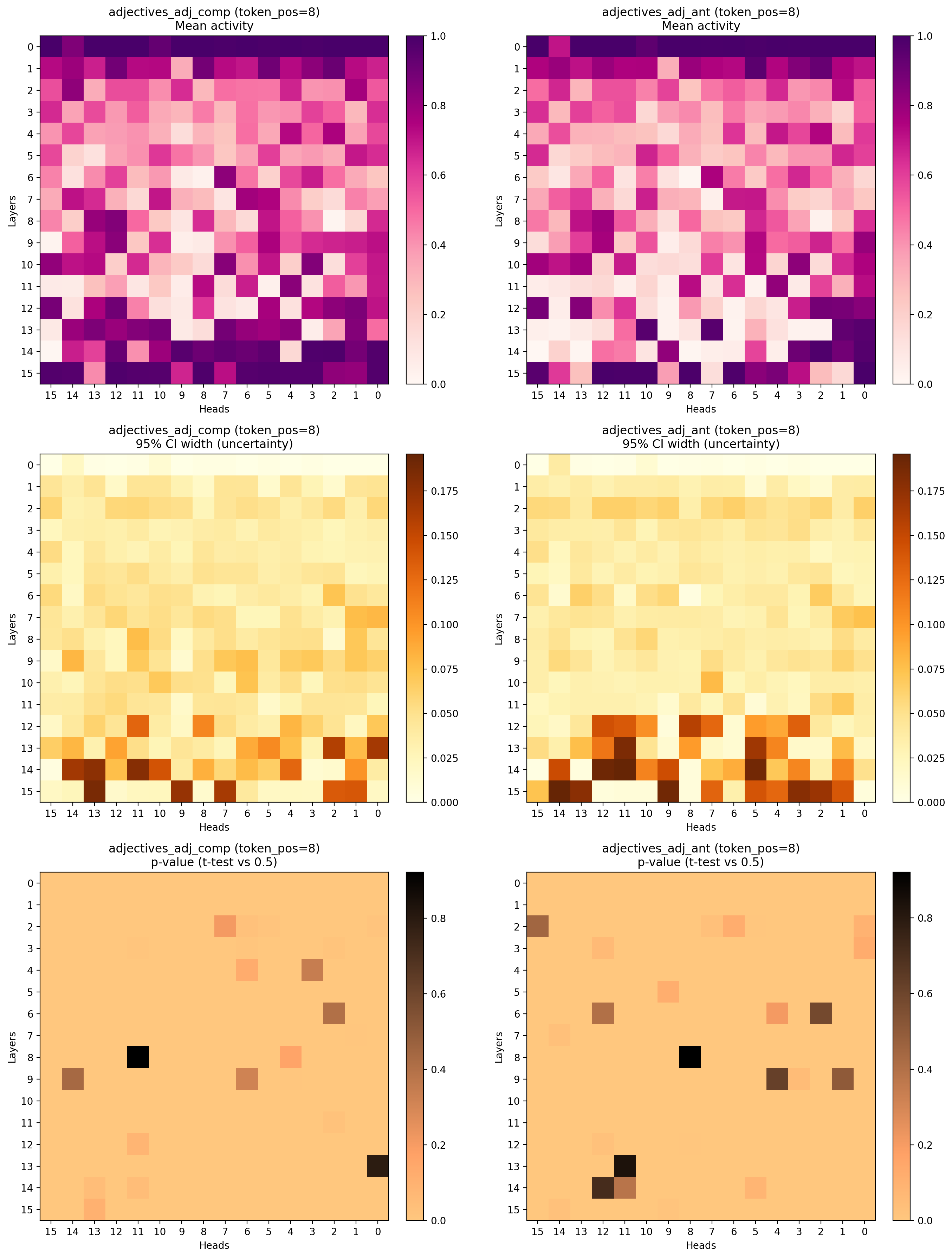}
    \caption{Results of our statistical tests of attention head activity, for OLMo-1B, k=3. Top: average activity across 100 samples. Middle: boostrapped confidence intervals, with lighter=narrower. Bottom: p-values showing significance of head variance against Gaussian noise, with lighter=lower p-value.}
    \label{fig:bootstrapped-7}
\end{figure}

\begin{figure}
    \centering
    \includegraphics[width=\linewidth,scale=0.7,trim={0 0cm 0cm 0cm},clip]{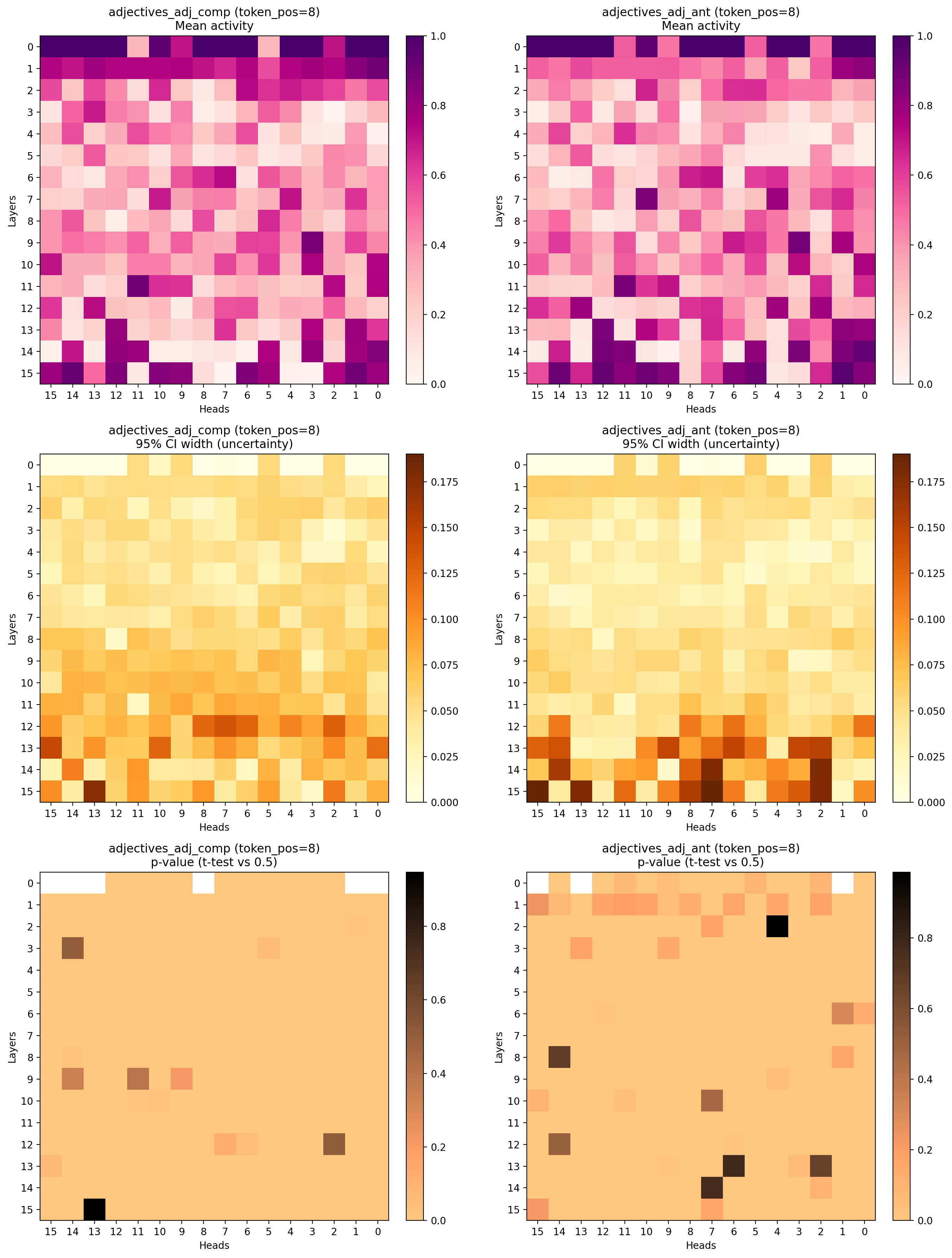}
    \caption{Results of our statistical tests of attention head activity, for OLMo-1B-DPO, k=3. Top: average activity across 100 samples. Middle: boostrapped confidence intervals, with lighter=narrower. Bottom: p-values showing significance of head variance against Gaussian noise, with lighter=lower p-value.}
    \label{fig:bootstrapped-8}
\end{figure}

\begin{figure*}
    \centering
    \includegraphics[width=\linewidth,scale=0.7,trim={0 0cm 0cm 0cm},clip]{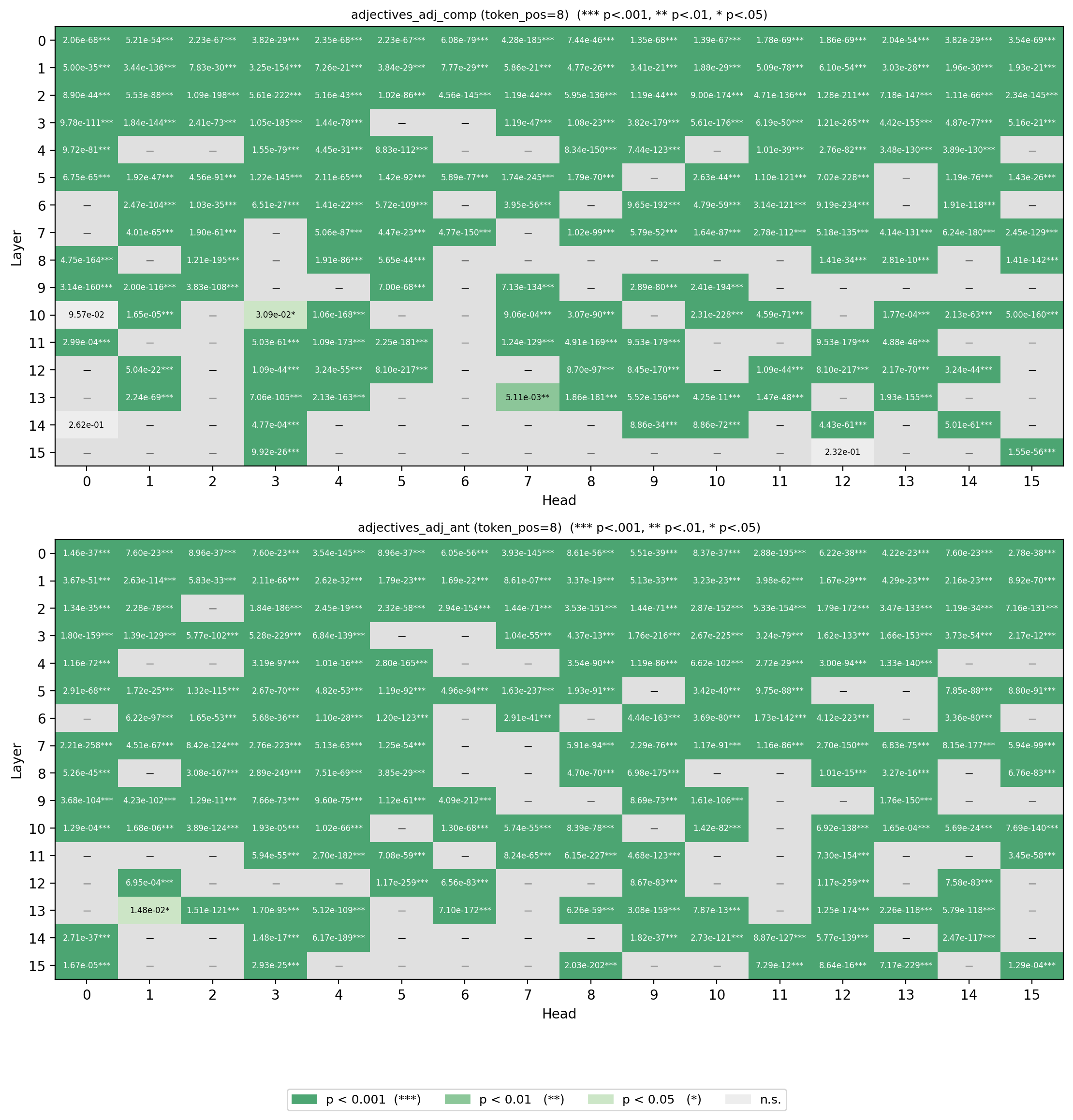}
    \caption{Raw p-values of attention head activity for OLMo-1B, $k{=}1$.}
    \label{fig:p-value-olmo1b-k1}
\end{figure*}

\begin{figure*}
    \centering
    \includegraphics[width=\linewidth,scale=0.7,trim={0 0cm 0cm 0cm},clip]{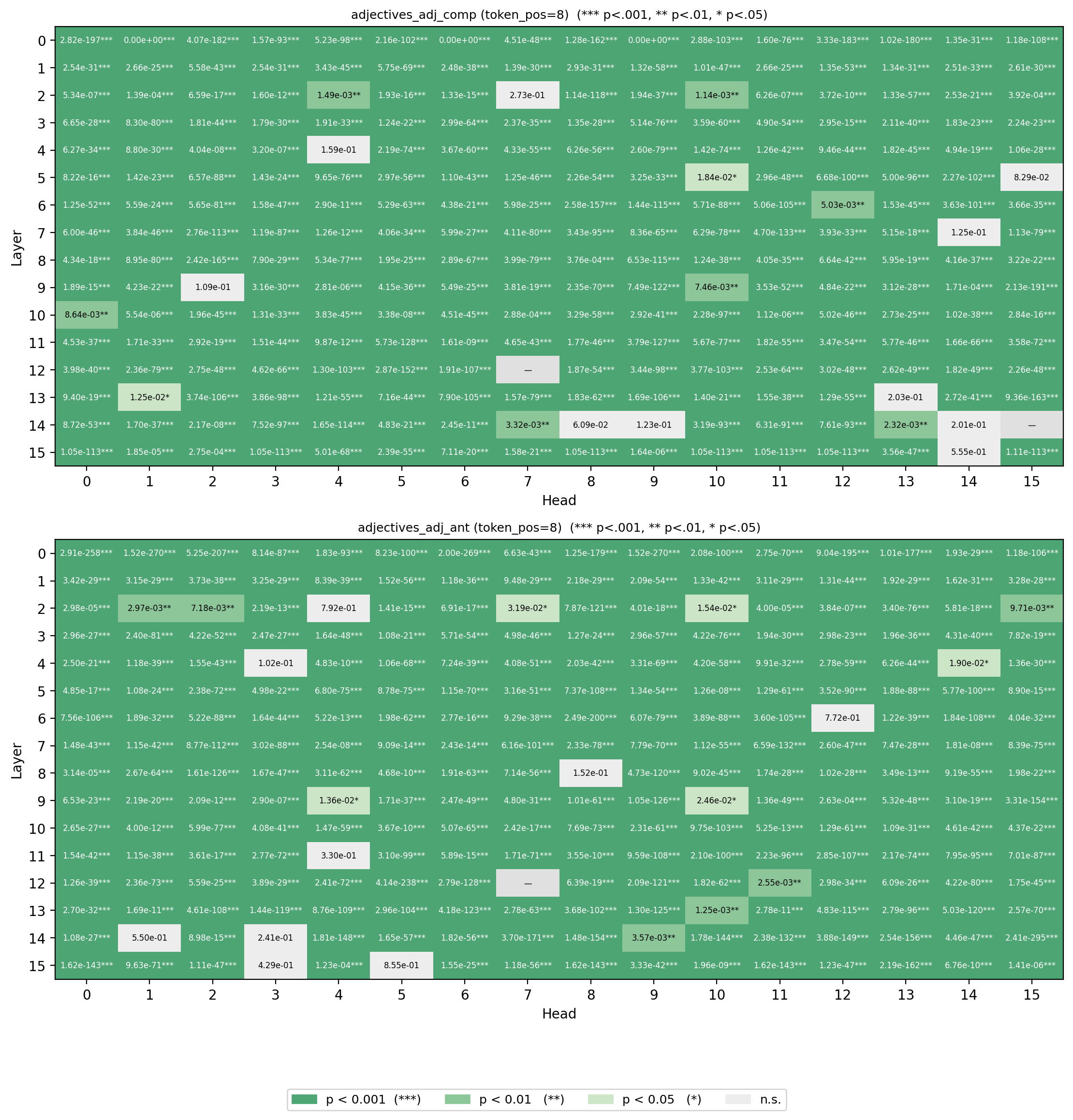}
    \caption{Raw p-values of attention head activity for OLMo-1B, $k{=}2$.}
    \label{fig:p-value-olmo1b-k2}
\end{figure*}

\begin{figure*}
    \centering
    \includegraphics[width=\linewidth,scale=0.7,trim={0 0cm 0cm 0cm},clip]{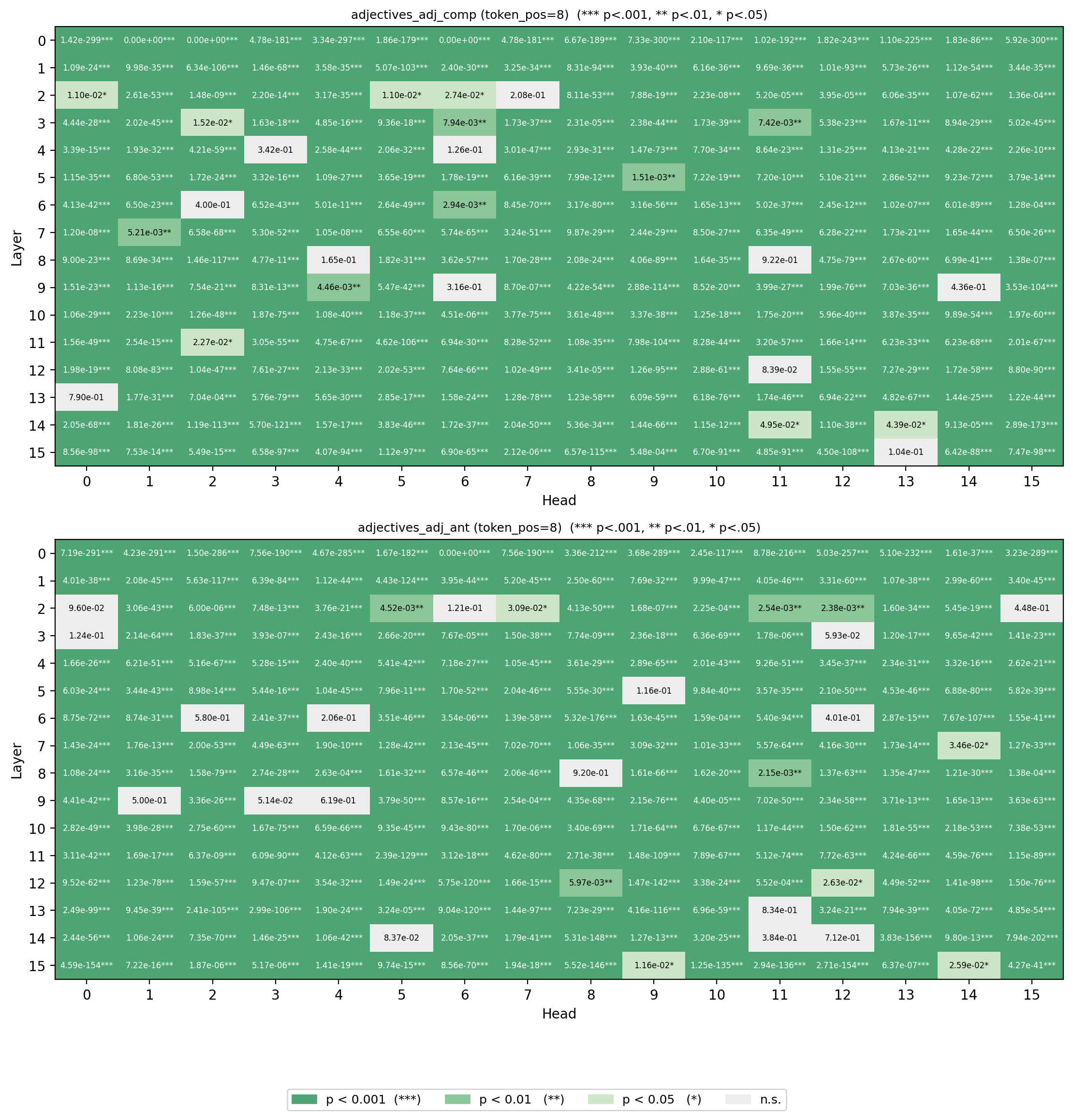}
    \caption{Raw p-values of attention head activity for OLMo-1B, $k{=}3$.}
    \label{fig:p-value-olmo1b-k3}
\end{figure*}

\begin{figure*}
    \centering
    \includegraphics[width=\linewidth,scale=0.7,trim={0 0cm 0cm 0cm},clip]{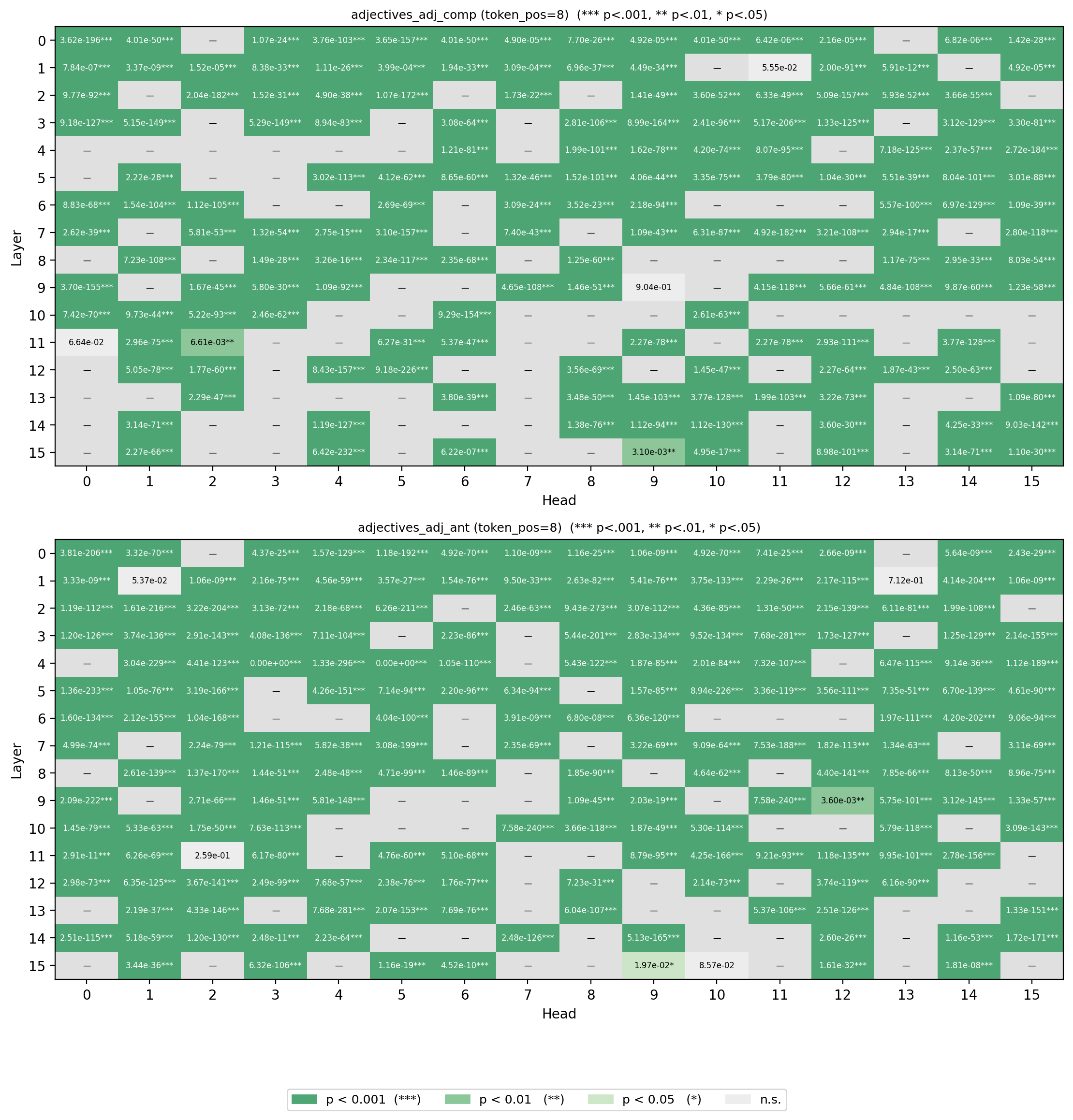}
    \caption{Raw p-values of attention head activity for OLMo-1B-DPO, $k{=}1$.}
    \label{fig:p-value-olmo1b-dpo-k1}
\end{figure*}

\begin{figure*}
    \centering
    \includegraphics[width=\linewidth,scale=0.7,trim={0 0cm 0cm 0cm},clip]{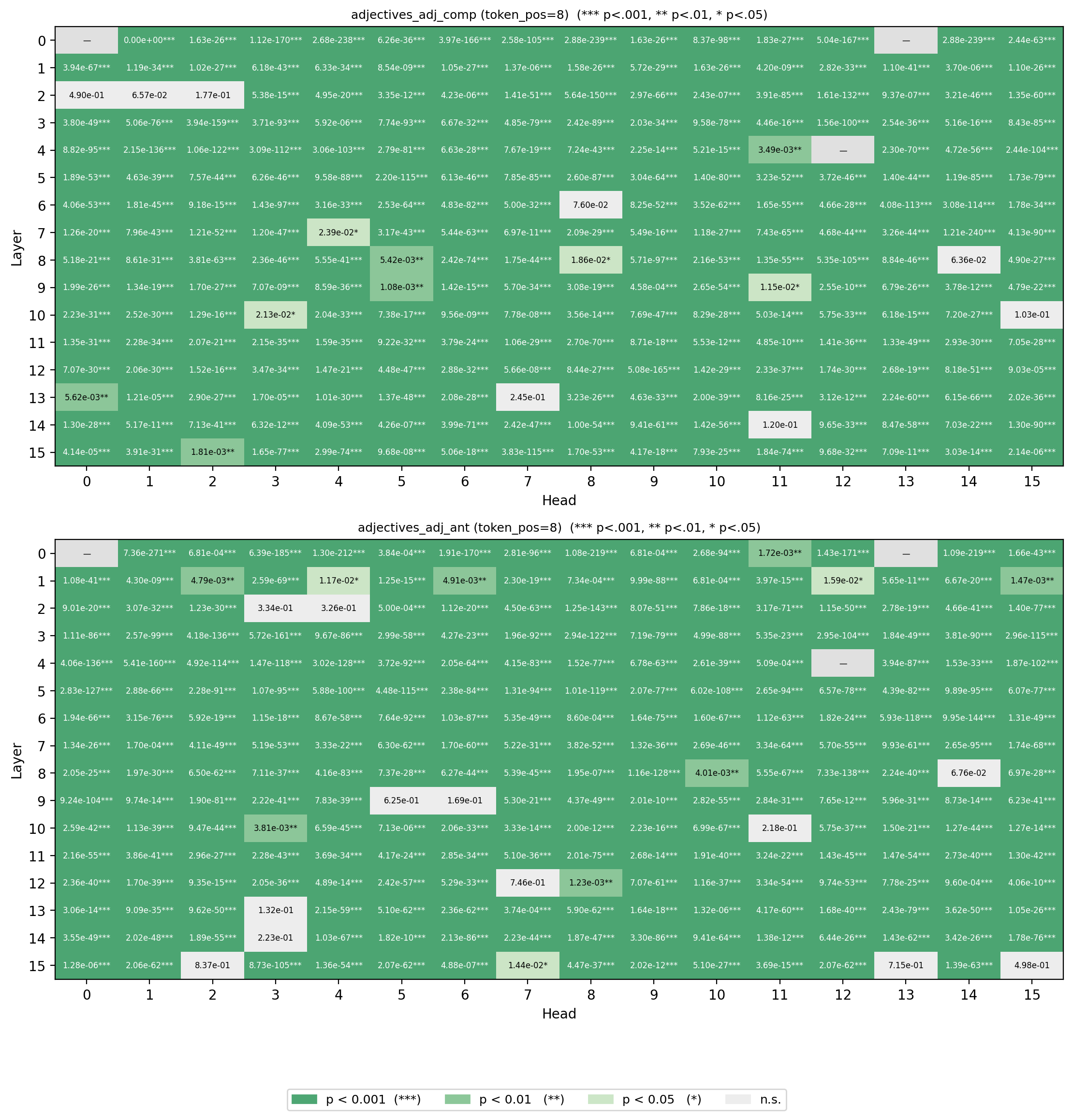}
    \caption{Raw p-values of attention head activity for OLMo-1B-DPO, $k{=}2$.}
    \label{fig:p-value-olmo1b-dpo-k2}
\end{figure*}

\begin{figure*}
    \centering
    \includegraphics[width=\linewidth,scale=0.7,trim={0 0cm 0cm 0cm},clip]{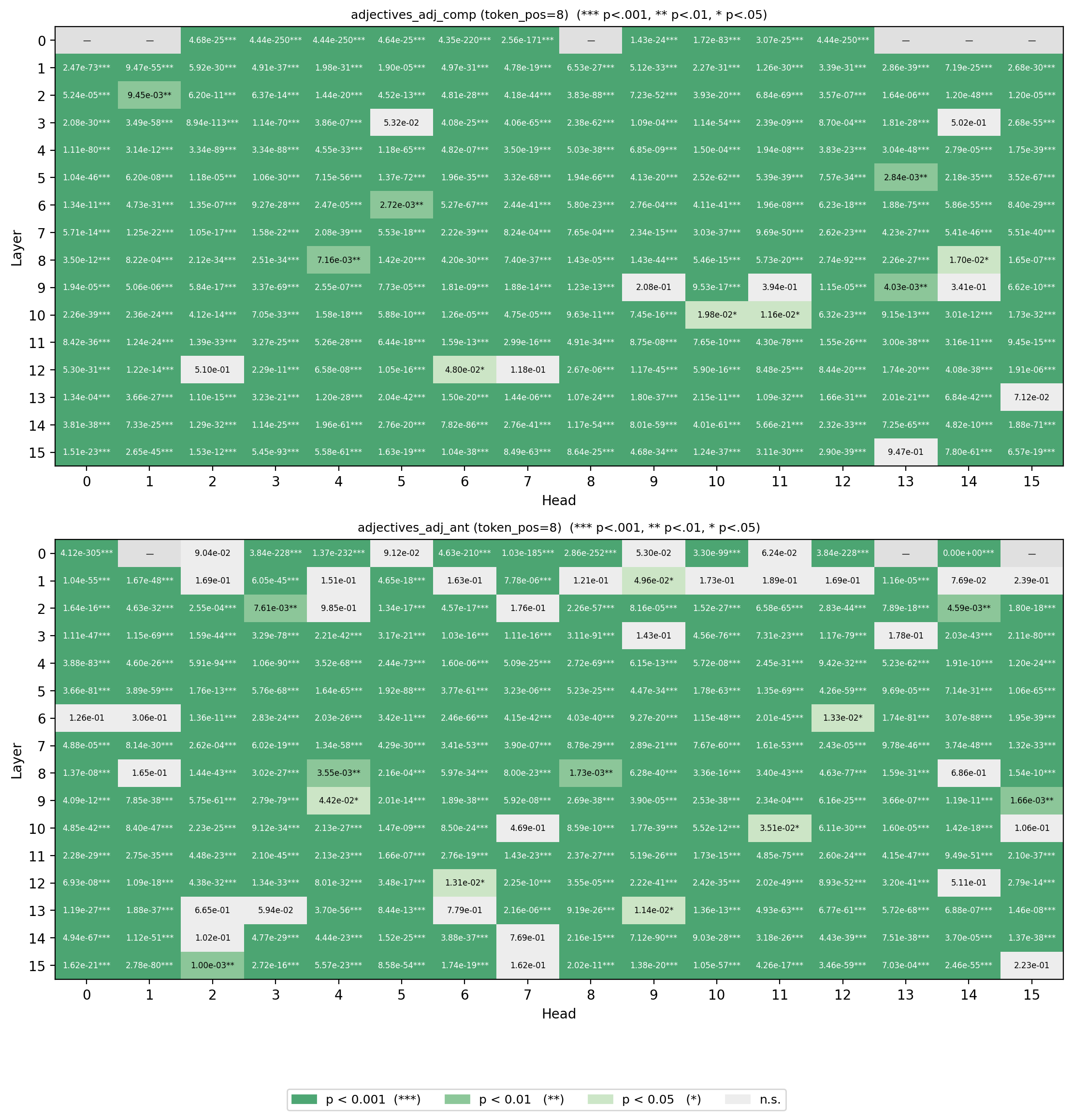}
    \caption{Raw p-values of attention head activity for OLMo-1B-DPO, $k{=}3$.}
    \label{fig:p-value-olmo1b-dpo-k3}
\end{figure*}

\begin{figure*}
    \centering
    \includegraphics[width=\linewidth,scale=0.7,trim={0 0cm 0cm 0cm},clip]{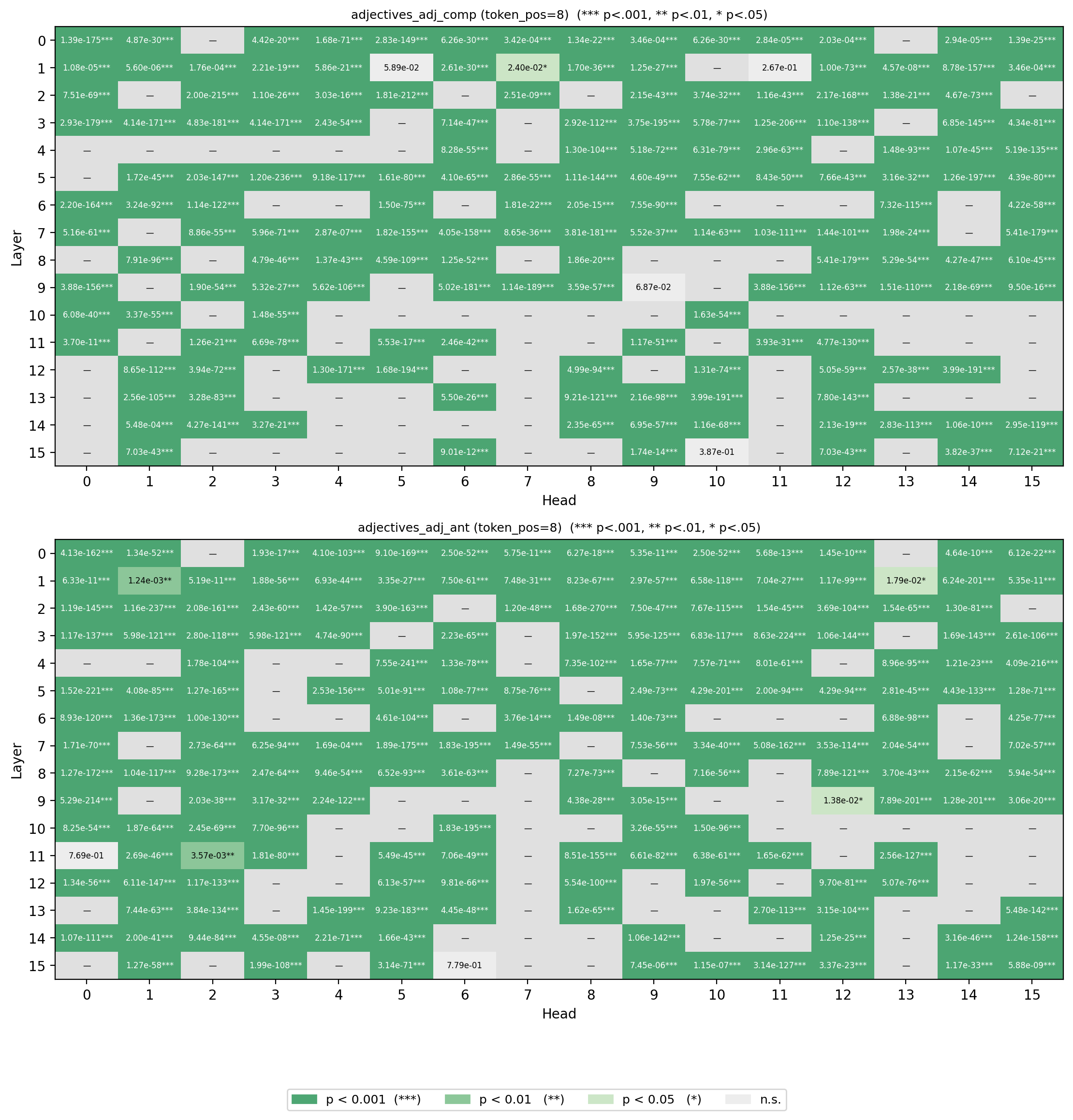}
    \caption{Raw p-values of attention head activity for OLMo-1B-SFT, $k{=}1$.}
    \label{fig:p-value-olmo1b-sft-k1}
\end{figure*}

\begin{figure*}
    \centering
    \includegraphics[width=\linewidth,scale=0.7,trim={0 0cm 0cm 0cm},clip]{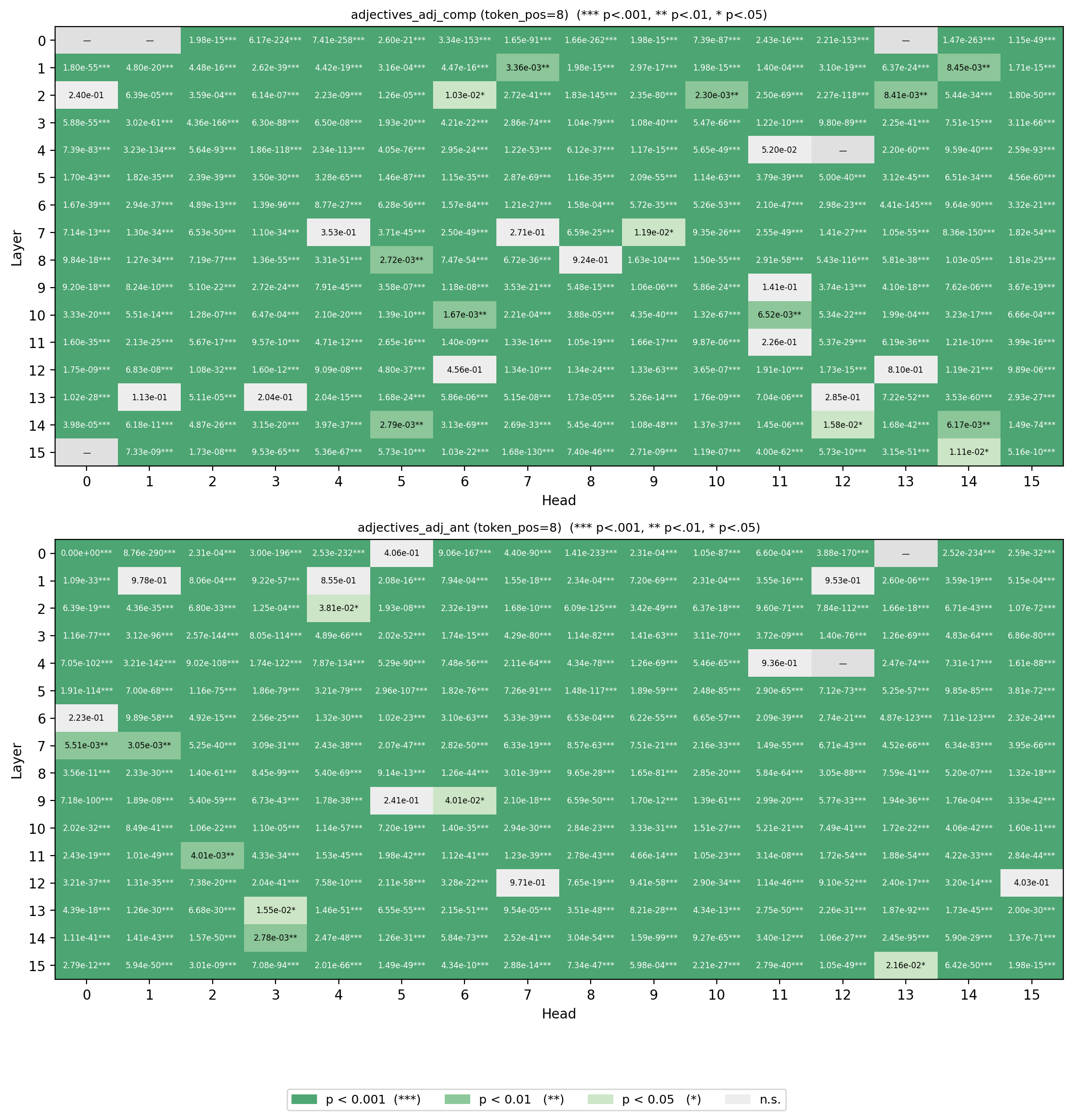}
    \caption{Raw p-values of attention head activity for OLMo-1B-SFT, $k{=}2$.}
    \label{fig:p-value-olmo1b-sft-k2}
\end{figure*}

\subsection{Similarity of Head Activity Across Contrastive Instructions and $k$ Values}
\label{app:subsec:jaccard-tables}

\paragraph{$k$ Values.}To determine the impact of top-$k$ attention with various values of $k$ on the head activity patterns, we calculate the Jaccard similarity of the average head activity between $k{=}1$ and $k{=}3$ (Table \ref{tab:app:jaccard-similarity2}), as well as between $k{=}2$ and $k{=}3$ (Table \ref{tab:app:jaccard-similarity3}). The comparison between $k{=}1$ and $k{=}2$ is given in Table \ref{tab:jaccard-similarity}.

We find that from $k{=}1$ to $k{=}2$, there is a moderate similarity in head activity, mostly in the early layers (Table \ref{tab:app:jaccard-similarity2}). The lower similarity in the later layers indicates that there is some change in head activity here. However, when going from $k{=}2$ to $k{=}3$, the similarity scores are higher overall -- this indicates that the head activity pattern has mostly stabilized and that few heads change significantly at this point (Table \ref{tab:app:jaccard-similarity3}).

\paragraph{Tasks.} The similarity scores for head activity between the two contrastive \textsc{Adjectives} is shown in Tables \ref{tab:app:jac-sim-across-tasks-k1} - \ref{tab:app:jac-sim-across-tasks-k3}. In accordance with our result that increasing $k$ also increases general head activity, we find that the similarity increases with $k$ for most of our selected layer groups. However, across $k$, we notice that Layers 0-2 have the highest similarity in activity at the $\emT_\text{inst}$ token position, and that the lowest similarity scores are observed somewhere in the later layer groups (6-9 or 10-15). Moreover, we observe that this high similarity does not occur before $\emT_\text{inst}$, and seldom occurs after. This indicates that the $\emT_\text{inst}$ position is the location of a particular kind of computation that does not occur at the other token positions.

\begin{table}[h!]
\centering
\resizebox{\linewidth}{!}{
\begin{tabular}{l ccc ccc}
\toprule
& \multicolumn{3}{c}{\texttt{adj\_comp}} & \multicolumn{3}{c}{\texttt{adj\_ant}} \\
\cmidrule(lr){2-4} \cmidrule(lr){5-7}
Model & L0--1 & L4--5 & L14--15 & L0--1 & L4--5 & L14--15 \\
\midrule
OLMo-1B         & \textbf{0.91} & 0.22 & 0.19 & \textbf{0.88} & 0.16 & 0.24 \\
OLMo-1B-DPO     & \textbf{0.84} & 0.20 & 0.35 & \textbf{0.81} & 0.18 & 0.31 \\
\bottomrule
\end{tabular}
}
\caption{Jaccard similarity of active heads (threshold $> 0.1$) between $k{=}1$ and $k{=}3$, by layer group. The activity in Layers 0-1 remains highly similar, while middle and later layer pairs are dissimilar, indicating that the increased $k$ changes activity patterns primarily in these layers.}
\label{tab:app:jaccard-similarity2}
\end{table}

\begin{table}[h!]
\centering
\resizebox{\linewidth}{!}{
\begin{tabular}{l ccc ccc}
\toprule
& \multicolumn{3}{c}{\texttt{adj\_comp}} & \multicolumn{3}{c}{\texttt{adj\_ant}} \\
\cmidrule(lr){2-4} \cmidrule(lr){5-7}
Model & L0--1 & L4--5 & L14--15 & L0--1 & L4--5 & L14--15 \\
\midrule
OLMo-1B         & \textbf{1.00} & \textbf{0.88} & \textbf{0.81} & \textbf{1.00} & \textbf{0.88} & \textbf{0.64} \\
OLMo-1B-DPO     & \textbf{1.00} & \textbf{0.76} & \textbf{0.90} & \textbf{1.00} & \textbf{0.68} & \textbf{0.85} \\
\bottomrule
\end{tabular}
}
\caption{Jaccard similarity of active heads (threshold $> 0.1$) between $k{=}2$ and $k{=}3$, by layer group. The high similarity scores indicate that there is little change in head activity from $k{=}2$ to $k{=}3$.}
\label{tab:app:jaccard-similarity3}
\end{table}

\begin{table}[h]
\centering
\small
\resizebox{\linewidth}{!}{
\begin{tabular}{lccc}
\toprule
\multicolumn{4}{c}{$\emT_\text{inst}-1$} \\
\midrule
\textbf{Model} & \textbf{Layers 0--2} & \textbf{Layers 6--9} & \textbf{Layers 10--15} \\
\midrule
OLMo-1B       & 0.375 & 0.000 & 0.714 \\
OLMo-1B-SFT   & 0.333 & 0.000 & 0.500 \\
OLMo-1B-DPO   & 0.000 & 0.000 & 0.444 \\
\midrule
\multicolumn{4}{c}{$\emT_\text{inst}$} \\
\midrule
\textbf{Model} & \textbf{Layers 0--2} & \textbf{Layers 6--9} & \textbf{Layers 10--15} \\
\midrule
OLMo-1B       & 0.762 & 0.500 & 0.714 \\
OLMo-1B-SFT   & 0.952 & 0.333 & 0.727 \\
OLMo-1B-DPO   & 0.909 & 0.286 & 0.500 \\
\midrule
\multicolumn{4}{c}{$\emT_\text{inst}+1$} \\
\midrule
\textbf{Model} & \textbf{Layers 0--2} & \textbf{Layers 6--9} & \textbf{Layers 10--15} \\
\midrule
OLMo-1B       & 0.250 & 0.500 & 0.714 \\
OLMo-1B-SFT   & 0.714 & 0.250 & 0.600 \\
OLMo-1B-DPO   & 0.625 & 0.500 & 0.500 \\
\bottomrule
\end{tabular}
}
\caption{Jaccard similarity of attention head activity ($k{=}1$) between the two contrastive \textsc{Adjectives} tasks at various token positions, over 100 samples.}
\label{tab:app:jac-sim-across-tasks-k1}
\end{table}

\begin{table}[h]
\centering
\small
\resizebox{\linewidth}{!}{
\begin{tabular}{lccc}
\toprule
\multicolumn{4}{c}{$\emT_\text{inst}-1$} \\
\midrule
\textbf{Model} & \textbf{Layers 0--2} & \textbf{Layers 6--9} & \textbf{Layers 10--15} \\
\midrule
OLMo-1B       & 0.769 & 0.759 & 0.577 \\
OLMo-1B-SFT   & 0.765 & 0.760 & 0.792 \\
OLMo-1B-DPO   & 0.667 & 0.708 & 0.826 \\
\midrule
\multicolumn{4}{c}{$\emT_\text{inst}$} \\
\midrule
\textbf{Model} & \textbf{Layers 0--2} & \textbf{Layers 6--9} & \textbf{Layers 10--15} \\
\midrule
OLMo-1B       & 1.000 & 0.931 & 0.577 \\
OLMo-1B-SFT   & 0.933 & 0.667 & 0.750 \\
OLMo-1B-DPO   & 0.935 & 0.700 & 0.826 \\
\midrule
\multicolumn{4}{c}{$\emT_\text{inst}+1$} \\
\midrule
\textbf{Model} & \textbf{Layers 0--2} & \textbf{Layers 6--9} & \textbf{Layers 10--15} \\
\midrule
OLMo-1B       & 0.524 & 0.583 & 0.577 \\
OLMo-1B-SFT   & 0.500 & 0.308 & 0.750 \\
OLMo-1B-DPO   & 0.538 & 0.429 & 0.846 \\
\bottomrule
\end{tabular}
}
\caption{Jaccard similarity of attention head activity ($k{=}2$) between the two contrastive \textsc{Adjectives} tasks at various token positions, over 100 samples.}
\label{tab:app:jac-sim-across-tasks-k2}
\end{table}

\begin{table}[h]
\centering
\small
\resizebox{\linewidth}{!}{
\begin{tabular}{lccc}
\toprule
\multicolumn{4}{c}{$\emT_\text{inst}-1$} \\
\midrule
\textbf{Model} & \textbf{Layers 0--2} & \textbf{Layers 6--9} & \textbf{Layers 10--15} \\
\midrule
OLMo-1B       & 0.931 & 1.000 & 0.839 \\
OLMo-1B-DPO   & 0.773 & 0.800 & 0.769 \\
\midrule
\multicolumn{4}{c}{$\emT_\text{inst}$} \\
\midrule
\textbf{Model} & \textbf{Layers 0--2} & \textbf{Layers 6--9} & \textbf{Layers 10--15} \\
\midrule
OLMo-1B       & 1.000 & 1.000 & 0.806 \\
OLMo-1B-DPO   & 0.969 & 0.679 & 0.769 \\
\midrule
\multicolumn{4}{c}{$\emT_\text{inst}+1$} \\
\midrule
\textbf{Model} & \textbf{Layers 0--2} & \textbf{Layers 6--9} & \textbf{Layers 10--15} \\
\midrule
OLMo-1B       & 0.577 & 0.871 & 0.839 \\
OLMo-1B-DPO   & 0.684 & 0.750 & 0.846 \\
\bottomrule
\end{tabular}
}
\caption{Jaccard similarity of attention head activity ($k{=}3$) between the two contrastive \textsc{Adjectives} tasks at various token positions, over 100 samples.}
\label{tab:app:jac-sim-across-tasks-k3}
\end{table}

\subsection{Initial Generalization to the \textsc{Animals} Tasks}
\label{app:subsec:animals}
\begin{figure}
    \centering
    \includegraphics[width=\linewidth,scale=0.7,trim={0 0cm 0cm 0cm},clip]{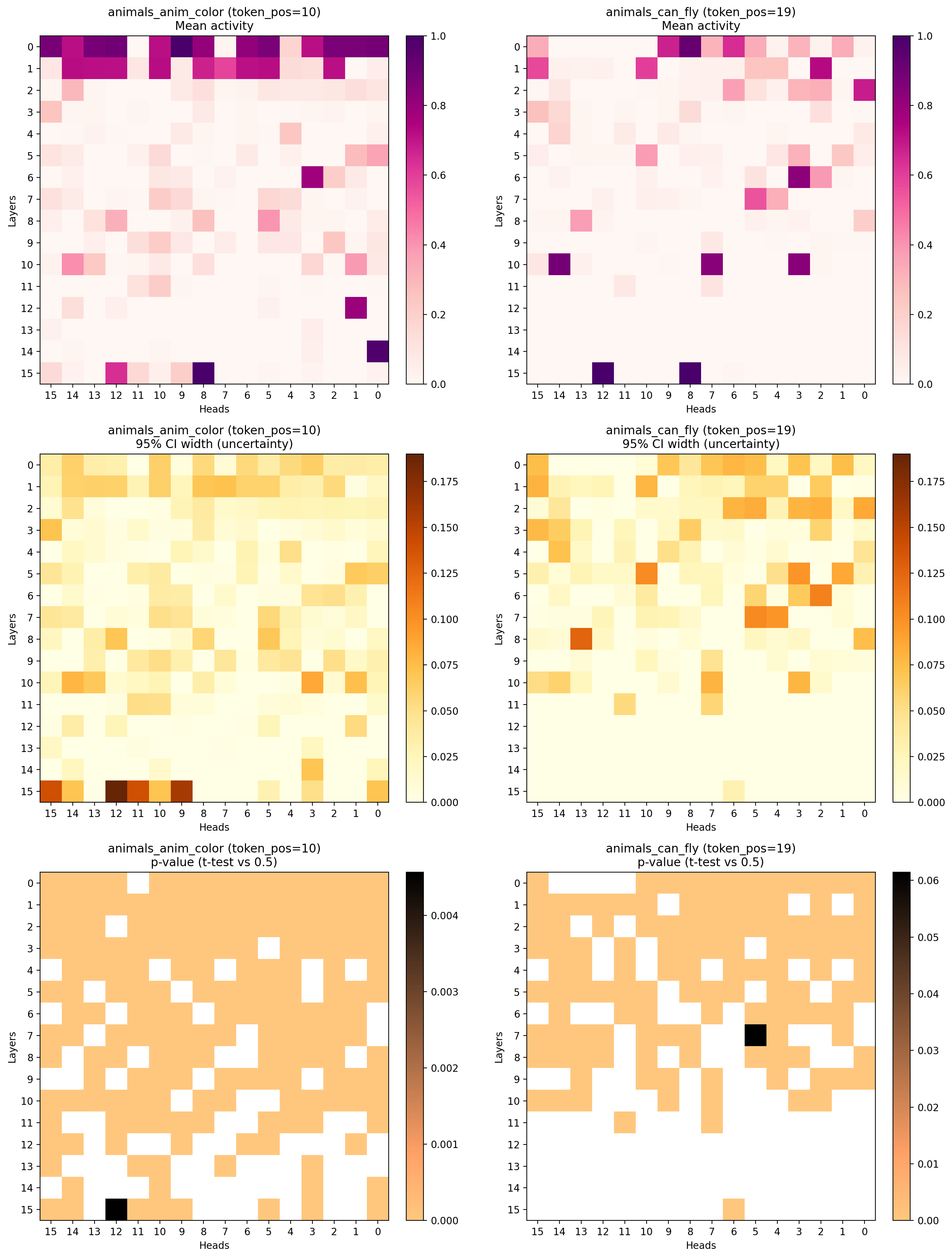}
    \caption{Results of our statistical tests of attention head activity, for OLMo-1B, $k{=}1$. Top: average activity across 100 samples. Middle: boostrapped confidence intervals, with lighter=narrower. Bottom: p-values showing significance of head variance against Gaussian noise, with lighter=lower p-value.}
    \label{fig:bootstrapped-animals-1}
\end{figure}

\begin{figure}
    \centering
    \includegraphics[width=\linewidth,scale=0.7,trim={0 0cm 0cm 0cm},clip]{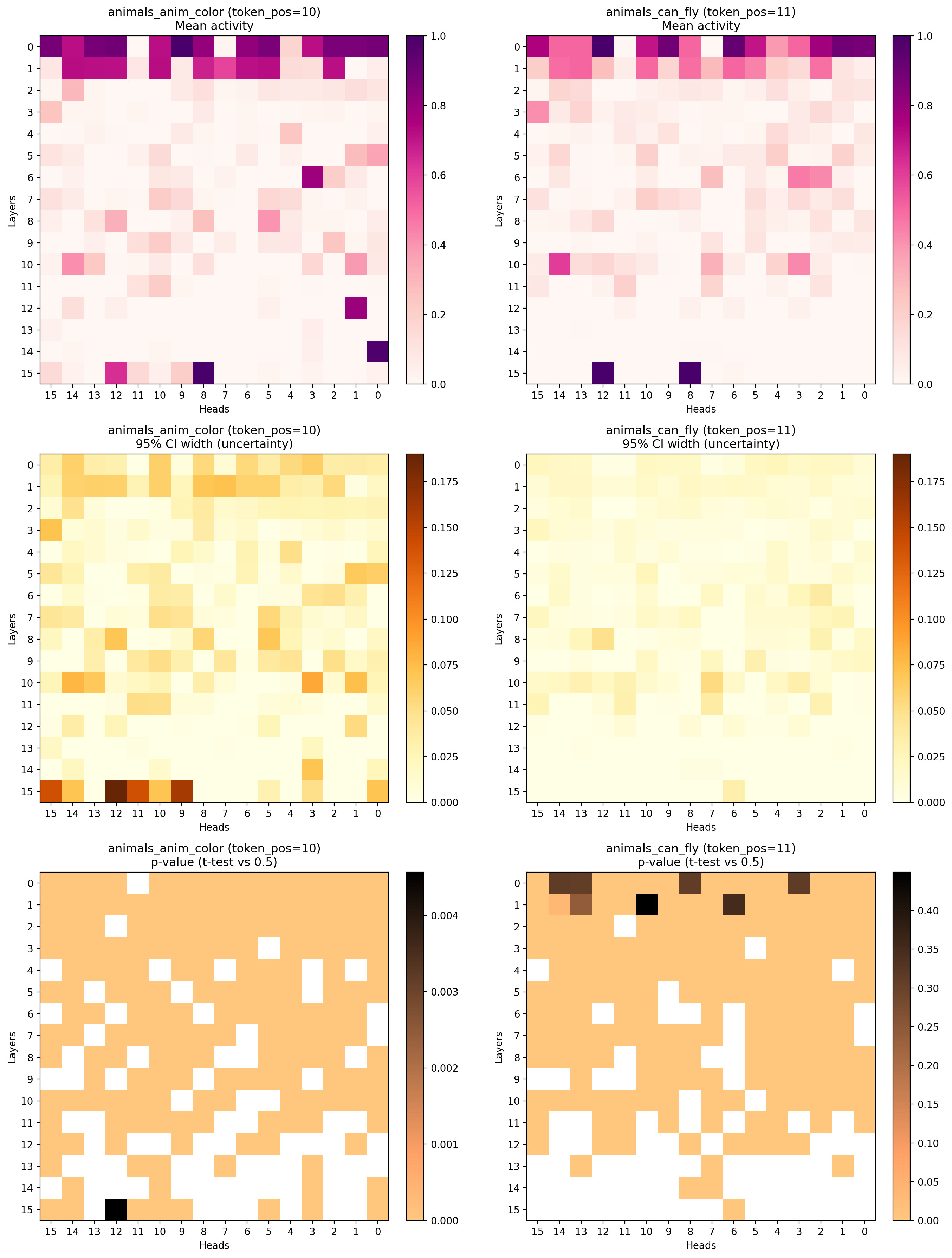}
    \caption{Results of our statistical tests of attention head activity, for OLMo-1B, $k{=}1$. Top: average activity across 100 samples. Middle: boostrapped confidence intervals, with lighter=narrower. Bottom: p-values showing significance of head variance against Gaussian noise, with lighter=lower p-value. \textbf{Note that the comparison is between the $\emT_\text{inst}$ token of \textsc{Animals: Color} and an earlier token of \textsc{Animals: Can\_Fly} that represents the first complete \enquote{sub-instruction}.}}
    \label{fig:bootstrapped-animals-2}
\end{figure}

\begin{figure}
    \centering
    \includegraphics[width=\linewidth,scale=0.7,trim={0 0cm 0cm 0cm},clip]{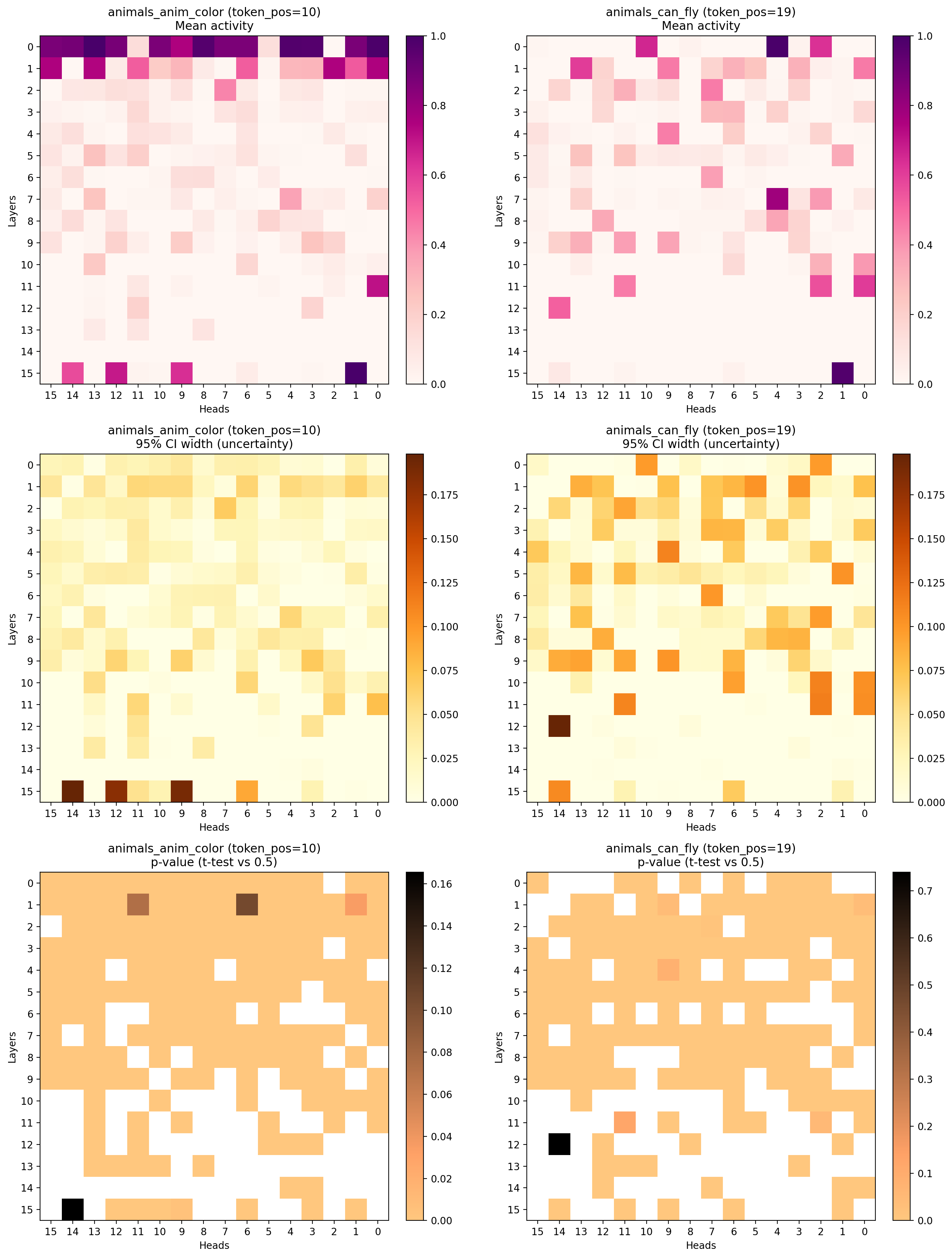}
    \caption{Results of our statistical tests of attention head activity, for OLMo-1B-SFT, $k{=}1$. Top: average activity across 100 samples. Middle: boostrapped confidence intervals, with lighter=narrower. Bottom: p-values showing significance of head variance against Gaussian noise, with lighter=lower p-value. }
    \label{fig:bootstrapped-animals-3}
\end{figure}

\begin{figure}
    \centering
    \includegraphics[width=\linewidth,scale=0.7,trim={0 0cm 0cm 0cm},clip]{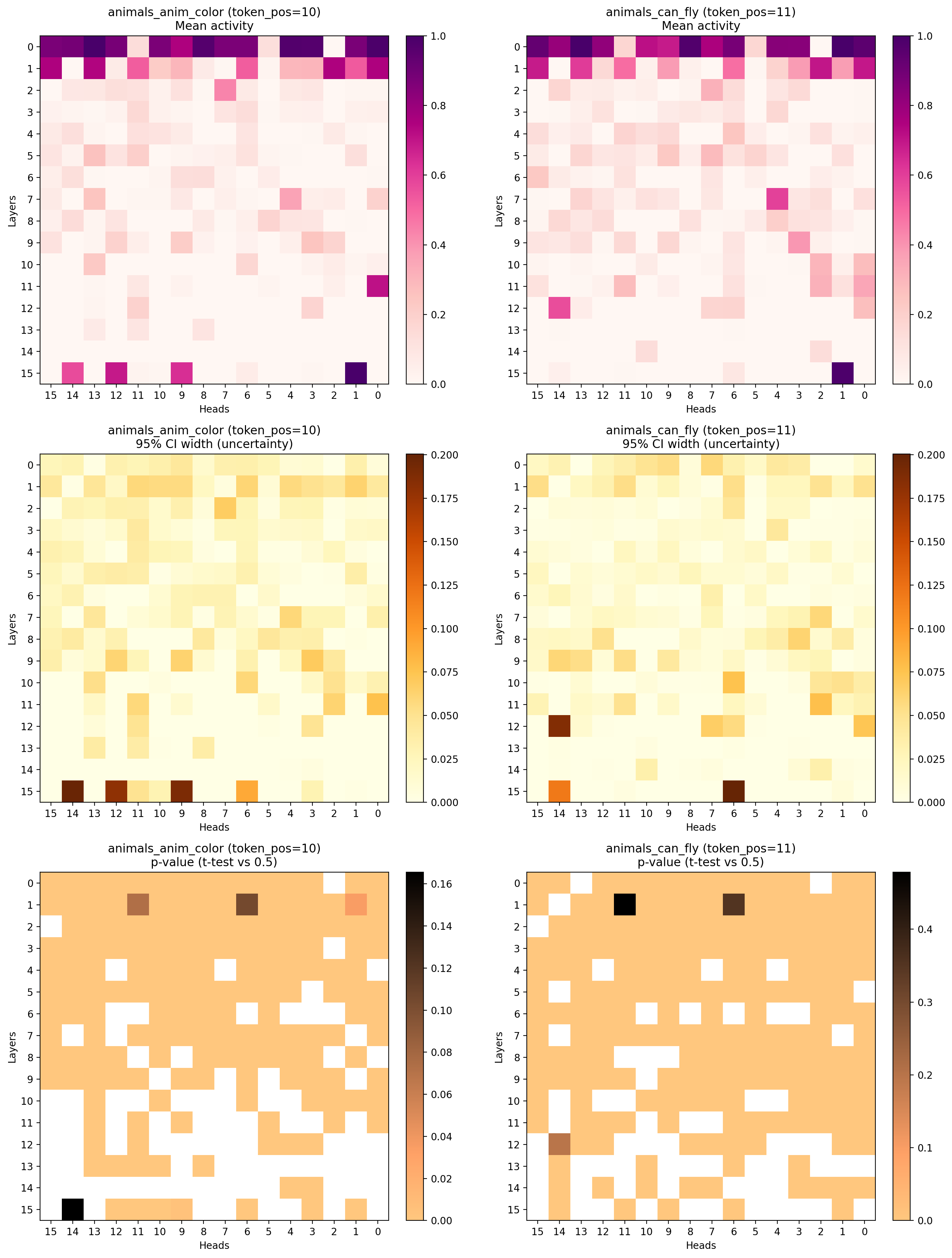}
    \caption{Results of our statistical tests of attention head activity, for OLMo-1B-SFT, $k{=}1$. Top: average activity across 100 samples. Middle: boostrapped confidence intervals, with lighter=narrower. Bottom: p-values showing significance of head variance against Gaussian noise, with lighter=lower p-value. \textbf{Note that the comparison is between the $\emT_\text{inst}$ token of \textsc{Animals: Color} and an earlier token of \textsc{Animals: Can\_Fly} that represents the first complete \enquote{sub-instruction}.}}
    \label{fig:bootstrapped-animals-4}
\end{figure}

\begin{figure}
    \centering
    \includegraphics[width=\linewidth,scale=0.7,trim={0 0cm 0cm 0cm},clip]{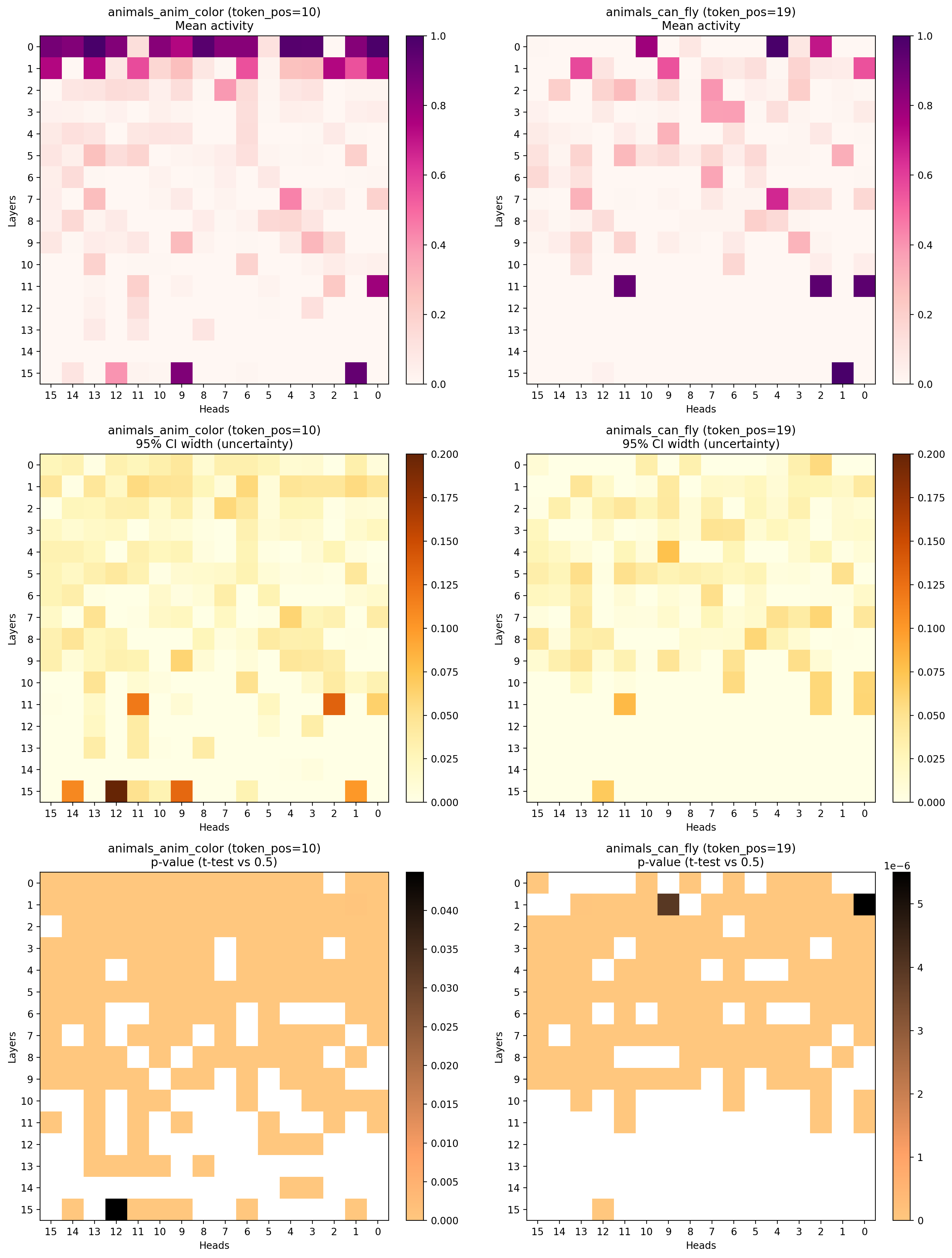}
    \caption{Results of our statistical tests of attention head activity, for OLMo-1B-DPO, $k{=}1$. Top: average activity across 100 samples. Middle: boostrapped confidence intervals, with lighter=narrower. Bottom: p-values showing significance of head variance against Gaussian noise, with lighter=lower p-value.}
    \label{fig:bootstrapped-animals-5}
\end{figure}

\begin{figure}
    \centering
    \includegraphics[width=\linewidth,scale=0.7,trim={0 0cm 0cm 0cm},clip]{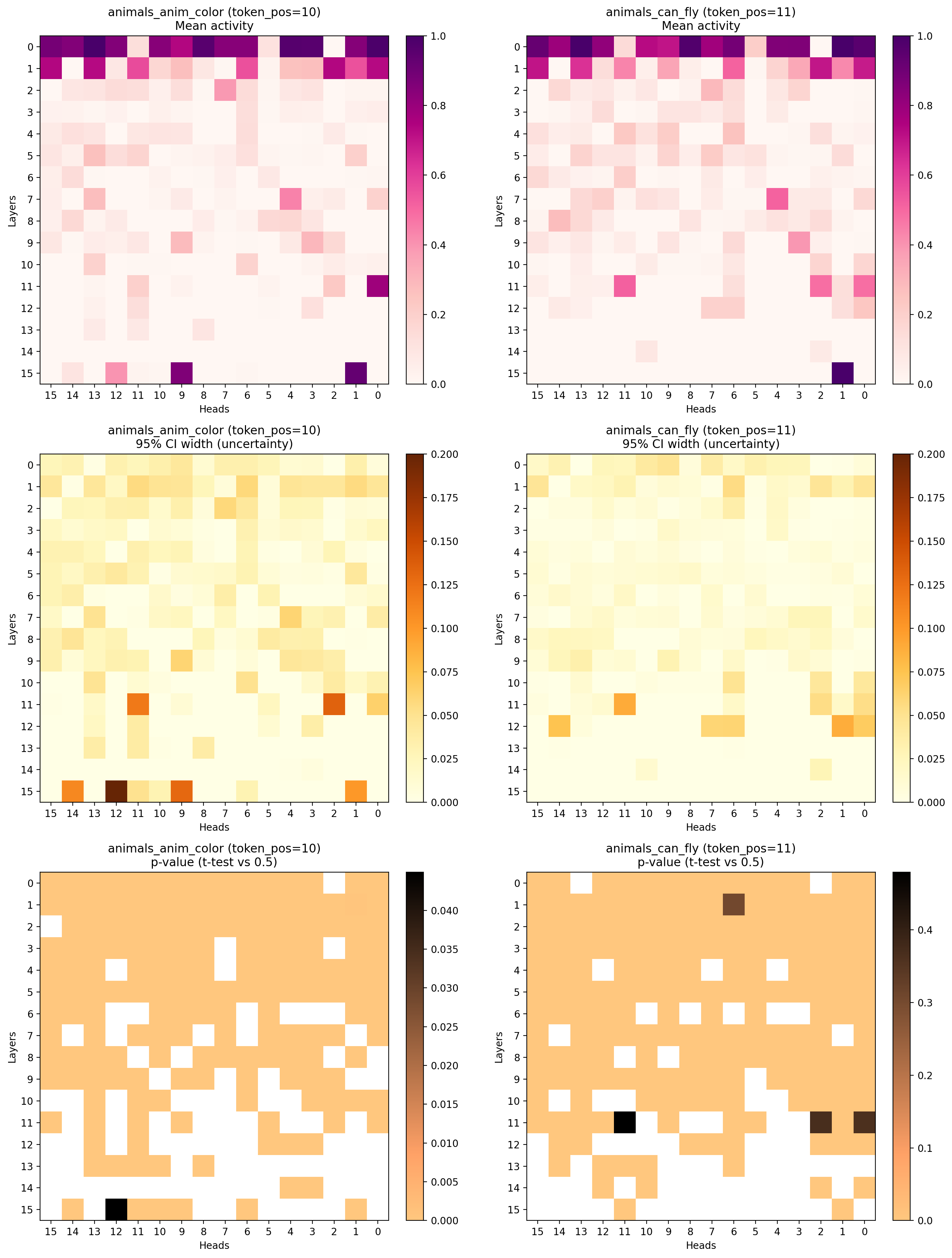}
    \caption{Results of our statistical tests of attention head activity, for OLMo-1B-DPO, $k{=}1$. Top: average activity across 100 samples. Middle: boostrapped confidence intervals, with lighter=narrower. Bottom: p-values showing significance of head variance against Gaussian noise, with lighter=lower p-value. \textbf{Note that the comparison is between the $\emT_\text{inst}$ token of \textsc{Animals: Color} and an earlier token of \textsc{Animals: Can\_Fly} that represents the first complete \enquote{sub-instruction}.}}
    \label{fig:bootstrapped-animals-6}
\end{figure}

In Figures \ref{fig:bootstrapped-animals-1} - \ref{fig:bootstrapped-animals-6}, we present the attention head activity patterns for \textsc{Animals: Color} and \textsc{Animals: Can\_Fly}, with $k{=}1$. Due to resource constraints, we do not conduct path analysis on these tasks. However, we observe that our overall conclusions for the \textsc{Adjectives} tasks hold, with some interesting insights regarding eager instruction representations, as discussed in Section \ref{sec:path_experiments}.

\end{document}

%% file: latex/math_commands.tex
\usepackage{amsmath,amsfonts,bm}
\usepackage{dsfont}

\def\eqref#1{equation~\ref{#1}}

\def\1{\bm{1}}

\def\vzero{{\bm{0}}}

\def\va{{\bm{a}}}

\def\ve{{\bm{e}}}

\def\vx{{\bm{x}}}
\def\vy{{\bm{y}}}

\def\eva{{a}}

\def\mD{{\bm{D}}}
\def\mE{{\bm{E}}}

\def\mI{{\bm{I}}}

\def\mP{{\bm{P}}}

\def\mT{{\bm{T}}}
\def\mU{{\bm{U}}}
\def\mV{{\bm{V}}}
\def\mW{{\bm{W}}}
\def\mX{{\bm{X}}}

\DeclareMathAlphabet{\mathsfit}{\encodingdefault}{\sfdefault}{m}{sl}
\SetMathAlphabet{\mathsfit}{bold}{\encodingdefault}{\sfdefault}{bx}{n}

\def\sR{{\mathbb{R}}}

\def\sV{{\mathbb{V}}}

\def\emT{{T}}

\def\emX{{X}}

\DeclareMathOperator*{\argmax}{arg\,max}